\documentclass[default,iicol]{sn-jnl}

\jyear{2023}%

\theoremstyle{thmstyletwo}%

\theoremstyle{thmstylethree}%

\usepackage{times}

\usepackage[numbers]{natbib}
\usepackage{multicol}

\usepackage{graphicx}
\usepackage{subfigure}
\usepackage{wrapfig}
\usepackage{times}
\usepackage{amsmath}
\usepackage{amssymb}
\usepackage{graphics}
\usepackage{epsfig}
\usepackage{multirow}
\usepackage{wrapfig}
\usepackage{amsthm}
\usepackage{mathtools}
\usepackage{makecell}
\usepackage{bbm}
\usepackage{booktabs}
\usepackage{graphicx}
\usepackage{color,soul}

\usepackage{amsmath}
\usepackage{amssymb}
\usepackage{amsthm}

\newcommand{\SE}{\mathrm{SE}}
\newcommand{\SO}{\mathrm{SO}}

\DeclareMathOperator*{\argmax}{arg\,max}

\begin{document}


\title[On Robot Grasp Learning Using Equivariant Models]{On Robot Grasp Learning Using Equivariant Models}


\author{\fnm{Xupeng} \sur{Zhu}}\email{zhu.xup@northeastern.edu}

\author{\fnm{Dian} \sur{Wang}}\email{wang.dian@northeastern.edu}

\author{\fnm{Guanang} \sur{Su}}\email{su.gu@northeastern.edu}

\author{\fnm{Ondrej} \sur{Biza}}\email{biza.o@northeastern.edu}

\author{\fnm{Robin} \sur{Walters}}\email{r.walters@northeastern.edu}

\author{\fnm{Robert} \sur{Platt}}\email{r.platt@northeastern.edu}

\affil{\orgdiv{Khoury College of Computer Science}, \orgname{Northeastern University}, \orgaddress{\street{Huntington Ave}, \city{Boston}, \postcode{02115}, \state{Massachusetts}, \country{USA}}}




\abstract{
Real-world grasp detection is challenging due to the stochasticity in grasp dynamics and the noise in hardware. Ideally, the system would adapt to the real world by training directly on physical systems. However, this is generally difficult due to the large amount of training data required by most grasp learning models. In this paper, we note that the planar grasp function is $\SE(2)$-equivariant and demonstrate that this structure can be used to constrain the neural network used during learning. This creates an inductive bias that can significantly improve the sample efficiency of grasp learning and enable end-to-end training from scratch on a physical robot with as few as $600$ grasp attempts. We call this method Symmetric Grasp learning (SymGrasp) and show that it can learn to grasp ``from scratch'' in less that 1.5 hours of physical robot time.



}




\maketitle

\section{Introduction}\label{sec1}

\emph{Grasp detection} detects good grasp poses in a scene directly from raw visual input (e.g., RGB or depth images) using machine learning. The learning-based method generalizes to novel objects.
This is in contrast to classical model-based methods that attempt to reconstruct the geometry and the pose of objects in a scene and then reason geometrically about how to grasp those objects. 


Most current grasp detection models are data-driven, i.e., they must be trained using large offline datasets. For example, \cite{mousavian20196} trains on a dataset consisting of over 7M simulated grasps, \cite{breyer2021volumetric} trains on over 2M simulated grasps, \cite{mahler2017dex} trains on grasp data drawn from over 6.7M simulated point clouds, and \cite{gpd} trains on over 700k simulated grasps. Some models are trained using datasets obtained via physical robotic grasp interactions. For example,  \cite{supersizing_self_supervision} trains on a dataset created by performing 50k grasp attempts over 700 hours, \cite{qt-opt} trains on over 580k grasp attempts collected over the course of 800 robot hours, and \cite{berscheid2021robot} train on a dataset obtained by performing 27k grasps over 120 hours. 


Such reliance on collecting large datasets necessitates either learning in simulation or using significant amounts of robot time to generate data, motivating the desire for a more sample efficient grasp detection model, i.e., a model that can achieve good performance with a smaller dataset. In this paper, we propose a novel grasp detection strategy that improves sample efficiency significantly by incorporating the equivariant structure into the model. We term our strategy Symmetric Grasp learning as SymGrasp. Our key observation is that the target grasp function (from images onto grasp poses) is $\SE(2)$-equivariant. That is, rotations and translations of the input image should correspond to the same rotations and translations of the detected grasp poses at the output of the function. In order to encode 
the $\SE(2)$-equivariance in the target function,
we constrain the layers of our model to respect this symmetry. Compared with conventional grasp detection models that must be trained using tens of thousands of grasp experiences, the equivariant structure we encode into the model enables us to achieve good grasp performance after only a few hundred grasp attempts. 

This paper makes several key contributions. First, we recognize that the grasp detection function from images to grasp poses is a $\SE(2)$-equivariant function. Then, we propose a neural network model using equivariant layers to encode this property. Finally, we introduce several algorithmic optimizations that enable us to learn to grasp online using a contextual bandit framework. Ultimately, our model is able to learn to grasp opaque (using depth images) and transparent objects (using RGB images) with a good success rate after only approximately 600 grasp trials -- 1.5 hours of robot time. Although the model we propose here is only for 2D grasping (i.e., we only detect top down grasps rather than all six dimensions as in 6-DOF grasp detection), the sample efficiency is still impressive and we believe the concepts could be extended to higher-DOF grasp detection models in the future.

These improvements in sample efficiency are important for several reasons. First, since our model can learn to grasp in only a few hundred grasp trials, it can be trained easily on a physical robotic system. This greatly reduces the need to train on large datasets created in simulation, and it therefore reduces our exposure to the risks associated with bridging the sim2real domain gap -- we can simply do all our training on physical robotic systems. Second, since we are training on a small dataset, it is much easier to learn on-policy rather than off-policy, i.e., we can train using data generated by the policy being learned rather than with a fixed dataset. This focuses learning on areas of state space explored by the policy and makes the resulting policies more robust in those areas. Finally, since we can learn efficiently from a small number of experiences, our policy has the potential to adapt relatively quickly at run time to physical changes in the robot sensors and actuators.

\section{Related Work}

\subsection{Equivariant convolutional layers}

Equivariant convolutional layers incorporate symmetries into the structure of convolutional layers, allowing them to generalize across a symmetry group automatically. This idea was first introduced as G-Convolution~\citep{g_conv} and Steerable CNN~\citep{steerable_cnns}. E2CNN is a generic framework for implementing $\mathrm{E}(2)$ Steerable CNN layers~\citep{e2cnn}. In applications such as dynamics ~\citep{walters2020trajectory,wang2020incorporating} and reinforcement learning~\citep{van2020mdp,mondal2020group,wang2021equivariant,wang2022equivariant} equivariant models demonstrate improvements over traditional approaches. 


\subsection{Sample efficient reinforcement learning}

Recent work has shown that data augmentation using random crops and/or shifts can improve the sample efficiency of standard reinforcement learning algorithms~\citep{rad, DrQ}. It is possible to improve sample efficiency even further by incorporating contrastive learning~\citep{oord2018representation}, e.g. CURL~\cite{curl}. The contrastive loss enables the model to learn an internal latent representation that is invariant to the type of data augmentation used. The FERM framework~\cite{FERM} applies this idea to robotic manipulation and is able to learn to perform simple manipulation tasks directly on physical robotic hardware. The equivariant models used in this paper are similar to data augmentation in that the goal is to leverage problem symmetries to accelerate learning. However, whereas data augmentation and contrastive approaches require the model to \emph{learn} an invariant or equivariant encoding, the equivariant model layers used in this paper \emph{enforce} equivariance as a prior encoded in the model. This simplifies the learning task and enables our model to learn faster (see Section~\ref{sect:experiments}).

\subsection{Grasp detection}

In grasp detection, the robot finds grasp configurations directly from visual or depth data. This is in contrast to classical methods which attempt to reconstruct object or scene geometry and then do grasp planning. See \cite{platt2022grasp} for a review on this topic.

\noindent
\underline{2D Grasping:} Several methods are designed to detect grasps in 2D, i.e., to detect the planar position and orientation of grasps in a scene based on top-down images. A key early example of this was DexNet 2.0, which infers the quality of a grasp centered and aligned with an oriented image patch~\cite{mahler2017dex}. Subsequent work proposed fully convolutional architectures, thereby enabling the model to quickly infer the pose of \emph{all} grasps in a (planar) scene~\cite{morrison2018closing,4Dof, jacquard,3d_generative_res_conv,zhou2018fully} (some of these models infer the $z$ coordinate of the grasp as well).

\noindent
\underline{3D Grasping:} There is much work in 3D grasp detection, i.e., detecting the full 6-DOF position and orientation of grasps based on truncated signed distance function (TSDF) or point cloud input. A key early example of this was GPD~\cite{tenpas_ijrr2017} which inferred grasp pose based on point cloud input. Subsequent work has focused on improving grasp candidate generation in order to improve efficiency, accuracy, and coverage~\cite{huang2022edge, mousavian20196,sundermeyer2021contact,jiang2021synergies,graspnet-1b,breyer2021volumetric,berscheid2021robot}. 

\noindent
\underline{On-robot Grasp Learning:} Another important trend has been learning to grasp directly from physical robotic grasp experiences. Early examples of this include~\cite{supersizing_self_supervision} who learn to grasp from 50k grasp experiences collected over 700 hours of robot time and~\cite{levine2018learning} who learn a grasp policy from 800k grasp experiences collected over two months. QT-Opt~\citep{qt-opt} learns a grasp policy from 580k grasp experiences collected over 800 hours and~\cite{james2019sim} extends this work by learning from an additional 28k grasp experiences. \cite{song2020grasping} learns a grasp detection model from 8k grasp demonstrations collected via demonstration and~\cite{synergy} learns a online pushing/grasping policy from just 2.5k grasps.

\noindent
\underline{Transparent objects grasping:} Commercial depth sensors that are based on structured-light or time-of-flight techniques often fail to sense transparent objects accurately~\cite{weng2020multi}. Pixels in the depth image are often dropped due to specularities or the object is simply invisible to the sensor because it is transparent~\cite{sajjan2020clear}. To avoid this type of failure, RGB or RGB-D sensors are commonly used. \cite{weng2020multi} infers grasp pose from RGBD images. They collect paired RGB and D images on opaque objects and utilize transfer learning from a trained D modality model to an RGBD modality model. \cite{sajjan2020clear} reconstructs depth images from RGBD images using CNN and then performs grasp detection. Likewise, \cite{ichnowski2021dex} and \cite{kerrevo} infer grasp pose from reconstructed depth images, but using neural radiance field (NeRF) \cite{mildenhall2021nerf} for reconstruction. These methods either rely on collecting paired images for training or require training a NeRF model per grasp during evaluation. In contrast, our method is trained directly on RGB images without requiring paired images or NeRF models.


\noindent
\underline{Equivariance through canonicalization in grasping:} An alternative to modeling rotational symmetry using equivariant neural network layers is an approach known as \emph{canonicalization} where we learn a model over the non-equivariant variables assuming a single ``canonical'' group element~\cite{wang2020incorporating,kofinas2021roto,gao2020vectornet}. Equivariance based on canonicalization is common in robotic grasping where it is not unusual to translate and rotate the input image so that it is expressed in the reference frame of the hand, e.g.~\cite{mahler2017dex,tenpas_ijrr2017,mousavian20196}. This way, the neural network model need only infer the quality of a grasp for a single canonical grasp pose rather than over arbitrary translations and orientations. In this paper, we compare our model-based approach to equivariance with VPG, a method that obtains rotational equivariance via canonicalization~\cite{song2020grasping}. Our results in Section~\ref{sect:baselines} suggest that the model-based approach has a significant advantage.

\section{Background}

\subsection{Equivariant Neural Network Models}






In this paper, we use equivariant neural network layers defined with respect to a finite group (e.g. a finite group of rotations). Equivariant neural network \cite{e2cnn, equi_theory, steerable_cnns, g_conv} encodes a function $f : X \rightarrow Y$ that satisfy the equivariance constraint: $gf(x) = f(gx)$, where $g \in G$ is an element of a finite group. $gx$ is shorthand for the action of $g$ on $x$, e.g. rotation of an image $x$. Similarly, $g(f(x))$ describes the action of $g$ on $f(x)$. Below, we make these ideas more precise and summarize how the equivariance constraint is encoded into a neural network layer.

\subsubsection{The cyclic group $C_n \leq \SO(2)$}

We are primarily interested in equivariance with respect to the group of planar rotations, $\SO(2)$. However, in practice, in order to make our models computationally tractable, we will use the cyclic subgroup $C_n$ of $\SO(2)$,  $C_n=\{ 2  \pi k / n  : 0\leq k < n\}$. $C_n$ is the group of discrete rotations by multiples of $2\pi/n$ radians.

\subsubsection{Representation of a group}

The way a group element $g \in G$ acts on $x$ depends on how $x$ is represented. If $x$ is a point in the plane, then $g$ acts on $x$ via the \emph{standard representation}, $\rho_1(g)x$, where $\rho_1(g)$ is the standard $2 \times 2$ rotation matrix corresponding to $g$. In the hidden layers of an equivariant neural network model, it is common to encode a separate feature map for each group element. For example, suppose $G$ is the order $n$ cyclic group and suppose $x$ is a set of features $x = (x_1, x_2, \dots, x_n) \in \mathbb{R}^{n \times \lambda}$ that maps the $k$th group element to a feature $x_k$. The \emph{regular representation} of $g$ acts on $x$ by permuting its elements: $\rho_{reg}(g)x = (x_{n-m+1}, \dots, x_n, x_1, x_2, \dots, x_{n-m})$ where $g$ is the $m$th element in $C_n$. Finally, it is sometimes the case that $x$ is invariant to the action of the group elements. In this case, we have the \emph{trivial representation}, $\rho_0(g)x = x$.

\subsubsection{Feature maps of equivariant convolutional layers} 

An equivariant convolutional layer maps between feature maps which transform by specified representations $\rho$ of the group. In the hidden layers of an equivariant model, generally, an extra dimension is added to the feature maps to encode group elements via the regular representation. 
So, whereas the feature map used by a standard convolutional layer is a tensor $\mathcal{F} \in \mathbb{R}^{m \times h \times w}$, 
an equivariant convolutional layer adds an extra dimension: $\mathcal{F} \in \mathbb{R}^{k \times m \times h \times w}$, where $k$ denotes the dimension of the group representation. This tensor associates each pixel $(u, v) \in \mathbb{R}^{h \times w}$ with a matrix $\mathcal{F}(u, v) \in \mathbb{R}^{k \times m}$.

\subsubsection{Action of the group operator on the feature map} 

Given a feature map $\mathcal{F} \in \mathbb{R}^{k \times m \times h \times w}$ associated with group $G$ and representation $\rho$, a group element $g \in G$ acts on $\mathcal{F}$ via:
\begin{equation}
(g \mathcal{F})(x) = \rho(g) \mathcal{F} (\rho_1(g)^{-1} x),
\label{eqn:group_operator}
\end{equation} 
where $x \in \mathbb{R}^2$ denotes pixel position. The RHS of this equation applies the group operator in two ways. First, $\rho_1(g)^{-1}$ rotates the pixel position $x$ using the standard representation. Second, $\rho$ applies the rotation to the feature representation. If the feature is invariant to the rotation, then we use the trivial representation $\rho_0(g)$. However, if the feature vector changes according to rotation (e.g. the feature denotes grasp orientation), then it must be transformed as well. This is accomplished by setting $\rho$ in Equation~\ref{eqn:group_operator} to be the regular representation that transforms the feature vector by a circular shift.




\subsubsection{The equivariant convolutional layer} 

An equivariant convolutional layer is a function $h$ from $\mathcal{F}_{in}$ to $\mathcal{F}_{out}$ that is constrained to represent only equivariant functions with respect to a chosen group $G$. The feature maps $\mathcal{F}_{in}$ and $\mathcal{F}_{out}$ are associated with representations $\rho_{in}$ and $\rho_{out}$ acting on feature spaces $\mathbb{R}^{k_{in}}$ and $\mathbb{R}^{k_{out}}$ respectively. 
Then the equivariant constraint for $h$ is~\cite{equi_theory}:
\begin{equation}
\label{eqn:equi_conv}
h(g \mathcal{F}_{in}) = g h(\mathcal{F}_{in}) = g \mathcal{F}_{out}.
\end{equation}
This constraint can be implemented by tying kernel weights $K(y) \in \mathbb{R}^{k_{out} \times k_{in}}$ in such a way as to satisfy the following constraint~\citep{equi_theory}:
\begin{equation}
K(gy) = \rho_{out}(g) K(y)\rho_{in}(g)^{-1}.
\end{equation}
Please see Appendix.\ref{sec:equ_example} for an example of equivariant convolustional layer.

When all hidden layers $h$ in a neural network satisfy Equation~\ref{eqn:equi_conv}, then by induction the entire neural network is equivariant~\cite{equi_theory}.

\subsection{Augmented State Representation (ASR)} 
\label{sect:asr}

We will formulate $\SE(2)$ robotic grasping as the problem of learning a function from an $m$ channel image, $s \in S = \mathbb{R}^{m \times h \times w}$, to a gripper pose $a \in A = \SE(2)$ from which an object may be grasped. Since we will use the contextual bandit framework, we need to be able to represent the $Q$-function, $Q : \mathbb{R}^{m \times h \times w} \times \SE(2) \rightarrow \mathbb{R}$. However, this is difficult to do using a single neural network due to the GPU memory limitation. To combat this, we will use the Augmented State Representation (ASR)~\citep{sharma2017learning,asr} to model $Q$ as a pair of functions, $Q_1$ and $Q_2$. Another advantage of using ASR is we can use different group order in $Q_1, Q_2$, as explained in Section~\ref{sec:equ_asr}.

We follow the ASR framework that factors $\SE(2) = \mathbb{R}^2 \times \SO(2)$ into a translational component $X \subseteq \mathbb{R}^2$ and a rotational component $\Theta \subseteq \SO(2)$. The first function is a mapping $Q_1 : \mathbb{R}^{m \times h \times w} \times X \rightarrow \mathbb{R}$ which maps from the image $s$ and the translational component of action $X$ onto value. This function is defined to be: $Q_1(s,x) = \max_{\theta \in \Theta} Q(s,(x,\theta))$. The second function is a mapping $Q_2 : \mathbb{R}^{m \times h' \times w'} \times \Theta \rightarrow \mathbb{R}$ with $h' \leq h$ and $w' \leq w$ which maps from an image patch and an orientation onto value. This function takes as input a cropped version of $s$ centered on a position $x$, $\mathrm{crop}(s,x)$, and an orientation, $\theta$, and outputs the corresponding $Q$ value: $Q_2(\mathrm{crop}(s,x),\theta) = Q(s,(x,\theta))$.

\begin{figure}
    \centering
    \includegraphics[width=0.45\textwidth]{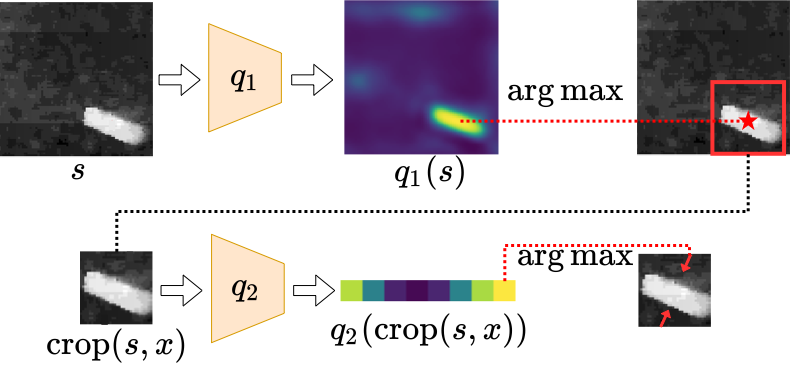}
    \caption{Illustration of the ASR representation. $Q_1$ selects the translational component of an action, $Q_2$ selects the rotational component.}
    \label{fig:asr}
\end{figure}

Inference is performed on the model by evaluating $x^* = \argmax_{x \in X} Q_1(s,x)$ first and then evaluating $\theta^* = \argmax_{\theta}Q_2(\mathrm{crop}(s,x^*),\theta)$. Since each of these two models, $Q_1$ and $Q_2$, are significantly smaller than $Q$ would be, the inference is much faster. Figure~\ref{fig:asr} shows an illustration of this process. The top of the figure shows the action of $Q_1$ while the bottom shows $Q_2$. Notice that the semantics of $Q_2$ imply that the $\theta$ depends only on $\mathrm{crop}(s,x)$, a local neighborhood of $x$, rather than on the entire scene. This assumption is generally true for grasping because grasp orientation typically depends only on the object geometry near the target grasp point.

\section{Problem Statement}

\subsection{Planar grasp detection}

The \emph{planar grasp detection function} $\Gamma : \mathbb{R}^{m \times h \times w} \rightarrow \SE(2)$ maps from a top down image of a scene containing graspable objects, $s \in S = \mathbb{R}^{m \times h \times w}$, to a planar gripper pose, $a \in A = \SE(2)$, from which an object can be grasped. This is similar to the formulations used by~\cite{mahler2017dex,morrison2018closing}.

\subsection{Formulation as a Contextual Bandit}
\label{sect:bandit}

We formulate grasp learning as a contextual bandit problem where the state is an image $s \in S = \mathbb{R}^{m \times h \times w}$ and the action $a \in A = \SE(2)$ is a grasp pose to which the robot hand will be moved and a grasp will be attempted, expressed in the reference frame of the image. After each grasp attempt, the agent receives a binary reward $R$ drawn from a Bernoulli distribution with unknown probability $r(s, a)$. The true $Q$-function denotes the expected reward of taking action $a$ from $s$. Since $R$ is binary, we have that $Q(s, a) = r(s, a)$. This formulation of grasp learning as a bandit problem is similar to that used by, e.g.~\cite{danielczuk2020exploratory,qt-opt,synergy}.

\subsection{Invariance Assumption}
\label{sec:inv_assumption}

We assume that the (unknown) reward function $r(s,a)$ that denotes the probability of a successful grasp is invariant to translations and rotations $g \in \SE(2)$. Let $gs$ denote the image $s$ translated and rotated by $g$. Similarly, let $ga$ denote the action translated and rotated by $g$. Therefore, our assumption is:
\begin{equation}
r(s,a) = r(gs,ga).
\label{eqn:grasp_invariance_assumption}
\end{equation}
Intuitively, when the image of a scene transforms, the grasp poses (located with respect to the image) transform correspondingly.

\section{SymGrasp: Symmetric Grasp Learning}
\label{sect:method}
\subsection{Equivariant Learning}
\label{sec:equi_learning}

We use equivariant neural networks \cite{e2cnn} to model the $Q$-function that enforces the invariance assumption in Section~\ref{sec:inv_assumption}. Therefore once the $Q$ function is fit to a data sample $r(s, a)$, it generalizes to any $g\in\SE(2)$ transformed data sample $r(gs, ga)$. This generalization could lead to a significant sample efficiency improvement during training.


\subsubsection{Invariance properties of $Q_1$ and $Q_2$} 

The assumption that the reward function $r$ is invariant to transformations $g \in \SE(2)$ implies that the optimal $Q$-function is also invariant to $g$, i.e., $Q(s, a) = Q(gs,ga)$.
In the context of the augmented state representation (ASR, see Section~\ref{sect:asr}), this implies separate invariance properties for $Q_1$ and $Q_2$:
\begin{align}
\label{eqn:q1_inv}
Q_1(gs, gx) &= Q_1(s, x) \\
\label{eqn:q2_inv}
Q_2(g_\theta (\mathrm{crop}(s, x)), g_\theta +\theta) &= Q_2(\mathrm{crop}(s, x), \theta),
\end{align}
where $g_\theta \in \SO(2)$ denotes the rotational component of $g\in \SE(2)$, $gx$ denotes the rotated and translated vector $x \in \mathbb{R}^2$, and $g_\theta (\mathrm{crop}(s, x))$ denotes the cropped image rotated by $g_\theta$.



\subsubsection{Discrete Approximation of $\SE(2)$} 

We implement the invariance constraints of Equation~\ref{eqn:q1_inv} and~\ref{eqn:q2_inv} using a discrete approximation to $\SE(2)$.  We constrain the positional component of the action to be a discrete pair of positive integers $x \in \{1\dots h\}\times  \{1\dots w\}\subset \mathbb{Z}^2$, corresponding to a pixel in $s$, and constrain the rotational component of the action to be an element of the finite cyclic group $C_n = \{2\pi k/n : 0\leq k < n, i\in \mathbb{Z}\}$. This discretized action space will be written $\hat{\SE}(2) = \mathbb{Z}^2 \times C_n$.


\subsubsection{Equivariant $Q$-Learning with ASR}
\label{sec:equ_asr}
In $Q$-Learning with ASR, we model $Q_1$ and $Q_2$ as neural networks. We model $Q_1$ as a fully convolutional UNet~\citep{u_net} $q_1: \mathbb{R}^{m \times h\times w}\to \mathbb{R}^{1\times h\times w}$ that generates $Q$ value for each discretized translational actions from the input state image.
We model $Q_2$ as a standard convolutional network $q_2: \mathbb{R}^{m \times h'\times w'}\to \mathbb{R}^n$ that evaluates $Q$ value for $n$ discretized rotational actions based on the image patch.
The networks $q_1$ and $q_2$ thus model the functions $Q_1$ and $Q_2$ by partially evaluating at the first argument and returning a function in the second.  As a result, the invariance properties of $Q_1$ and $Q_2$ (Equation~\ref{eqn:q1_inv} and~\ref{eqn:q2_inv}) imply the equivairance of $q_1$ and $q_2$:
\begin{align}
\label{eqn:q1_net_equi}
q_1(gs) &= g q_1(s) \\
\label{eqn:q2_net_equi}
q_2(g_\theta \mathrm{crop}(s,x)) &= \rho_{reg}(g_\theta) q_2(\mathrm{crop}(s,x))
\end{align}
where $g\in \hat{\SE}(2)$ acts on the output of $q_1$ through rotating and translating the $Q$-map, and $g_\theta\in C_n$ acts on the output of $q_2$ by performing a circular shift of the output $Q$ values via the regular representation $\rho_{reg}$.

\begin{figure}
    \centering
    \subfigure[Illustration of Equation~\ref{eqn:q1_net_equi}]{\includegraphics[width=0.23\textwidth]{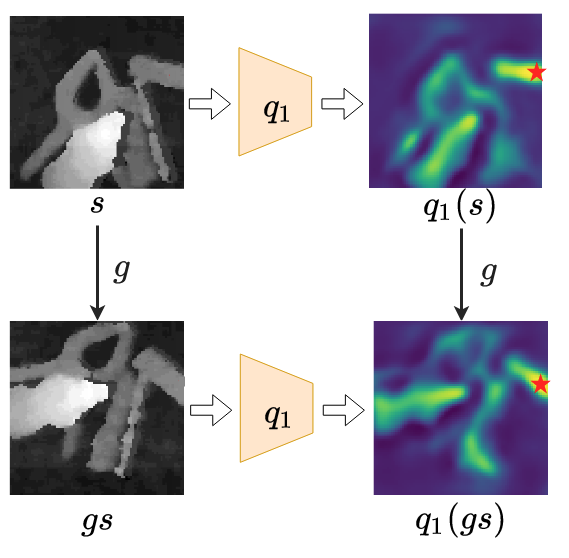}}
    \subfigure[Illustration of Equation~\ref{eqn:q2_net_equi}.]{\includegraphics[width=0.23\textwidth]{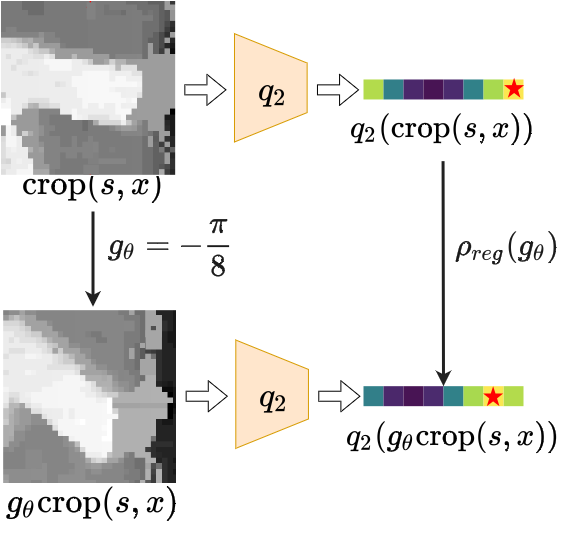}}
    \caption{Equivariance relations expressed by Equation~\ref{eqn:q1_net_equi} and Equation~\ref{eqn:q2_net_equi}.}
    \label{fig:q1_q2_net_equi}
\end{figure}

This is illustrated in Figure~\ref{fig:q1_q2_net_equi}. In Figure~\ref{fig:q1_q2_net_equi}a we take an example of a depth image $s$ in the upper left corner. If we rotate and translate this image by $g$ (lower left of Figure~\ref{fig:q1_q2_net_equi}a) and then evaluate $q_1$, we arrive at $q_1(gs)$. This corresponds to the LHS of Equation~\ref{eqn:q1_net_equi}. However, because $q_1$ is an equivariant function, we can calculate the same result by first evaluating $q_1(s)$ and \emph{then} applying the transformation $g$ (RHS of Equation~\ref{eqn:q1_net_equi}). Figure~\ref{fig:q1_q2_net_equi}b illustrates the same concept for Equation~\ref{eqn:q2_net_equi}. Here, the network takes the image patch $\mathrm{crop}(s,x)$ as input. If we rotate the image patch by $g_\theta$ and then evaluate $q_2$, we obtain the LHS of Equation~\ref{eqn:q2_net_equi}, $q_2(g_\theta \mathrm{crop}(s,x))$. However, because $q_2$ is equivariant, we can obtain the same result by evaluating $q_2(\mathrm{crop}(s,x))$ and circular shifting the resulting vector to denote the change in orientation by one group element.



\subsubsection{Model Architecture of Equivariant $q_1$} 

As a fully convolutional network, $q_1$ inherits the translational equivariance property of standard convolutional layers. The challenge is to encode rotational equivariance so as to satisfy Equation~\ref{eqn:q1_net_equi}. We accomplish this using equivariant convolutional layers that satisfy the equivariance constraint of Equation~\ref{eqn:equi_conv} 
where we assign $\mathcal{F}_{in}=s\in{\mathbb{R}^{1\times m\times h\times w}}$ to encode the input state $s$ and $\mathcal{F}_{out} \in \mathbb{R}^{1\times 1\times h\times w}$ to encode the output $Q$-map. Both feature maps are associated with the trivial representation $\rho_0$ such that the rotation $g$ operates on these feature maps by rotating pixels without changing their values. We use the regular representation $\rho_{reg}$ for the hidden layers of the network to encode more comprehensive information in the intermediate layers. We found we achieved the best results when we defined $q_1$ using the dihedral group $D_4$ which expresses the group generated by rotations of multiples of $\pi/2$ in combination with vertical reflections.





\subsubsection{Model Architecture of Equivariant $q_2$} 

Whereas the equivariance constraint in Equation~\ref{eqn:q1_net_equi} is over $\hat{\SE}(2)$, the constraint in Equation~\ref{eqn:q2_net_equi} is over $C_n$ only. We implement Equation~\ref{eqn:q2_net_equi} using Equation~\ref{eqn:equi_conv} with an input of $\mathcal{F}_{in} = \mathrm{crop}(s, x) \in \mathbb{R}^{1\times m\times h'\times w'}$ as a trivial representation, and an output of $\mathcal{F}_{out} \in \mathbb{R}^{n\times1\times 1\times 1}$ as a regular representation. $q_2$ is defined in terms of the group $C_n$, assuming the rotations in the action space are defined to be multiples of $2\pi/n$.  


\subsubsection{$q_2$ Symmetry Expressed as a Quotient Group} 
\label{sect:quotient}

It turns out that additional symmetries exist when the gripper has a bilateral symmetry. In particular, it is often the case that rotating a grasp pose by $\pi$ radians about its forward axis does not affect the probability of grasp success, i.e.,$r$ is invariant to rotations of the action by $\pi$ radians. When this symmetry is present, we can model it using the quotient group $C_n / C_2 \cong \{ 2 \pi k /n : 0\leq k < n/2, k\in \mathbb{Z}, 0 \equiv \pi \}$ which pairs orientations separated by $\pi$ radians into the same equivalence classes.





\subsection{Other Optimizations}

\label{sec:other_optimizations}

While our use of equivariant models to encode the $Q$-function is responsible for most of our gains in sample efficiency (Section~\ref{sect:ablations}), there are several additional algorithmic details that, taken together, have a meaningful impact on performance.


\subsubsection{Loss Function}
\label{sec:loss_function}

In the standard ASR loss function, given a one step reward $r(s, a)$, where $a=(x, \theta)$, $Q_1$ and $Q_2$ have targets~\citep{asr}:
\begin{align}
    \label{eqn:asr}
    \mathcal{L} &= \mathcal{L}_{1} + \mathcal{L}_{2} \\ 
    \label{eqn:asrl1}
    \mathcal{L}_{1} &= \tfrac{1}{2}(Q_1(s, x) - \max_{\theta'} Q_2(\mathrm{crop}(s, x), \theta'))^2 \\
    \mathcal{L}_{2} &= \tfrac{1}{2}(Q_2(\mathrm{crop}(s, x), \theta) - r(s, a))^2.
\end{align}

In $\mathcal{L}_1$ term, however, since the reward $r(s, a)$ is the ground truth return of $Q_2(\mathrm{crop}(s, x), \theta)$, we correct $Q_2$ with $r(s, a)$. Denoted $\hat{Q}_2$ as the corrected $Q_2$
\begin{equation}
    \hat{Q}_2(\mathrm{crop}(s, x), {\theta'}) = 
    \begin{cases}
        Q_2(\mathrm{crop}(s, x), {\theta'}), & \text{if } {\theta'}\neq\theta\\
        r(s, a), & \text{if } {\theta'}=\theta
    \end{cases}
\end{equation}

We then modify $\mathcal{L}_1$ to learn from $\hat{Q}_2$:

\begin{equation}
\mathcal{L}_1' = \frac{1}{2}\Big(Q_1(s, x) - \max_{\theta'} \left[ \hat{Q}_2(\mathrm{crop}(s, x), {\theta'}) \right] \Big)^2.
\end{equation}

In addition to the above, we add an off-policy loss term $\mathcal{L}_1''$ that is evaluated with respect to an additional $k$ grasp positions $\bar{X} \subset X$ sampled using a Boltzmann distribution from $Q_1(s)$: 
\begin{equation}
\label{eqn:q1_offline}
\begin{split}
\mathcal{L}_1'' = &\frac{1}{2k}\sum_{x_i\in \bar{X}} \Big(Q_1(s, x_i) - \\
&\max_{\theta^{'}} \left[ Q_2(\mathrm{crop}(s, x_i), \theta^{'}) \right] \Big)^2,
\end{split}
\end{equation}
where $Q_2$ provide targets to train $Q_1$. This off-policy loss minimizes the gap between $Q_1$ and $Q_2$. Our combined loss function is therefore $\mathcal{L} = \mathcal{L}_{1}' + \mathcal{L}_{1}'' + \mathcal{L}_{2}$.

\subsubsection{Prioritizing failure experiences in minibatch sampling}

In the contextual bandit setting, we want to avoid the situation where the agent repeats incorrect actions in a row. This can happen because some failure grasps left the scene intact, thus the image and the $Q$-map are intact.
We address this problem by prioritizing failure experiences. When a grasp failed, the experience is included in the sampled minibatch on the next SGD step~\cite{synergy}, thereby updating the $Q$-function prior to reevaluating it on the next time step. The updates in $Q$ reduce the chance of selecting the same (bad) action.




\subsubsection{Boltzmann exploration}

We find empirically Boltzmann exploration is better compares to $\epsilon$-greedy exploration in our grasp setting. We use a temperature of $\tau_\text{training}$ during training and a lower temperature of $\tau_\text{test}$ during testing. Using a non-zero temperature at test time helped reduce the chances of repeatedly sampling a bad action.


\subsubsection{Data augmentation}

Even though we are using equivariant neural networks to encode the $Q$-function, it can still be helpful to perform data augmentation as well. This is because the granularity of the rotation group encoded in $q_1$ ($D_4$) is coarser than that of the action space ($C_n/C_2$). We address this problem by augmenting the data with translations and rotations sampled from $\hat{\SE}(2)$. For each experienced transition, we add eight additional $\hat{\SE}(2)$-transformed images to the replay buffer.

\subsubsection{Softmax at the output of $q_1$ and $q_2$} 

Since we are using a contextual bandit with binary rewards and the reward function $r(s,a)$ denotes the parameter of a Bernoulli distribution at $s,a$, we know that $Q_1$ and $Q_2$ must each take values between zero and one. We encode this prior using an element-wise softmax layer at the output of each of the $q_1$ and $q_2$ networks.


\subsubsection{Selection of the $z$ coordinate} 

In order to execute a grasp, we must calculate a full $x,y,\theta, z$ goal position for the gripper. Since our model only infers a planar grasp pose, we must calculate a depth along the axis orthogonal to this plane (the $z$ axis) using other means. In this paper, we calculate $z$ by taking the average depth over a $5 \times 5$ pixel region centered on the grasp point in the input depth image. The commanded gripper height is set to an offset value from this calculated height. While executing the motion to this height, we monitor force feedback from the arm and halt the motion prematurely if a threshold is exceeded.

\subsection{Optimizations for transparent object grasping using RGB input}
\label{sec:opt_rgb}

In order to grasp transparent objects using our model, we found it was helpful to make a few small modifications to our model and setup. Most importantly, we found it was essential to use RGB rather than depth-only image input to the model.

\noindent
\underline{Bin color:} We found that for transparent objects, our system performed much better using a black bin color rather than a white or transparent bin color. Figure~\ref{fig:robot_setup} illustrates this difference in setup. We believe that the performance difference is due to the larger contrast between the transparent objects and the bin in the RGB spectrum.


\noindent
\underline{Dihedral group in $q2$:} Another optimization we used in our transparent object experiments was to implement Equation~\ref{eqn:q2_net_equi} using dihedral group $D_n$ which expresses the group of multiples of $2\pi/n$ and reflections, rather than $C_n$. As in Section~\ref{sect:quotient}, we use a quotient group that encodes gripper symmetries. Here, this quotient group becomes $D_n/D_2$, which pairs orientations separated by $\pi$ and reflected orientations into the same equivalent class.


\noindent
\underline{Collision penalty:} In our experiments with transparent objects, we found that collision was a more significant problem than it was with opaque objects. We believe this was a result of the fact that since the transparent objects did not completely register in the depth image, standard collision checking between the object point cloud and the gripper did not suffice to prevent collisions. Therefore, in our transparent object experiments, we penalized successful grasps that produced collisions during grasping by awarding those grasps only 0.8 reward instead of the full 1.0 reward.

\section{Experiments in Simulation}
\label{sect:experiments}

\subsection{Setup}
\label{sec:simulation_experiments}

\begin{figure}
    \centering
    \subfigure[86 GraspNet-1B objects used]{\includegraphics[height=0.2\textwidth]{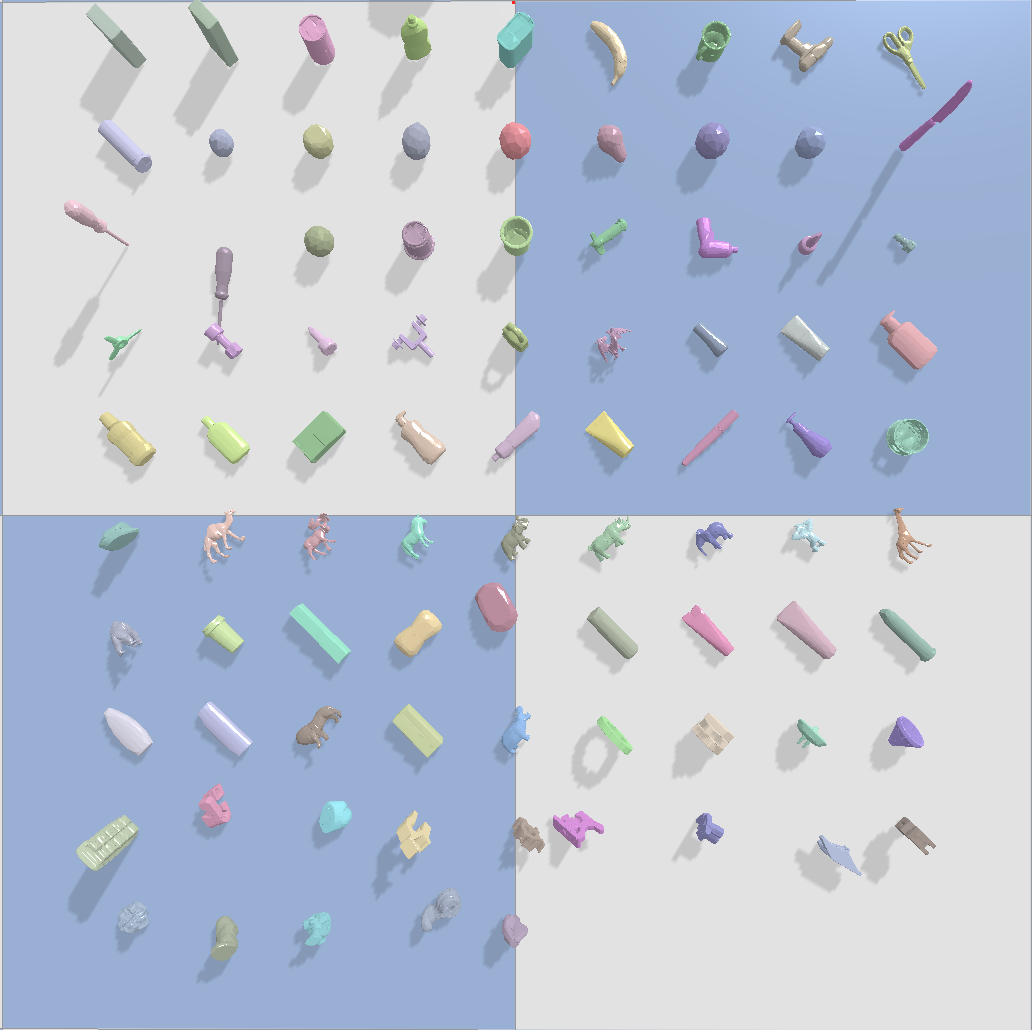}}
    \subfigure[Pybullet simulation]{\includegraphics[height=0.2\textwidth]{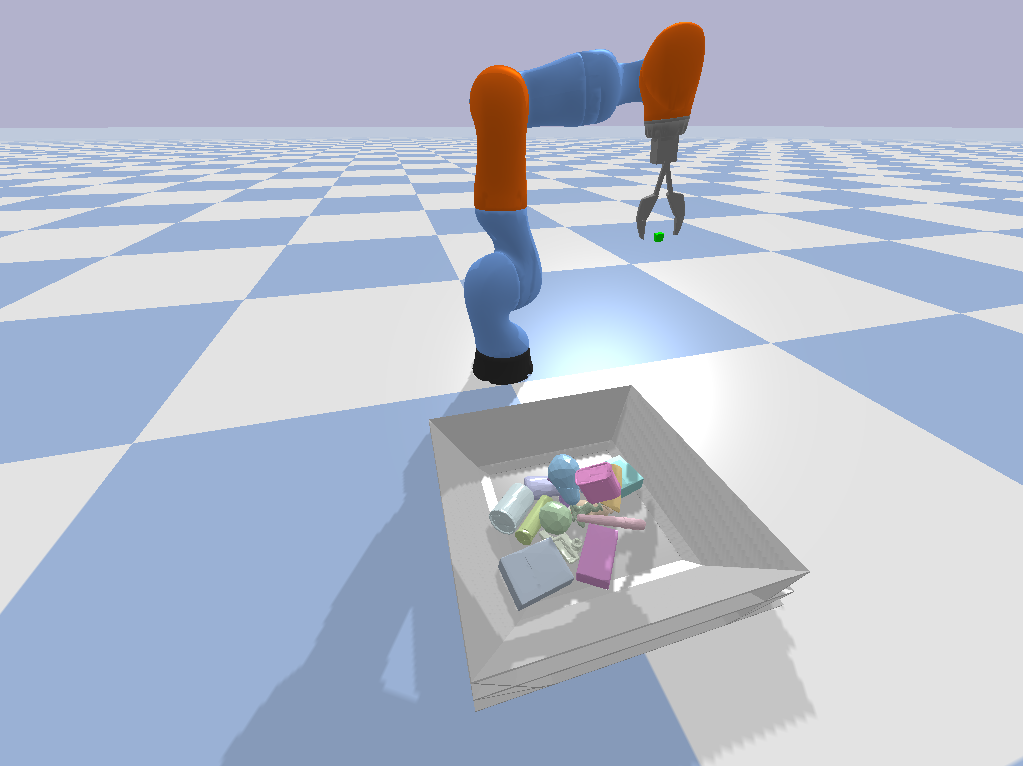}}
    \subfigure[Depth image]{\includegraphics[height=0.22\textwidth]{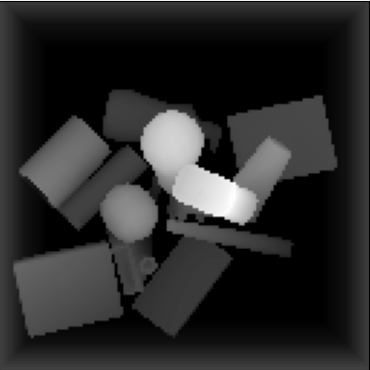}}
    \subfigure[RGB image]{\includegraphics[height=0.22\textwidth]{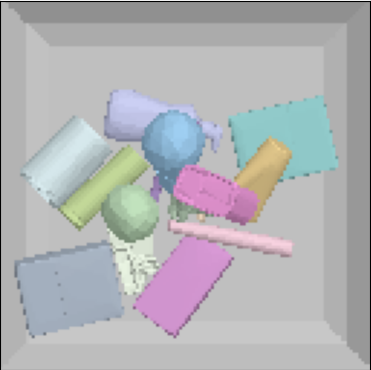}}
    \caption{(a) The 86 objects used in our simulation experiments are drawn from the GraspNet-1Billion dataset~\cite{graspnet-1b}. (b) Phybullet simulation. (c) and (d) are top-down depth and RGB images of the grasping scene.}
    \label{fig:simulation_exps}
\end{figure}


\subsubsection{Object Set}

All simulation experiments are performed using objects drawn from the GraspNet-1Billion dataset~\cite{graspnet-1b}. This includes 32 objects from the YCB dataset~\cite{ycb}, 13 adversarial objects used in DexNet 2.0~\cite{mahler2017dex}, and 43 additional objects unique to GraspNet-1Billion~\cite{graspnet-1b} (a total of 88 objects). Out of these 88 objects, we exclude two bowls because they can be stably placed in non-graspable orientations, i.e., they can be placed upside down and cannot be grasped in that orientation using standard grippers. Also, we scale these objects so that they are graspable from any stable object configuration. Lastly, objects are assigned with random RGB values drawn from uniform distribution $U((0.6, 0.6, 0.6), (1, 1, 1))$. We refer to these 86 mesh models as our simulation ``object set'', shown in Figure~\ref{fig:simulation_exps}a.



\subsubsection{Simulation Details}

Our experiments are performed in Pybullet~\citep{pybullet}. The environment includes a Kuka robot arm and a $0.3\text{m}\times0.3\text{m}$ tray with inclined walls (Figure~\ref{fig:simulation_exps}b). At the beginning of each episode, the environment is initialized with $15$ objects drawn uniformly at random from our object set and dropped into the tray from a height of $40$ cm so that they fall into a random configuration. The state is a depth image (depth modality) or RGB image (RGB modality) captured from a top-down camera (Figure~\ref{fig:simulation_exps}c and d). On each time step, the agent perceives a state and selects an action to execute which specifies the planar pose to which to move the gripper. A grasp is considered to have been successful if the robot is able to lift the object more than $0.1$m above the table. The environment will be reinitialized when all objects have been removed from the tray or $30$ grasp attempts have been made.

\subsection{Comparison Against Baselines on Depth Modality}
\label{sect:baselines}

\subsubsection{Baseline Model Architectures}


We compare our method against two different model architectures from the literature: VPG~\cite{synergy} and FC-GQ-CNN~\citep{4Dof}. Each model is evaluated alone and then with two different data augmentation strategies (soft equ and RAD). In all cases, we use the contextual bandit formulation described in Section~\ref{sect:bandit}. The baseline model architectures are: \underline{VPG:} Architecture used for grasping in~\citep{synergy}. This model is a fully convolutional network (FCN) with a single-channel output. The $Q$ value of different gripper orientations is evaluated by rotating the input image. We ignore the pushing functionality of VPG. \underline{FC-GQ-CNN:} Model architecture used in~\citep{4Dof}. This is an FCN with $8$-channel output that associates each grasp rotation to a channel of the output. During training, our model uses Boltzmann exploration with a temperature of $\tau = 0.01$ while the baselines use $\epsilon$-greedy exploration starting with $\epsilon = 50\%$ and ending with $\epsilon = 10\%$ over 500 grasps (this follows the original implementation in~\cite{synergy}).

\subsubsection{Data Augmentation Strategies}

The data augmentation strategies are: 
\underline{$n\times$ RAD:} The method from~\cite{rad} that augments each sample in the mini-batch with respect to Equation~\ref{eqn:grasp_invariance_assumption}. Specifically, for each SGD step, we first draw $bs$ (where $bs$ is the batch size) samples from the replay buffer. Then for each sample, we augment both the observation and the action using a random $\SE(2)$ transformation, while the reward is unchanged. We perform $n$ SGD steps on the RAD augmented mini-batch after each grasp sample. \underline{$n\times$ soft equ:} similar to $n \times$ RAD except that we produce a mini-batch by drawing $bs/n$ samples, randomly augment those samples $n$ times with respect to equation. \ref{eqn:grasp_invariance_assumption}, then perform a single SGD step. Details can be found in Appendix~\ref{sec:aug_baseline}.






\subsubsection{Results and Discussion}

\begin{figure}
    \centering
    \subfigure[Training]{\includegraphics[width=0.23\textwidth]{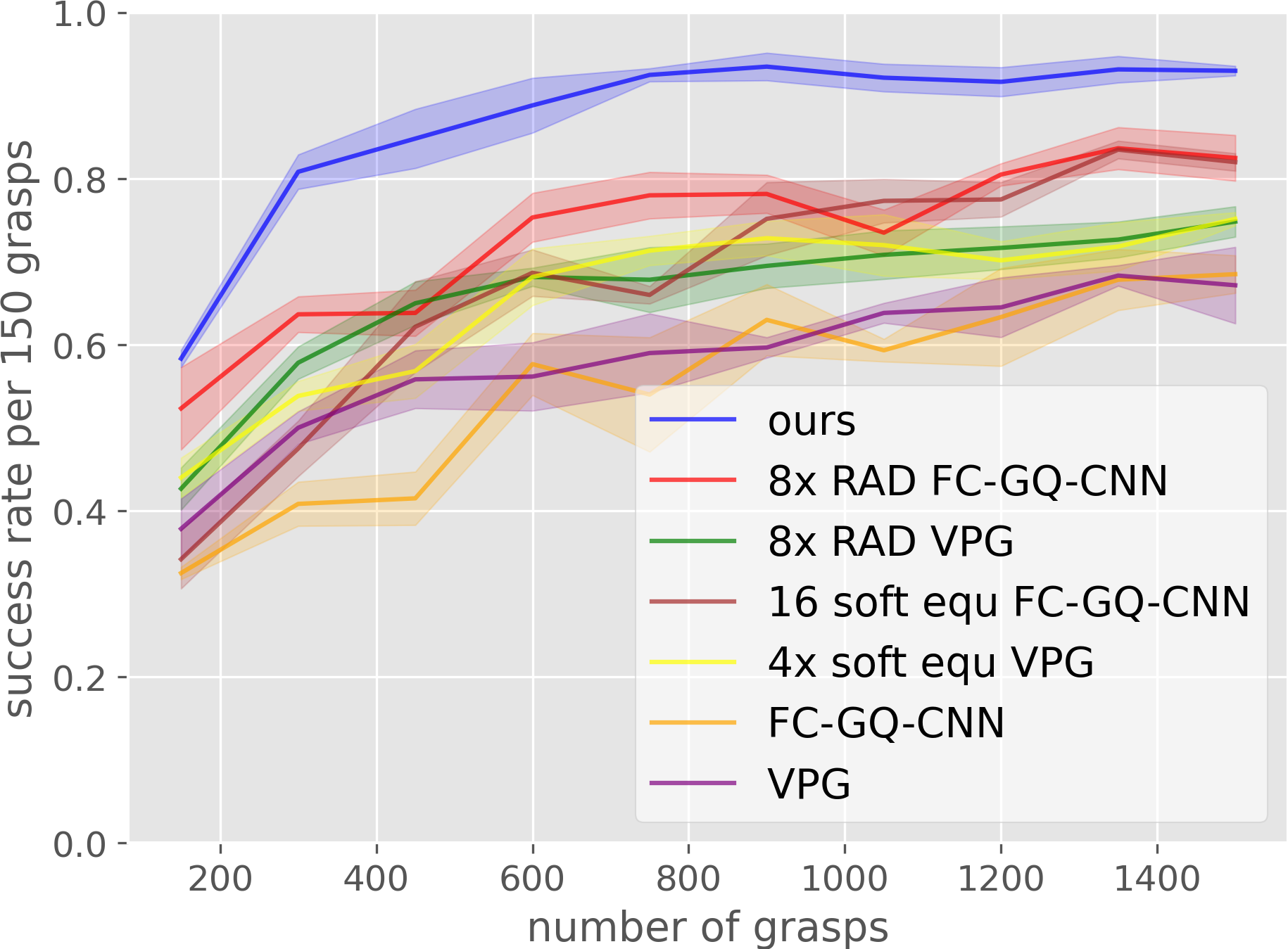}}
    \subfigure[Testing]{\includegraphics[width=0.23\textwidth]{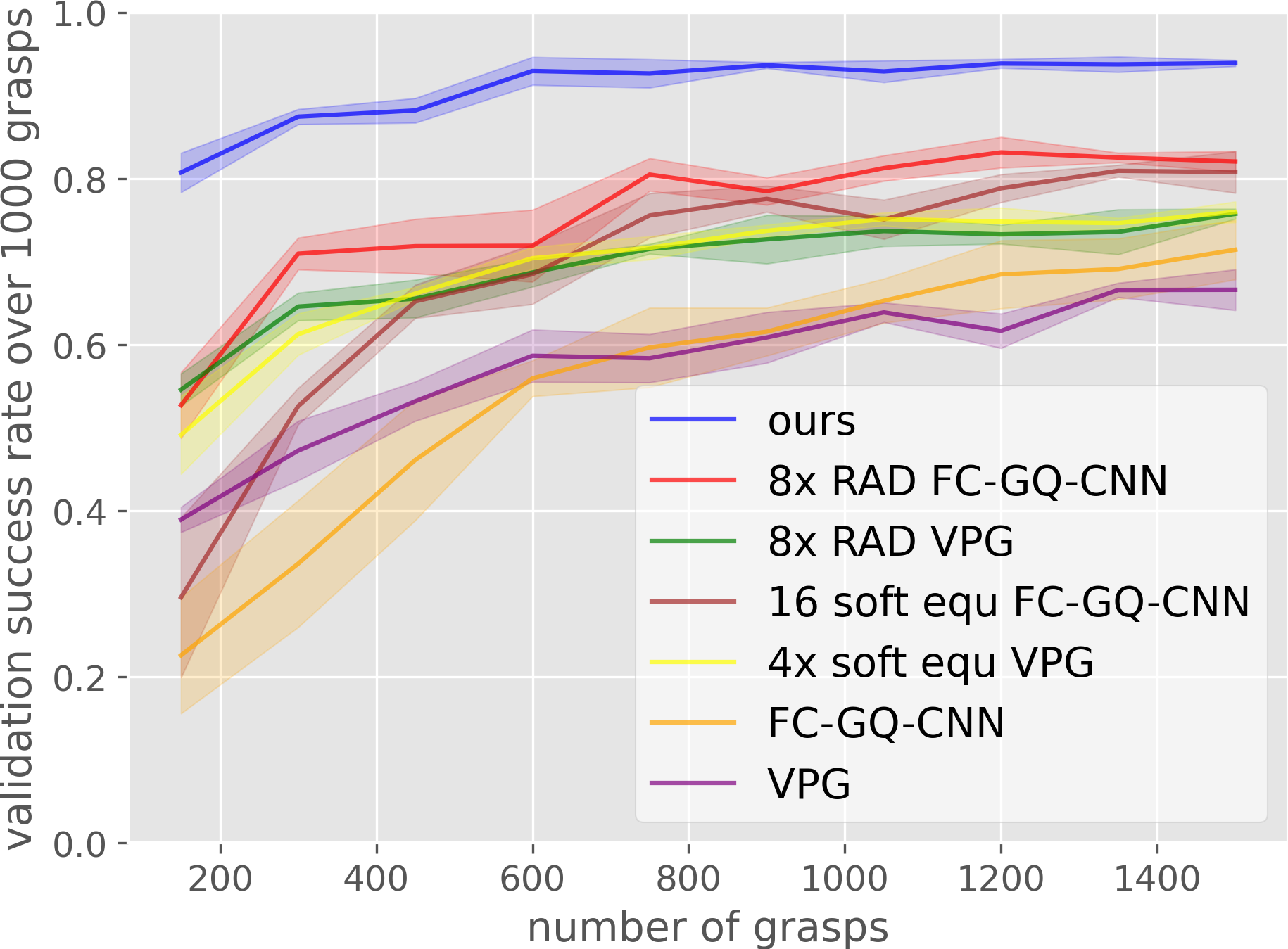}}
    \caption{Comparison with baselines. All lines are an average of four runs. Shading denotes standard error. (a) shows learning curves as a running average over the last 150 training grasps. (b) shows the average near-greedy performance of 1000 validation grasps performed every 150 training steps.}
    \label{fig:simulation_baseline}
\end{figure}

The learning curves of Figure~\ref{fig:simulation_baseline} show the grasp success rate versus the number of grasping attempts on depth modality. Figure~\ref{fig:simulation_baseline}a shows online learning performance. Each data point is the average success rate over the last 150 grasps (therefore, the first data point occurs at 150). Figure~\ref{fig:simulation_baseline}b shows testing performance by stopping training every 150 grasp attempts and performing 1000 test grasps and reporting average performance over these 1000 test grasps. Our method tests at a lower test temperature of $\tau=0.002$ while the baselines test pure greedy behavior. 



Generally, our proposed equivariant model convincingly outperforms the baseline methods and data augmentation strategies with depth modality. In particular, Figure~\ref{fig:simulation_baseline}b shows that the testing success rate of the equivariant model after 150 grasp attempts has the same or better performance than all of the baseline methods after 1500 grasp attempts. Notice that each of the two data augmentation methods we consider (RAD and soft equ) has a positive effect on the baseline methods. However, after training for the full 1500 grasp attempts, our equivariant model converges to the highest grasp success rate ($93.9\pm0.4\%$). Please see Appendix.\ref{sec:add_ablations} for a comparison with longer training horizon.


\subsection{Ablation Study}
\label{sect:ablations}

There are three main parts of SymGrasp as described in this paper: 1) the use of equivariant convolutional layers instead of standard convolution layers; 2) the use of the augmented state representation (ASR) instead of a single network; 3) the various optimizations described in Section~\ref{sec:other_optimizations}. Here, we evaluate the performance of the method when ablating each of these three parts in the depth modality. For additional ablations, see Appendix~\ref{sec:add_ablations}.

\subsubsection{Baselines}

In \underline{no equ}, we replace all equivariant layers with standard convolutional layers. In \underline{no ASR}, we replace the equivariant $q_1$ and $q_2$ models described in Section~\ref{sect:asr} by a single equivariant network. In \underline{no opt}, we remove the optimizations described in Section~\ref{sec:other_optimizations}. In addition to the above, we also evaluated \underline{rot equ} which is the same as \underline{no ASR} except that we replace ASR with a U-net\cite{u_net} and apply $4\times$ RAD\cite{rad} augmentation. Detailed network architectures can be found in Appendix~\ref{sec:network_architecture}.



\subsubsection{Results and Discussion}

Figure~\ref{fig:simulaiton_ablation}a and b shows the results where they are reported exactly in the same manner as in Section~\ref{sect:baselines}. \underline{no equ} does worst, suggesting that our equivariant model is critical. We can improve on this somewhat by adding data augmentation (\underline{rot equ}), but this sill underperforms significantly. The other ablations, \underline{no ASR} and \underline{no opt} demonstrate that those parts of the method are also important.

\subsection{Effect of background color}

This experiment evaluates the effect of background color with RGB modality. 
We compare training with different tray colors which have an offset from the mean RGB value of the objects $(0.8, 0.8, 0.8)$. For example, in \underline{mean$-0.8$}, the tray color would be black $(0, 0, 0)$. We test four different colors from black (mean$-0.8$) to white (mean$+0.2$).

\subsubsection{Results and Discussion}
\begin{figure}
    \centering
    \subfigure[Training]{\includegraphics[width=0.23\textwidth]{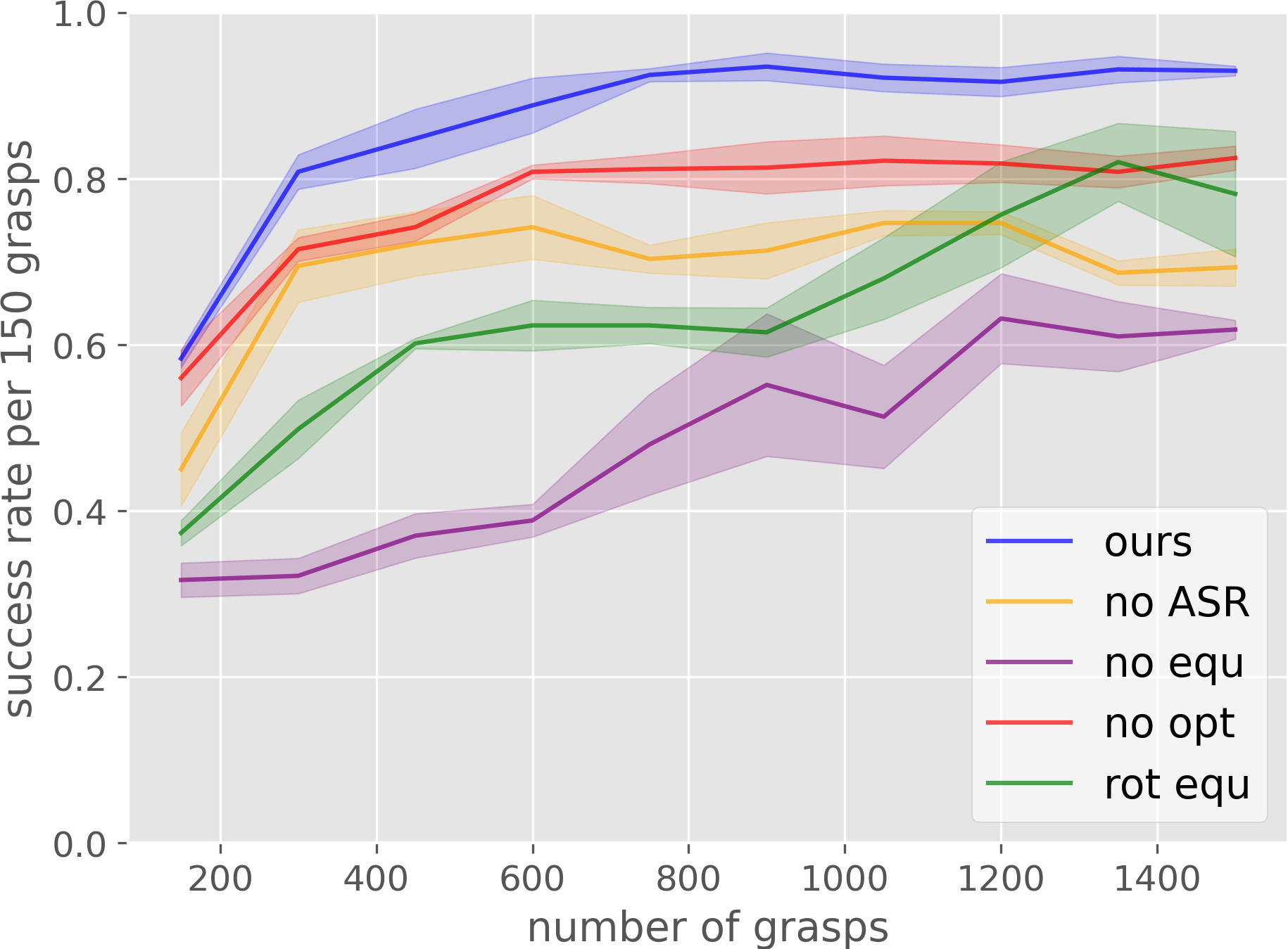}}
    \subfigure[Testing]{\includegraphics[width=0.23\textwidth]{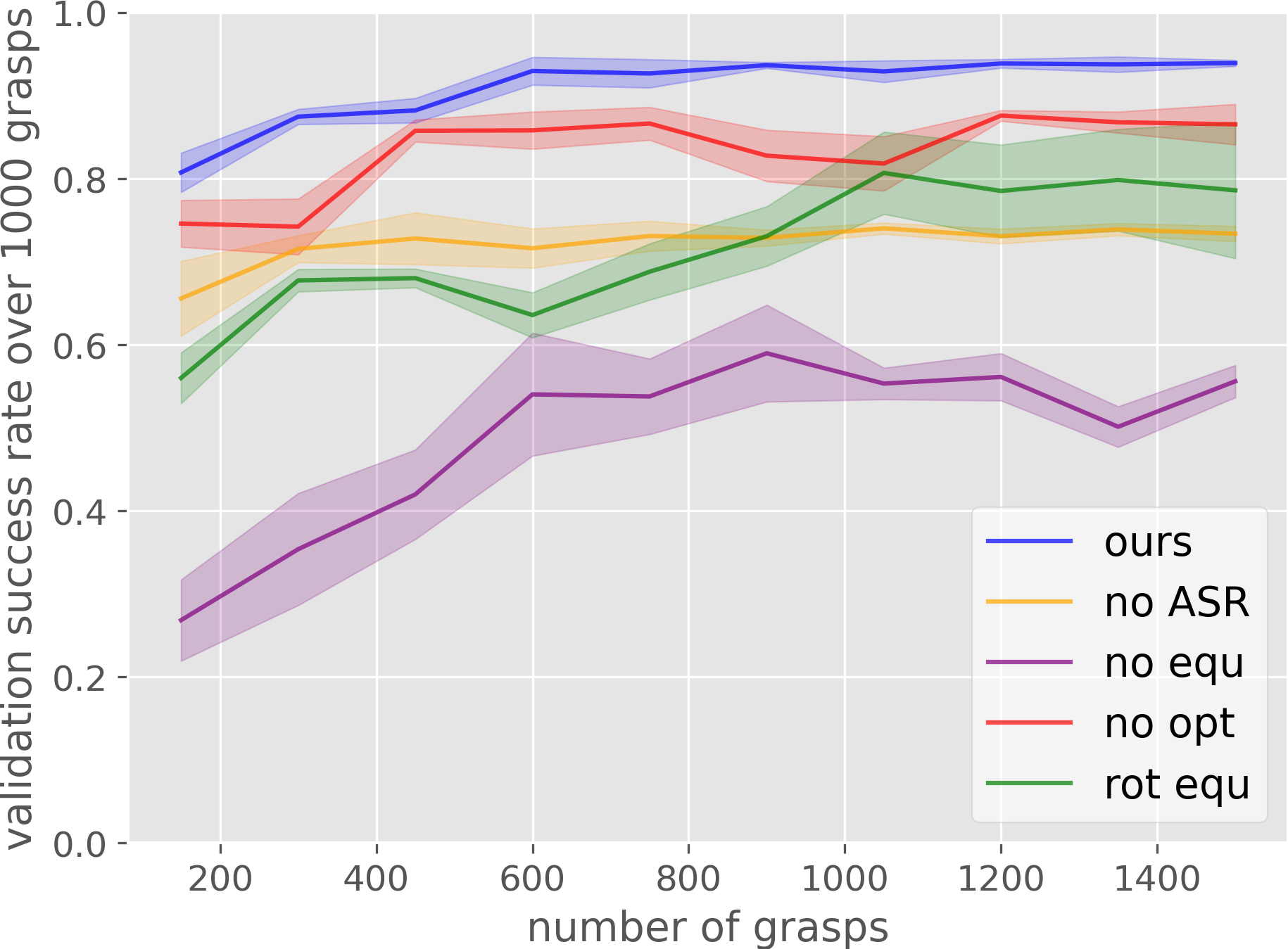}}
    \subfigure[Training]{\includegraphics[width=0.23\textwidth]{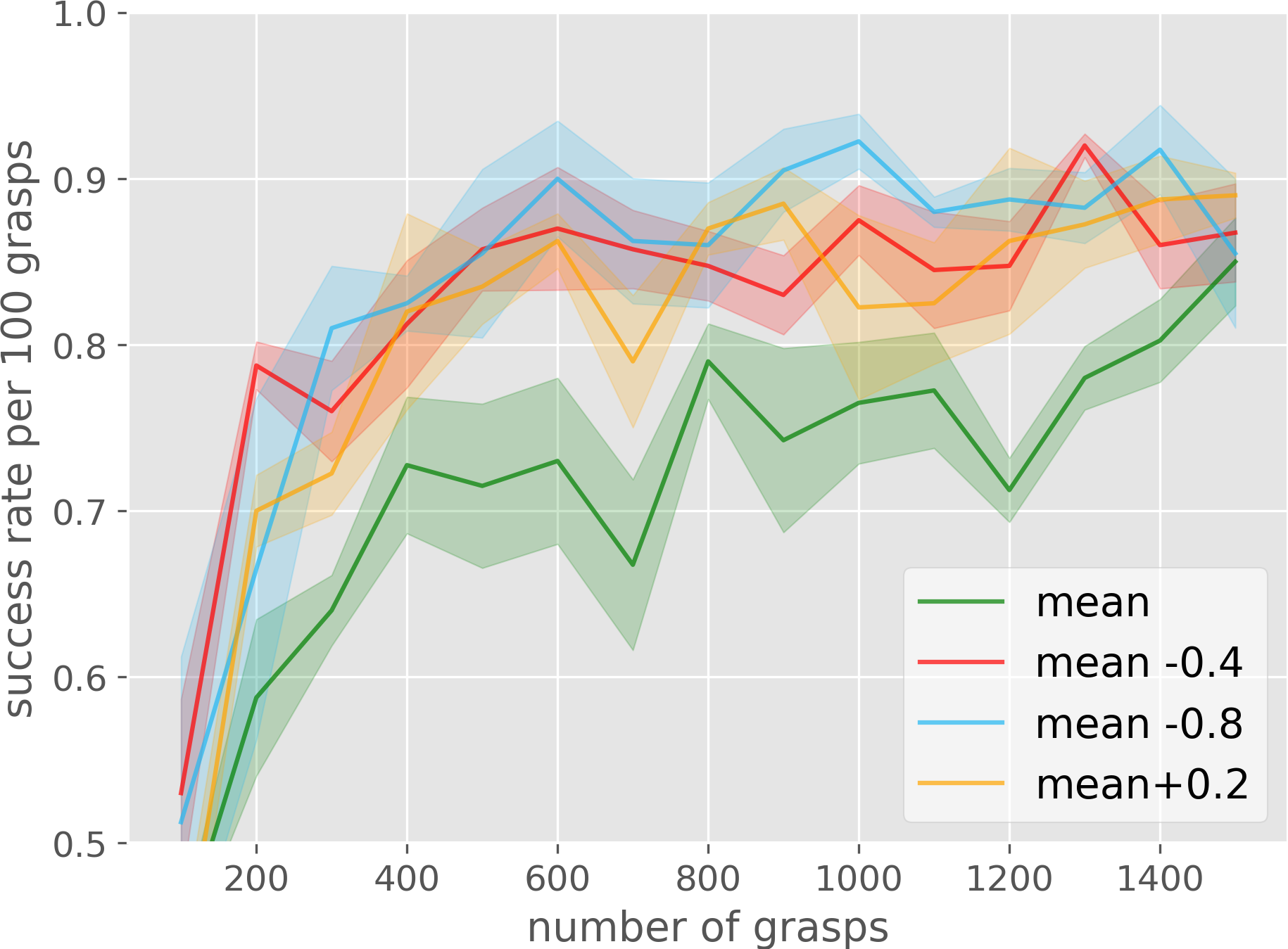}}
    \subfigure[Testing]{\includegraphics[width=0.23\textwidth]{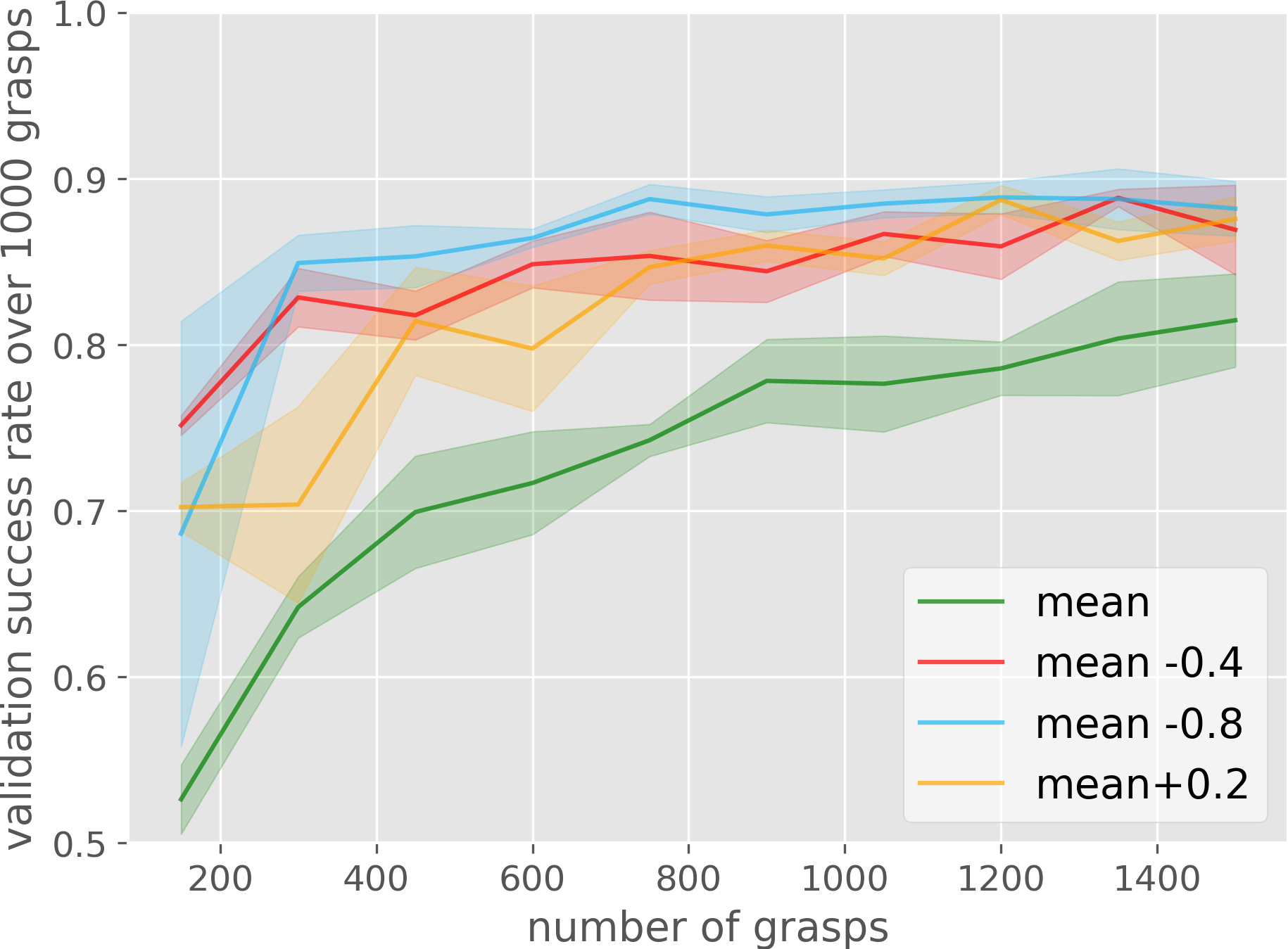}}
    \caption{Ablation study for depth observation. Lines are an average over $4$ runs. Shading denotes standard error. In the left column, learning curves as a running average over the last 150 training grasps. In the right column is the average near-greedy performance of 1000 validation grasps performed every 150 training steps. The first row is in the depth modality while the second row is in the RGB modality.}
    \label{fig:simulaiton_ablation}
\end{figure}

Figure~\ref{fig:simulaiton_ablation}c and d show that the contrast between background and object color has a big impact on grasp learning. In particular, the model has the worst performance when the background color is the same as the mean of the object color (\underline{mean}). The model performs best when the background color has the most significant contrast from the mean of the object color (\underline{mean $-0.8$}).



\begin{figure}
    \centering
    \subfigure[Opaque objects grasping]{\includegraphics[width=0.2\textwidth]{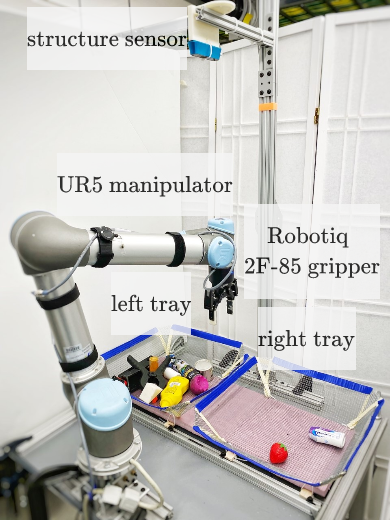}}
    \subfigure[Transparent objects grasping]{\includegraphics[width=0.21\textwidth]{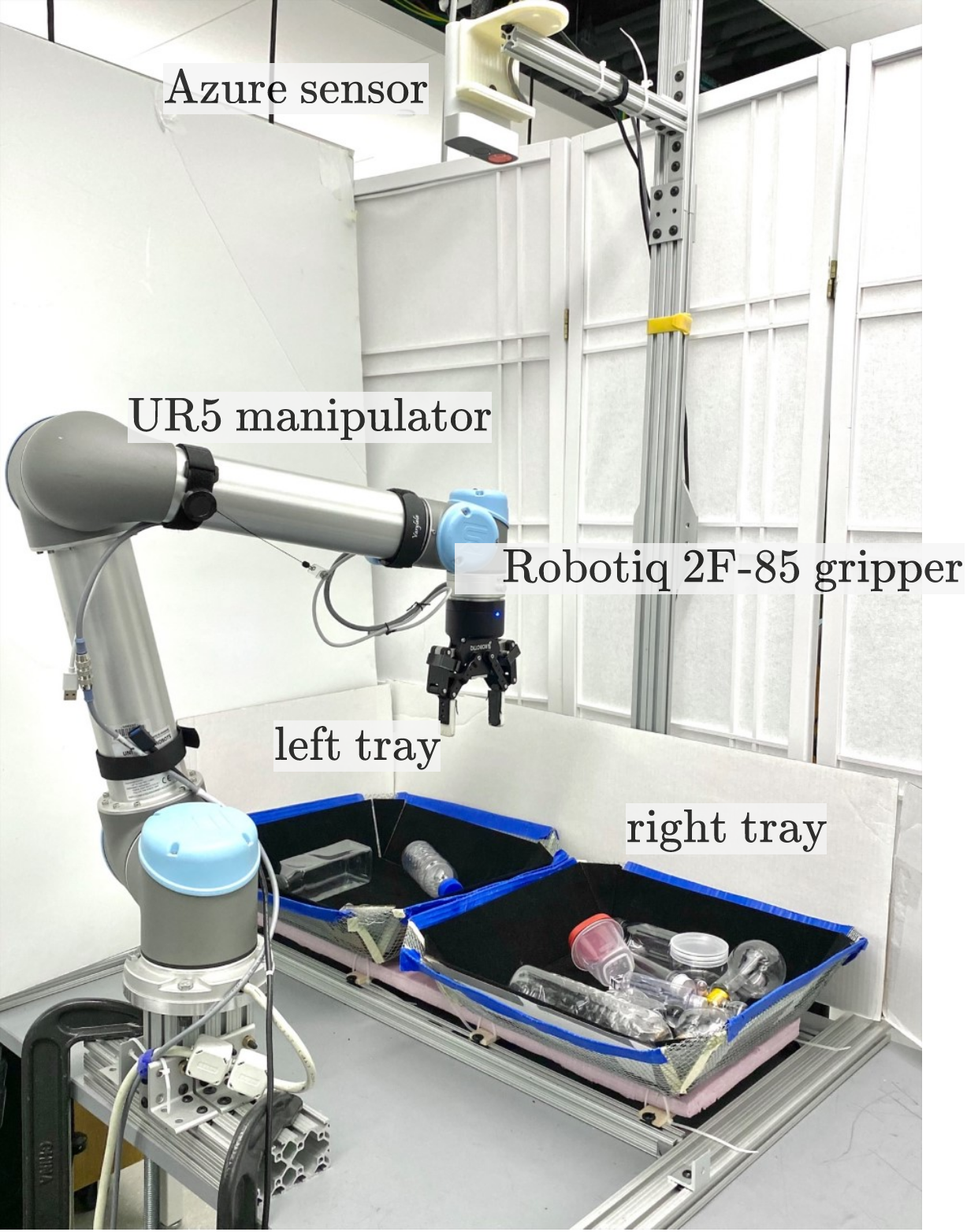}}
    \caption{Setup for self-supervised training on the robot.}
    \label{fig:robot_setup}
\end{figure}

\section{Experiments in Hardware}
\label{sec:physical_robot_experiments}

\subsection{Setup}

\subsubsection{Robot Environment}

Our experimental platform is comprised of a Universal Robots UR5 manipulator equipped with a Robotiq 2F-85 parallel-jaw gripper. For depth modality, we use an Occipital Structure Sensor and  for RGB modality we use a Kinect Azure Sensor paired with two black trays. The dual-tray grasping environment is shown in Figure~\ref{fig:robot_setup}. The workstation is equipped with an Intel Core i7-7800X CPU and an NVIDIA GeForce GTX 1080 GPU.

\subsubsection{Self-Supervised Training}

The training begins with the 15 training objects (Figure~\ref{fig:obj_sets}a for opaque object grasping and Figure~\ref{fig:transparent_obj_sets}a for transparent object grasping) being dropped into one of the two trays by the human operator. Then, the robot attempts to pick objects from one tray and drop them on another. The drop location is sampled from a Gaussian distribution centered in the middle of the receiving tray. All grasp attempts are generated by the contextual bandit. When all 15 objects have been transported in this way, training switches to attempting to grasp from the other tray and drop into the first. Whether all objects are transported or not is decided by the heuristic. For opaque objects, we threshold the depth sensor reading. For transparent objects, we compare the RGB image of the current scene with an RGB image of an empty tray, similar to \cite{qt-opt}. During training, the robot performs 600 grasp attempts in this way (that is 600 grasp \emph{attempts}, not 600 successful grasps). A reward $r$ will be set to $1$ if the gripper was blocked by the grasped object, otherwise $r=0$.



\subsubsection{In-Motion Computation}

We are able to nearly double the speed of robot training by doing all image processing and model learning while the robotic arm was in motion. This was implemented in Python as a producer-consumer process using mutexs. As a result, our robot is constantly in motion during training and the training speed for our equivariant algorithm is limited by the velocity of robot motion. This improvement enabled us to increase robot training speed from approximately 230 grasps per hour to roughly 400 grasps per hour.

\subsubsection{Evaluation procedure}

The evaluation begins with the human operator dropping objects into one tray, after this, no human interference is allowed. We evaluate the success rate of robot grasping objects. A key failure mode during testing is repeated failure grasps. To combat this, we use the procedure of~\cite{synergy} to reduce the chances of repeated grasp failures.
The procedure is that after a grasp failure, we perform multiple SGD steps using that experience to ``discourage'' the model from selecting the same action and then use that updated model for the subsequent grasp. After a successful grasp, we discard these updates and reload the original network.

 All runs are evaluated by freezing the corresponding model and executing 100 greedy (or near greedy) test grasps for each object set in the easy-object test set (Figure~\ref{fig:obj_sets}b), the hard-object test set (Figure~\ref{fig:obj_sets}c).



\subsubsection{Model Details}

For all methods, prior to training on the robot, model weights are initialized randomly using an independent seed. No experiences from simulation are used, i.e.,we train from scratch.
For our depth algorithm, the $q_1$ network is defined using $D_4$-equivariant layers and the $q_2$ network is defined using $C_{16}/C_2$-equivariant layers. For ours RGB algorithm, the $q_2$ network is defined using $D_{16}/D_2$-equivariant layers. During training, we use Boltzmann exploration with a temperature of $0.01$. During testing, the temperature is reduced to $0.002$ (near-greedy). For more details, see Appendix~\ref{sec:evaluation_details}.




\begin{table}
 \centering
    \begin{tabular}{ m{6em} m{5em} m{5em} m{3em}}
        \toprule
        Baseline &test set easy & test set hard & $t_\textit{SGD}$ \\
        \midrule
        $8\times$ RAD VPG &  61.8 $\pm$3.59 & 52.5 $\pm$5.33 & 12.1s\\
        $8\times$ RAD FC-GQ-CNN &  82.8 $\pm$1.65 & 74.3 $\pm$3.04 & 1.55s \\
        Ours & 95.0 $\pm$1.47 & 87.0 $\pm$1.87 & 1.11s\\
        \bottomrule
    \end{tabular}
    \caption{Evaluation success rate ($\%$), standard error, and training time per grasp $t_\textit{SGD}$ (in seconds) in the hardware experiments. Results are an average of 100 grasps per training run averaged over four runs, performed on the held out test objects shown in Figure~\ref{fig:obj_sets}b and c.}
    \label{tab:opaque_test_results}
\end{table}

\begin{table}
 \centering
    \begin{tabular}{ m{5em} m{6em} m{5.5em} m{3em}}
        \toprule
        Baseline & trans. train set& trans. test set & $t_\text{capture}$ \\
        \midrule
        Ours RGB & 85.3 $\pm$8.62 & 83.3 $\pm$3.20  & 1.8s\\
        \bottomrule
    \end{tabular}
    \caption{Evaluation success rate ($\%$), standard error, and capture time per grasp (in seconds) in the hardware experiments of transparent objects. Results are an average of 100 grasps per training run averaged over four runs, performed on the train and held out test objects shown in Figure~\ref{fig:transparent_obj_sets} a and b.}
    \label{tab:transparent_test_results}
\end{table}


\subsection{Opaque object grasping}

\subsubsection{Objects}

\begin{figure}
    \centering
    \subfigure[Training set]{\includegraphics[width=0.145\textwidth]{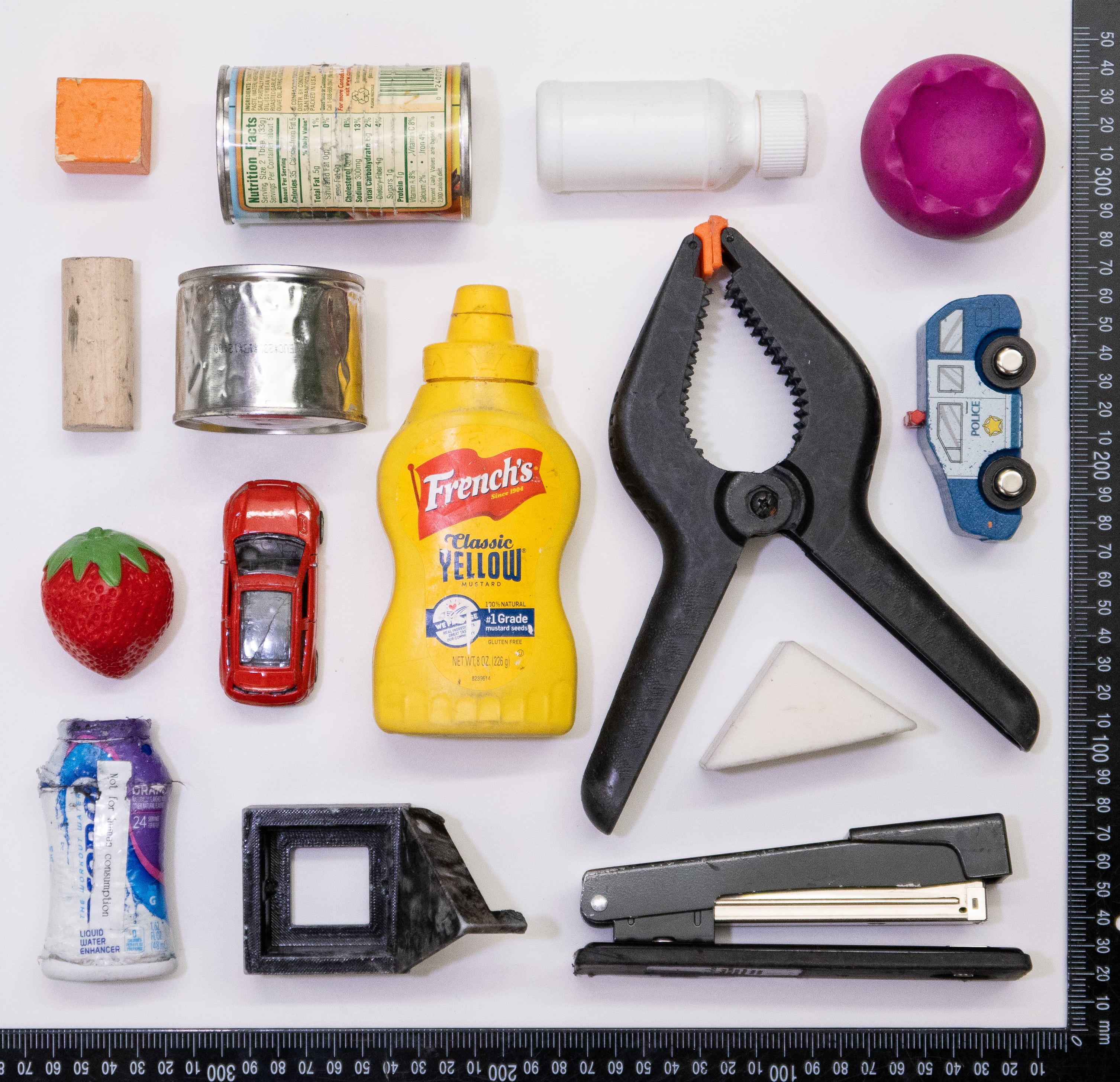}}
    \subfigure[Testing set, easy]{\includegraphics[width=0.142\textwidth]{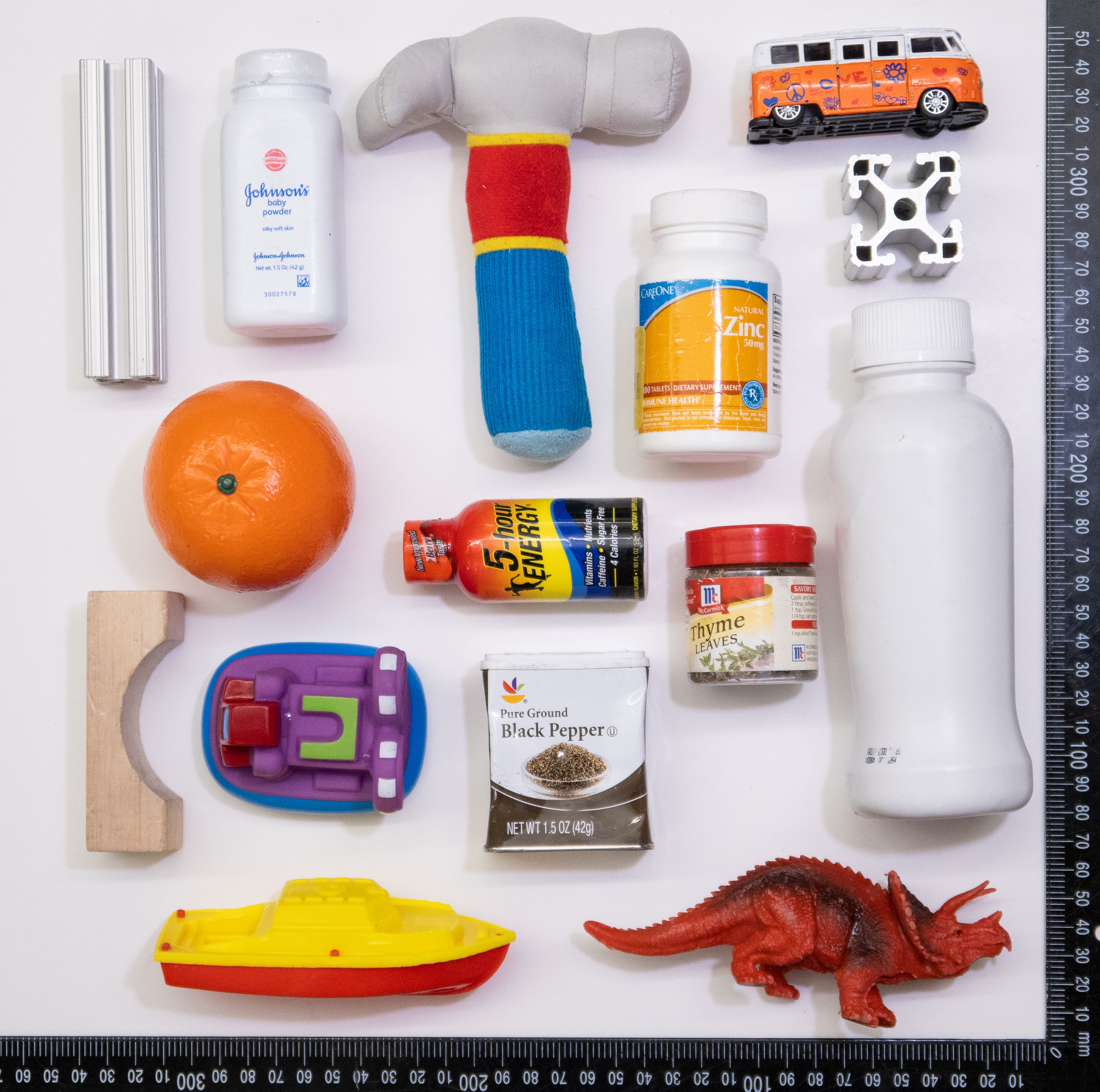}}
    \subfigure[Testing set, hard]{\includegraphics[width=0.177\textwidth]{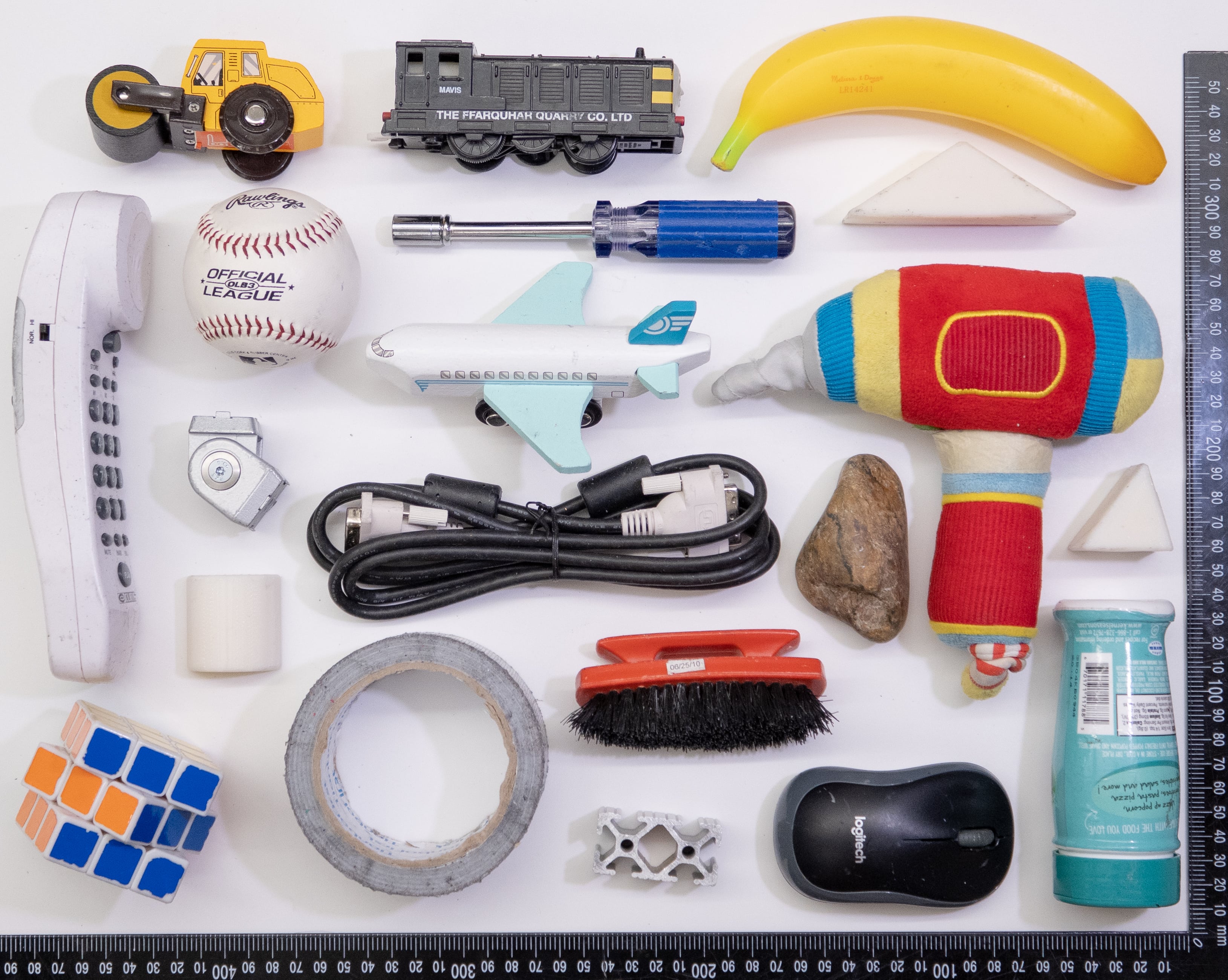}}
    \caption{Object sets used for training and testing. Both training and the test set easy include 15 objects while the test set hard has 20 objects. Objects were curated so that they were graspable by the Robotiq 2F-85 parallel jaw gripper from any configuration and visible to the Occipital Structure Sensor.}
    \label{fig:obj_sets}
\end{figure}

For opaque objects grasping, all training happens using the 15 objects shown in Figure~\ref{fig:obj_sets}a. After training, we evaluate grasp performance on both the ``easy'' test objects (Figure~\ref{fig:obj_sets}b) and the ``hard'' test objects (Figure~\ref{fig:obj_sets}c). Note that both test sets are novel with respect to the training set.

\subsubsection{Baselines}

In our robot experiments for opaque object grasping, we compare our method against \underline{$8\times$ RAD VPG}~\citep{synergy}~\cite{rad} and \underline{$8\times$ RAD FC-GQ-CNN}~\citep{4Dof}~\cite{rad}, the two baselines we found to perform best in simulation. As before, \underline{$8\times$ RAD VPG}, uses a fully convolutional network (FCN) with a single output channel. The $Q$-map for each gripper orientation is calculated by rotating the input image. After each grasp, we perform $8\times$ RAD data augmentation (8 optimization steps with a mini-batch containing randomly translated and rotated image data). \underline{$8\times$ RAD FC-GQ-CNN} also has an FCN backbone, but with eight output channels corresponding to each gripper orientation. It uses $8\times$ RAD data augmentation as well. All exploration is the same as it was in simulation except that the $\epsilon$-greedy schedule goes from $50\%$ to $10\%$ over 200 steps rather than over 500 steps.



\subsubsection{Results and Discussion}

\begin{figure}
    \centering
    \subfigure[Opaque object grasping]{\includegraphics[width=0.23\textwidth]{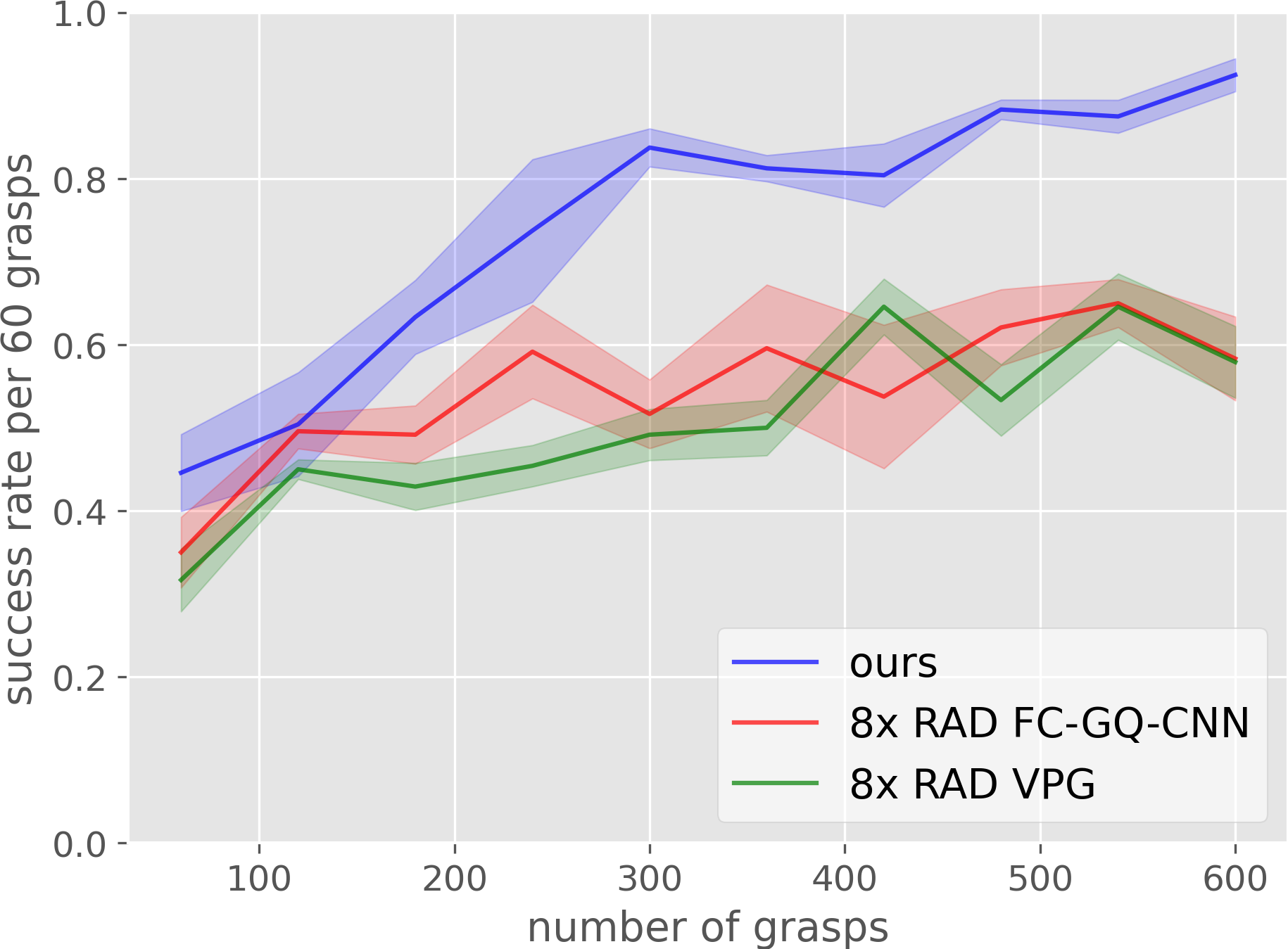}}
    \subfigure[Transparent object grasping]{\includegraphics[width=0.23\textwidth]{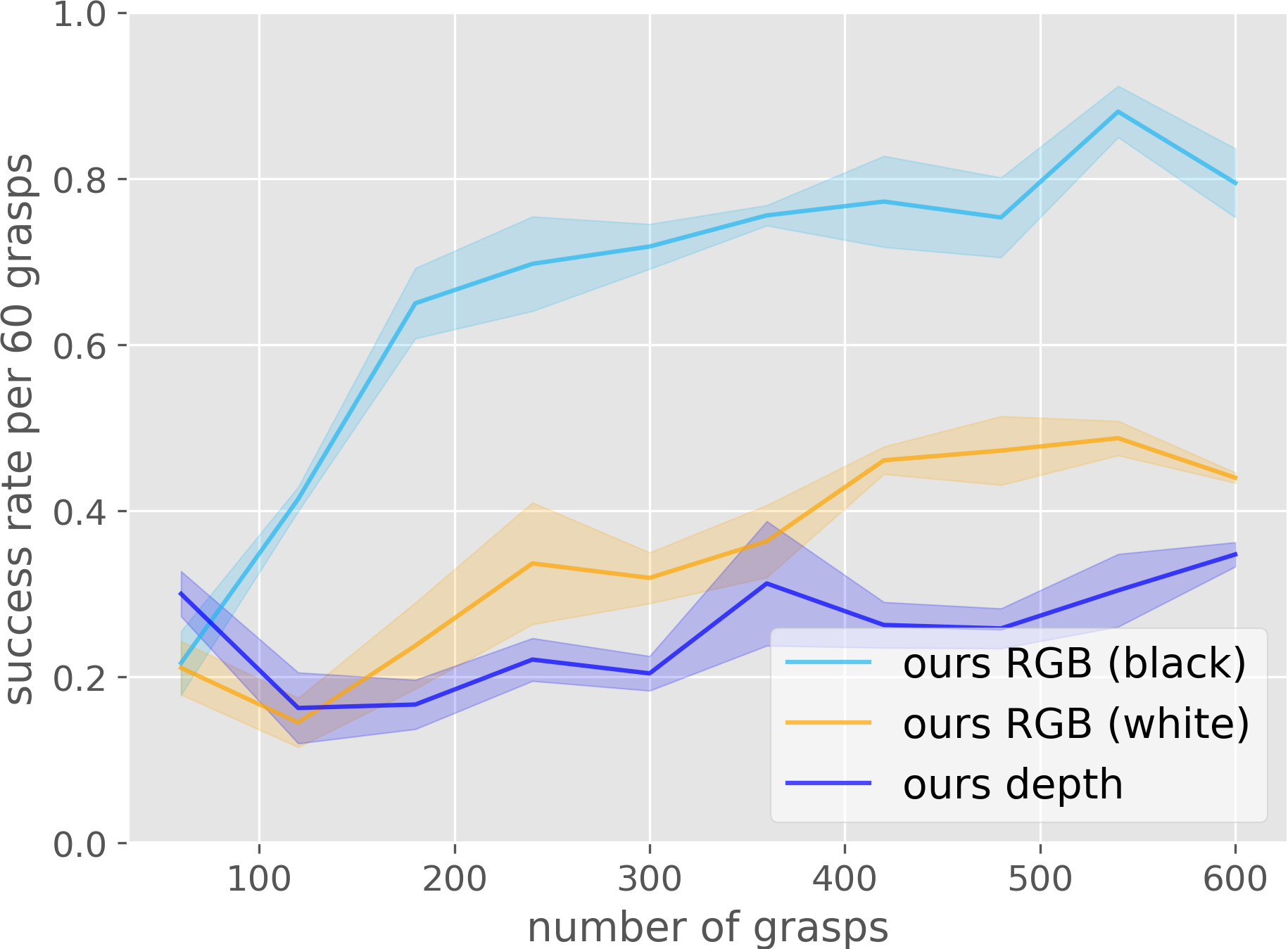}}
    \caption{Learning curves for (a) the opaque object grasping and (b) the transparent object grasping hardware experiment, the parenthesis indicates the color of trays. All curves are averaged over $4$ runs with different random seeds and random object placement. Each data point in the curve is the average grasp success over the last 60 grasp attempts. Shading denotes standard error.}
    \label{fig:physical_experiments}
\end{figure}

\begin{figure}
    \centering
    \subfigure[Depth observation]{\includegraphics[width=0.23\textwidth]{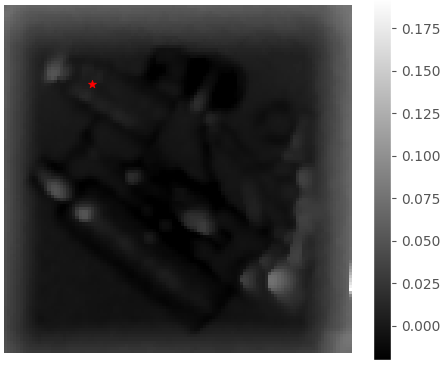}}
    \subfigure[RGB observation]{\includegraphics[width=0.225\textwidth]{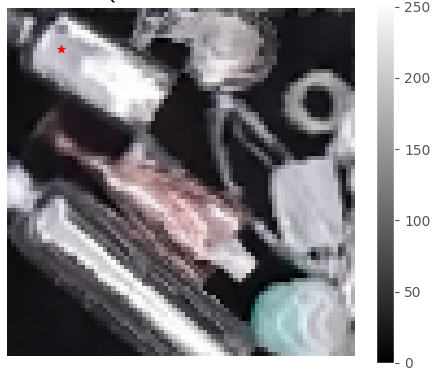}}
    \subfigure[$Q_1$ from RGB observation]{\includegraphics[width=0.22\textwidth]{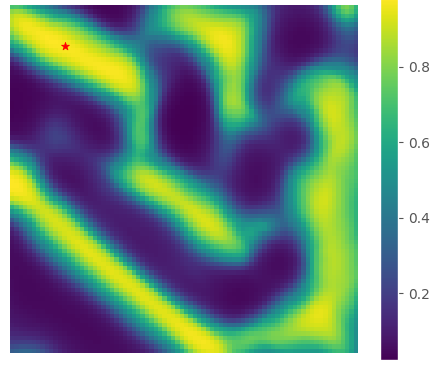}}
    \hspace{8pt}
    \subfigure[Execute the grasp]
    {\includegraphics[width=0.19\textwidth]{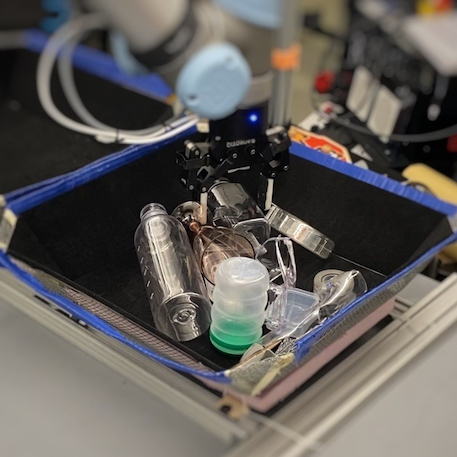}}
    \hspace{10pt}
    \caption{Illustrations for transparent object grasping. (a) shows the depth observation, notice that most transparent objects are not sensed. (b) shows the same scene with RGB modality, notice that objects are sensed up to orthographic projection error. (c) shows the $Q_1$ map from the observation in (b) and the selected action in the red dot. (d) shows the executed grasp.}
    \label{fig:trans_obj_grasp}
\end{figure}



Figure~\ref{fig:physical_experiments}a shows the learning curves on opaque object grasping for the three methods during learning. An important observation is that the results from training on the physical robot Figure~\ref{fig:physical_experiments}a match the simulation training results Figure~\ref{fig:simulation_baseline}a. Figure~\ref{fig:physical_experiments}a shows for opaque object grasping, our method achieves a success rate of $>90\%$ after 600 grasp attempts while the baselines are near $70\%$. Table~\ref{tab:opaque_test_results} shows the testing performance and also demonstrates our method significantly outperforms the baselines during testing. However, performance is lower on the ``hard'' test set. We hypothesize that the lower ``hard'' set performance is due to a lack of sufficient diversity in the training set. The other observation is that since each of these 600-grasp training runs takes approximately 1.5 hours, suggests these methods could efficiently directly learn from real-world data thus avoiding the simulation-to-real-world gap and adapting to physical changes in the robot.

\subsection{Transparent object grasping}

\subsubsection{Objects}

\begin{figure}
    \centering
    \subfigure[Trans. training set]{\includegraphics[width=0.192\textwidth]{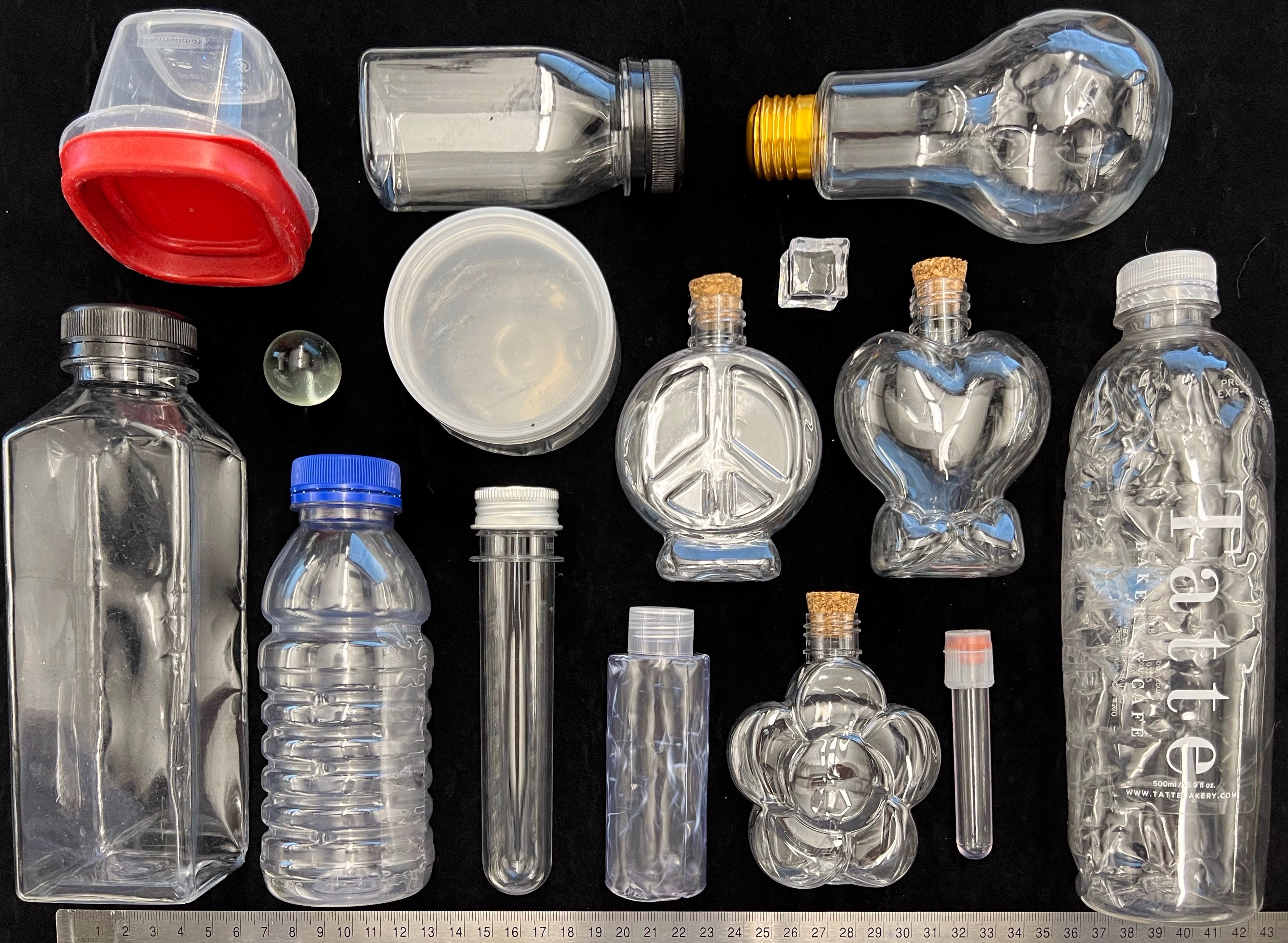}}\hspace{15pt}
    \subfigure[Trans. testing set]{\includegraphics[width=0.149\textwidth]{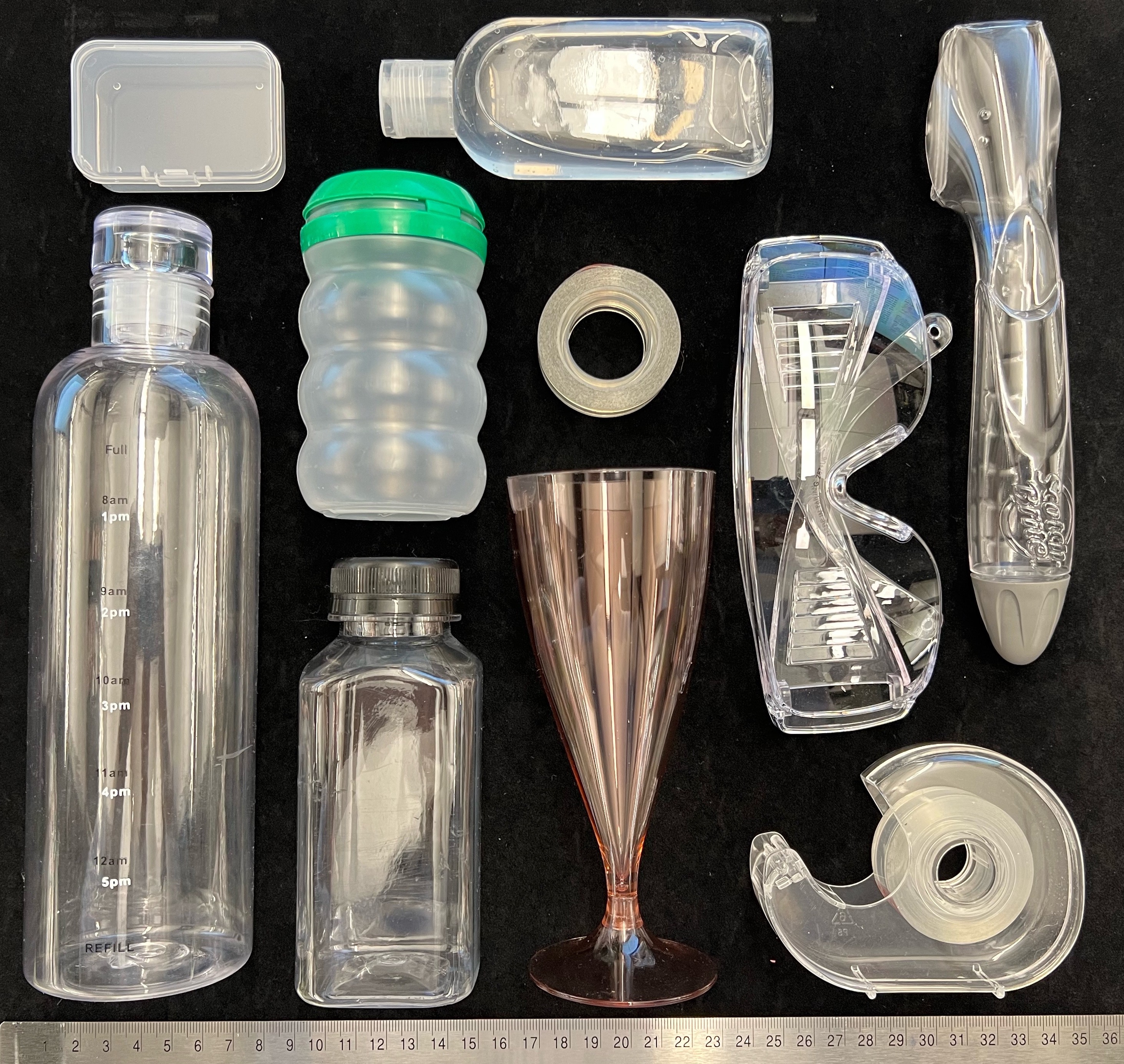}}
    \caption{Transparent object sets used for training and testing. The Training set includes 15 objects and the testing set includes 10 objects. Objects were curated so that they were graspable by the Robotiq 2F-85 parallel jaw gripper from any configuration.}
    \label{fig:transparent_obj_sets}
\end{figure}

For transparent object grasping, all training happens using the 15 objects shown in Figure~\ref{fig:transparent_obj_sets}a. After training, we evaluate grasp performance on the in-distribution training set and out-of-distribution testing objects (Figure~\ref{fig:transparent_obj_sets}a, b). Note that the test set is novel with respect to the training set.

\subsubsection{RGB Modality Details}

The top-down RGB images are orthographic projections of noisy RGB point clouds. RGB values are normalized to fit a Gaussian distribution $\mathcal{N}(0, 0.01)$. During training, each mini-batch of images is augmented in brightness with a $(0.9, 1.1)$ range.

\subsection{Baselines}

For transparent object grasping, we compare \underline{Ours RGB (black)}, \underline{Ours RGB (white)}, and \underline{Ours D}. \underline{Ours RGB (black)} is the baseline in our proposed methods, \underline{Ours RGB (white)} changes black trays to white trays, see figure \ref{fig:tray_color}. \underline{Ours D} is the best baseline in physical opaque objects grasping experiments, here we train and test on transparent objects.

\subsection{Results and Discussion}

Figure~\ref{fig:physical_experiments}(b) shows the learning curves for transparent object grasping and Table~\ref{tab:transparent_test_results} shows the grasping performance of \underline{ours RGB} (black) on the transparent object set. Figure~\ref{fig:trans_obj_grasp} shows a transparent object grasping process during evaluation.


Figure~\ref{fig:physical_experiments}b shows that for transparent object grasping, \underline{ours RGB} (black) produces a significant ($>40\%$) improvement in success rate compared to \underline{ours depth}, suggesting that the RGB information is needed to perform well in this setting.
The background color is also important: changing the tray from black to white leads to a $>30\%$ decrement in success rate. Table~\ref{tab:transparent_test_results} summarizes that \underline{ours RGB} achieves $> 80\%$ success rate on both the training set and testing set, indicating the method learns a good grasp function of in-distribution object set and generalizes to novel transparent objects. Also, notice that from a computational perspective, our method is significantly cheaper ($1.8$ seconds) than NeRF based approaches to transparent object grasping like EVO Nerf~\cite{kerrevo} and Dex Nerf~\cite{ichnowski2021dex} which take $9.5$ second and $16$ seconds, respectively.


\section{Conclusions and Limitations}

Our main contribution is to recognize that planar grasp detection is $\SE(2)$-equivariant and to leverage this structure using an $\SO(2)$-equivariant model architecture.
This model is significantly more sample efficient than other grasp-learning methods and can learn a good grasp function with less than $600$ grasp samples. This increase in sample efficiency enables us to learn to grasp on a physical robotic system in a practical amount of time (approximately and hour and a half) without relying on any sort of pretraining or semi-supervised learning.
Training completely on physical systems is important for two main reasons. First, it eliminates the
need to train in simulation, thereby avoiding the sim2real gap. Second, it enables our system to adapt to changes or idiosyncrasies of the robot hardware or the physical environment that would be hard to simulate.


A key limitation in both simulation and the real world is that despite the fast learning rate, grasp success rates (after training) still seems to be limited to the low/mid $90\%$ for opaque object and $80\%$ for transparent objects. This is the same success rate seen in with other grasp detection methods~\cite{mahler2017dex,tenpas_ijrr2017,mousavian20196}, but it is disappointing here because one might expect faster adaptation to lead ultimately to better grasp performance. This could simply be an indication of the complexity of the grasp function to be learned or it could be a result of stochasticity in the simulator and on the real robot. Another limitation of our work is that it is limited to an open-loop control setting where the model infers only a goal pose. Nevertheless, we expect that the equivariant model structure leverage in this paper would also be useful in the closed loop setting, as suggested by the results of~\cite{wang2022equivariant}.

\section{Declarations}
 
\textbf{Ethical Approval} Not applicable.
 
\textbf{Competing interests} The authors declare that they have no relevant financial or non-financial interests to disclose
 
\textbf{Authors' contributions} Xupeng Zhu proposed the method in discussions with Dian Wang. Xupeng Zhu and Guanang Su conducted the robot experiment. All authors wrote and reviewed the manuscript.
 
\textbf{Funding} This work was supported in part by NSF 1724257, NSF 1724191, NSF 1763878, NSF 1750649, NASA 80NSSC19K1474, the Roux Institute, the Harold Alfond Foundation, NSF grants 2107256, and 2134178.
 
\textbf{Availability of data and materials} Code is available at \url{https://github.com/ZXP-S-works/SE2-equivariant-grasp-learning}.


\clearpage

\bibliographystyle{plainnat}
\bibliography{main}

\clearpage

\begin{appendices}

\section{Neural network architecture}
\label{sec:network_architecture}

\begin{figure*}
    \centering
    \subfigure[ours and no opt]{\includegraphics[width=0.6\textwidth]{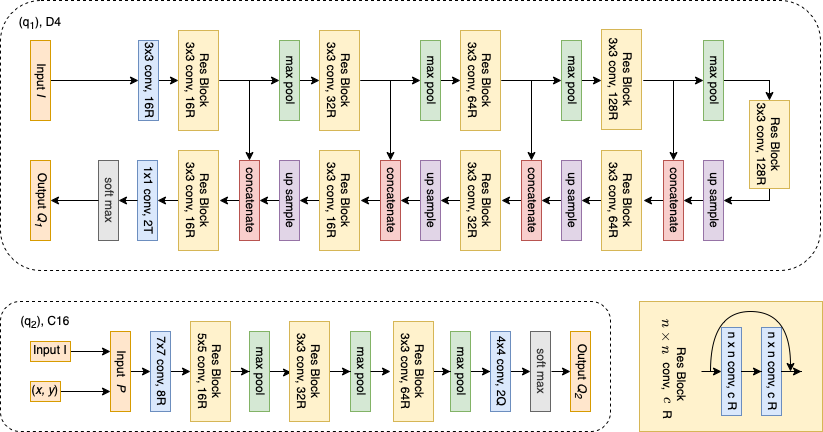}}
    \subfigure[no asr]{\includegraphics[width=0.6\textwidth]{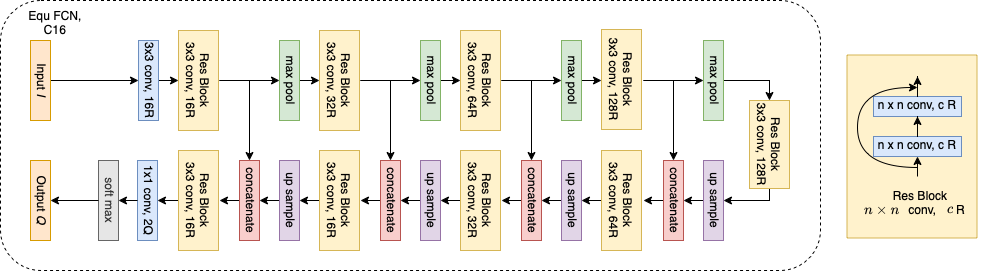}}
    \subfigure[no equ]{\includegraphics[width=0.6\textwidth]{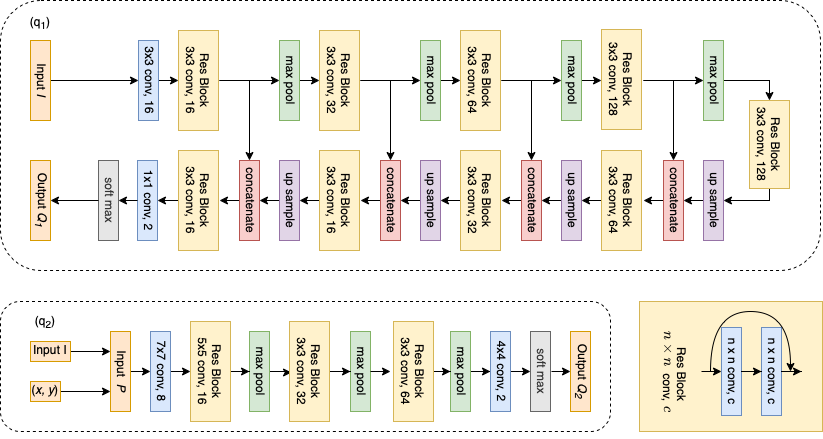}}
    \subfigure[Rot equ]{\includegraphics[width=0.6\textwidth]{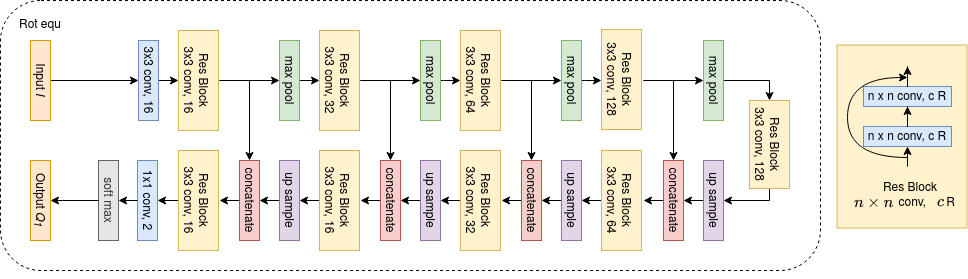}}
    \caption{The neural network architecture for ours and ablations. R means regular representation, T means trivial representation, and Q means quotient representation.}
    \label{fig:q1q2_architecture}
\end{figure*}

Our network architecture is shown in Figure~\ref{fig:q1q2_architecture}a.
The $q1$ network is a fully convolutional UNet \cite{u_net}. The $q2$ network is a residual neural network \cite{resnet}. These networks are implemented using PyTorch \cite{pytorch}, and the equivariant networks are implemented using the E2CNN library \cite{e2cnn}. Adam optimizer \cite{adam} is used for the SGD step. The ablation no opt has the same architecture as above. The ablation no asr (Figure~\ref{fig:q1q2_architecture}b) ablated the $q2$ network and is defined with respect to group $C16$. The ablation no equ   
(Figure~\ref{fig:q1q2_architecture}c) has a similar network architecture as ours with approximately the same number of free weights. However, the equivariant network is replaced with an FCN. The ablation rot equ (Figure~\ref{fig:q1q2_architecture}d) has a similar network architecture as no asr method with approximately the same number of free weights. However, the equivariant network is replaced with an FCN.

\section{Examples of Equivariant Layer}
\label{sec:equ_example}

\begin{figure}
    \centering
    \subfigure[With original input]{\includegraphics[width=0.4\textwidth]{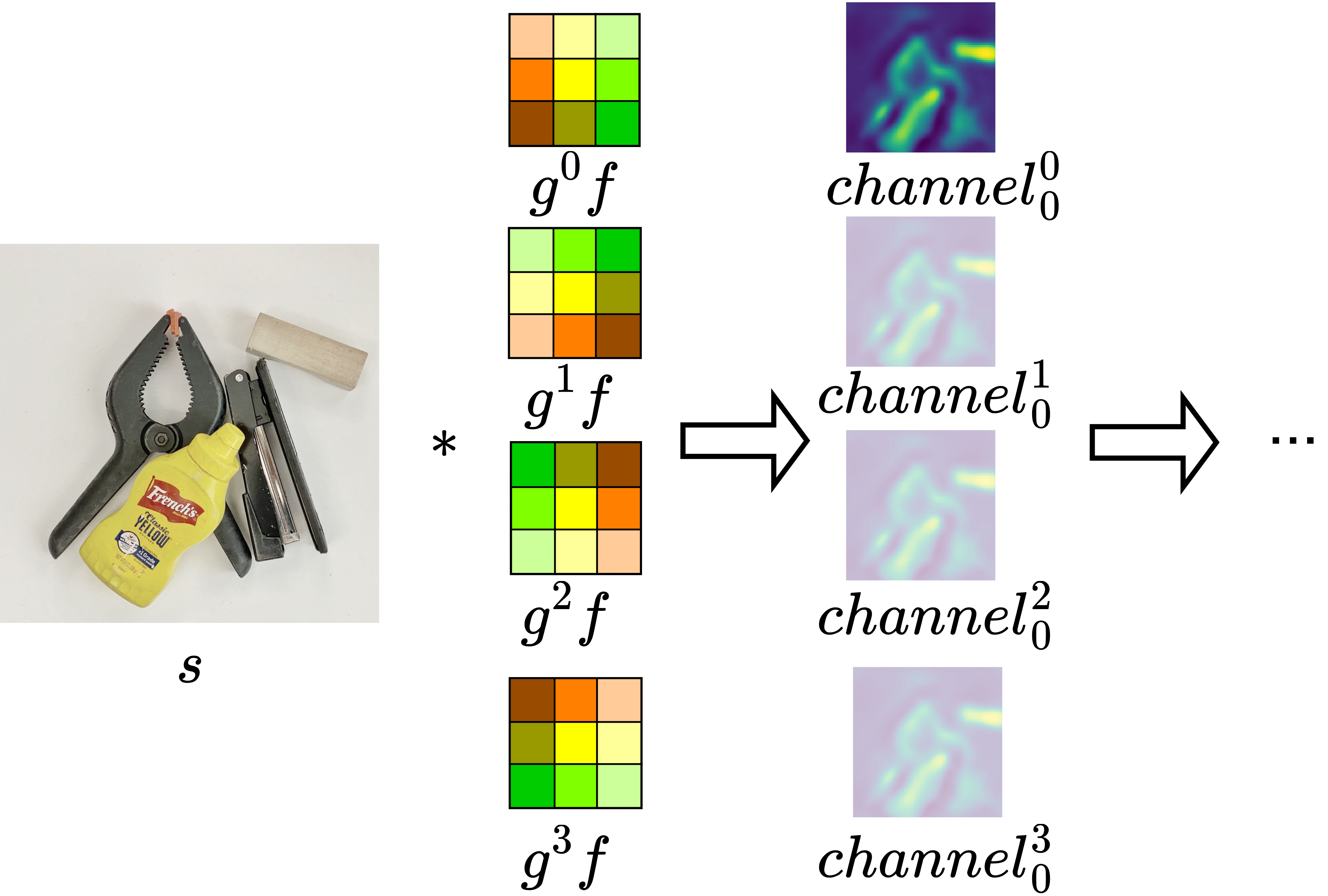}}
    \subfigure[With rotated input]{\includegraphics[width=0.4\textwidth]{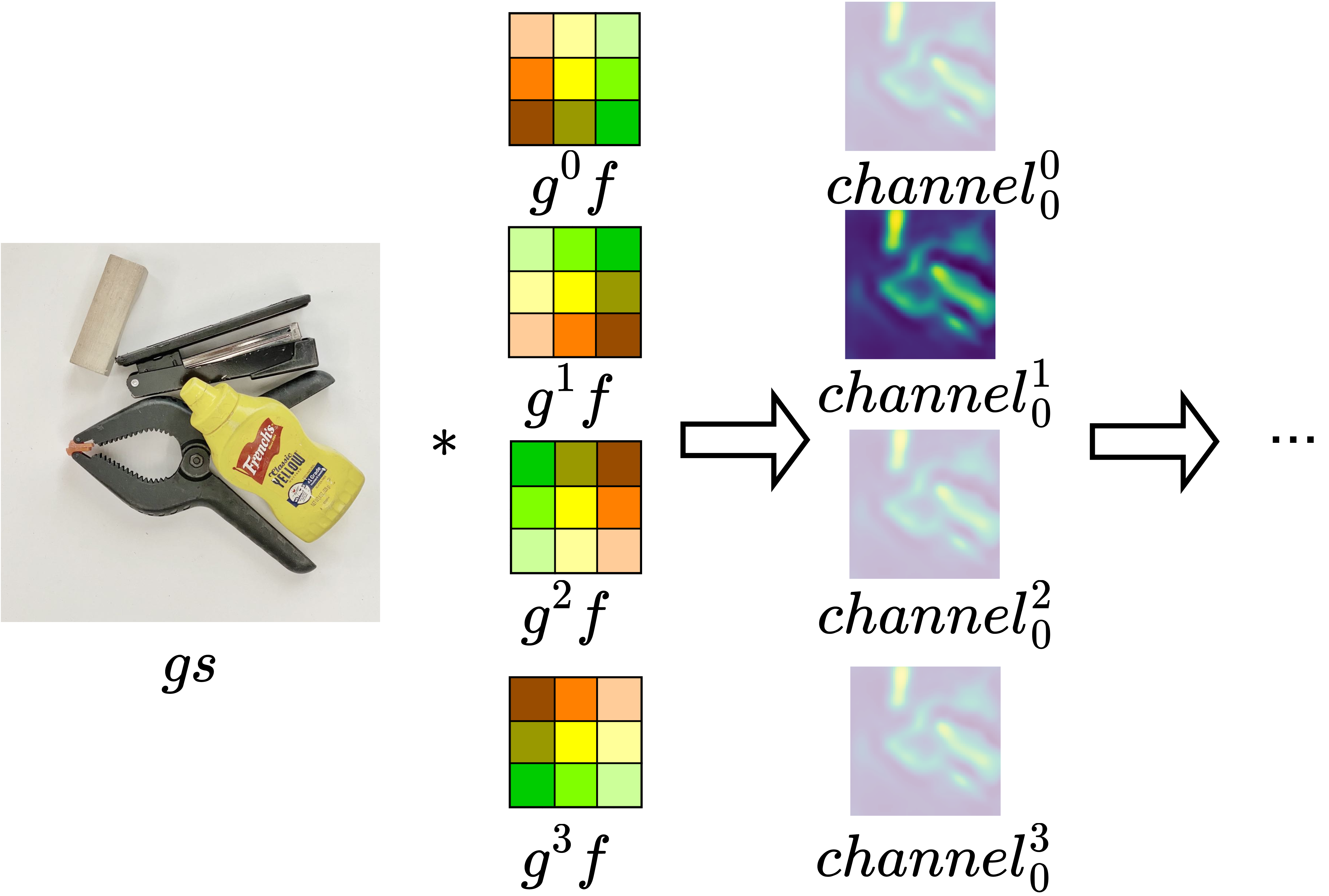}}
    \caption{An example of $C4$ equivariant layer that maps 3 channels of trivial representation to 1 channel of regular representation. $s$: the 3 trivial channel RGB image. $f$: the $3\time3$ convolutional filter. $g\in C4$: the group action that rotates the signal 90 degrees. $channel_0^i$: one regular channel in the neural network with $i = 0, 1, 2, 3$ fiber channels for each group element in $C4$. $*$ stands for standard convolution in CNN. Rotating image $s$ by $g$ leads to rotating the $channel_0$ spatially by $g$ and shifting the $channel_0$ along fiber channel by 1}
    \label{fig:equ_layer}
\end{figure}

Figure.\ref{fig:equ_layer} shows an example of an equivariant layer with $C4$ group that maps trivial representation to regular representation. In Figure.\ref{fig:equ_layer}. (a) the equivariant layer will initialize a convolutional filter $f$, then rotates the filter by each group element to obtain $f, gf, g^2f, g^3f$. In Figure.\ref{fig:equ_layer}. (b) since each filter is a rotated copy of $f$ and is correspond to a fiber channel in the regular representation, rotating image $s$ by $g$ leads to rotating the $channel_0$ spatially by $g$ and shifting the $channel_0$ along the fiber channel by 1. There could be more layers following this layer, as denoted by arrows and ellipsis. When each layer is equivariant, the entire neural network is equivariant.

\section{Augmentation baseline choices}
\label{sec:aug_baseline}

The data augmentation strategies are: \underline{$n\times$ RAD:} The method from~\cite{rad} where we perform $n$ SGD steps after each grasp sample, where each SGD step is taken over a mini-batch $bs$ of samples that have been randomly translated and rotated by $g\in \SE(2)$ according to equation. \ref{eqn:grasp_invariance_assumption}. Specifically, applying random $\SE(2)$ transformation $g$ for both observation and action: $(gs, ga, r)$.
\underline{$n\times$ soft equ:} a data augmentation method \cite{wang2021equivariant} that performs $n$ soft equivariant SGD steps per grasp, where each SGD step is taken over $n$ times randomly $\hat{\SE}(2)$ augmented mini-batch.
Specifically, we sample $bs/n$ samples ($bs$ is the batch size), augment it $n$ times, and train on this mini-batch. We perform these SGD steps $n$ times so that $bs$ transitions are sampled. This augmentation aims at achieving equivariance in the mini-batch. 

We apply $n\times$ RAD and $n\times$ soft equ data augmentation to both VPG and FC-GQ-CNN baselines, with $n={2, 4, 8}$. The best data augmentation parameters $n$ are chosen for each baseline in the comparison in Figure~\ref{fig:simulation_baseline}.



\begin{figure*}
    \centering
    \subfigure[]{\includegraphics[width=0.13\textwidth]{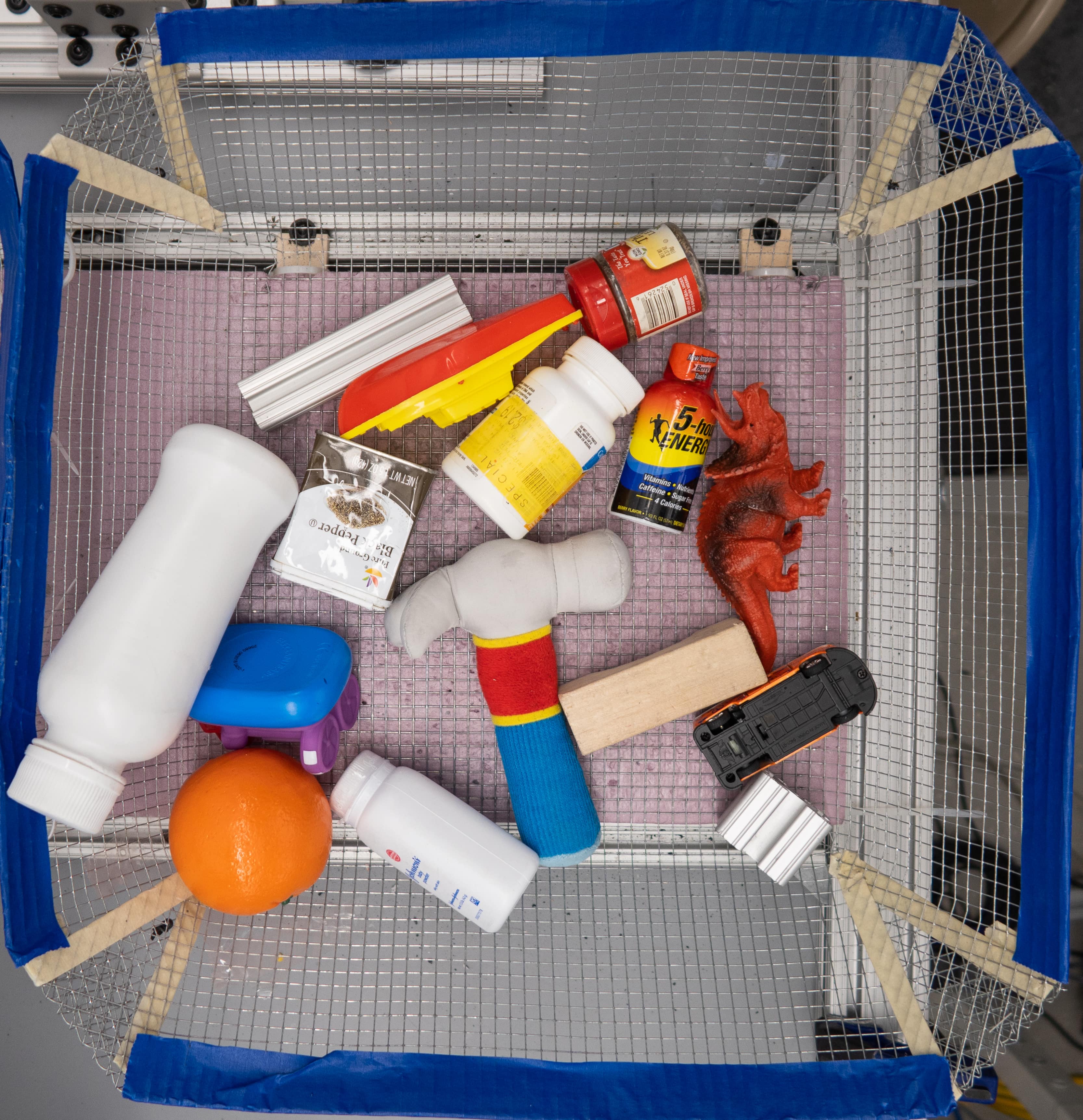}}
    \subfigure[]{\includegraphics[width=0.16\textwidth]{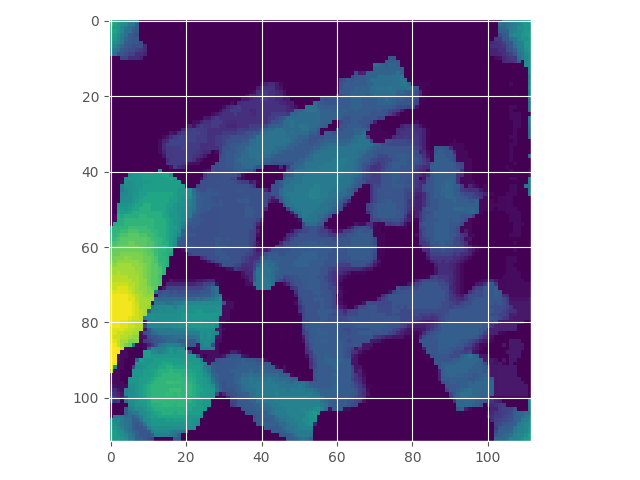}}
    \subfigure[]{\includegraphics[width=0.16\textwidth]{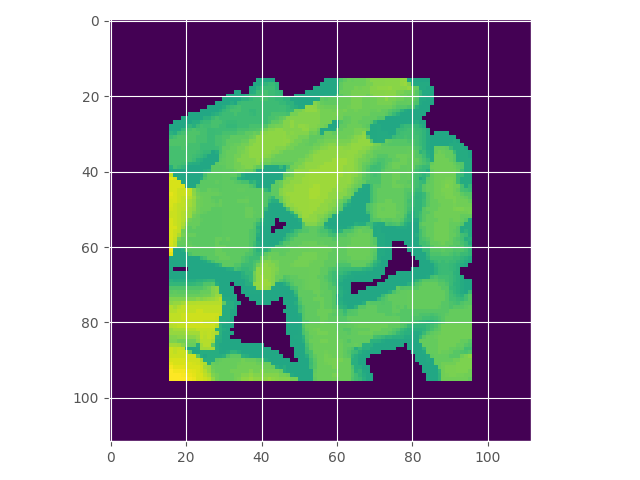}}
    \subfigure[]{\includegraphics[width=0.15\textwidth]{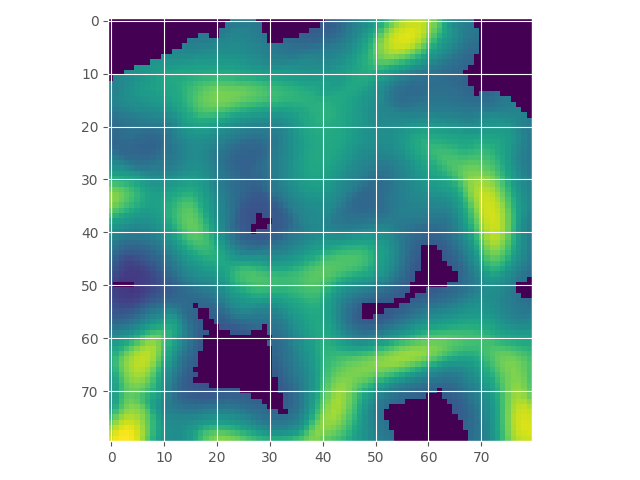}}
    \subfigure[]{\includegraphics[width=0.13\textwidth]{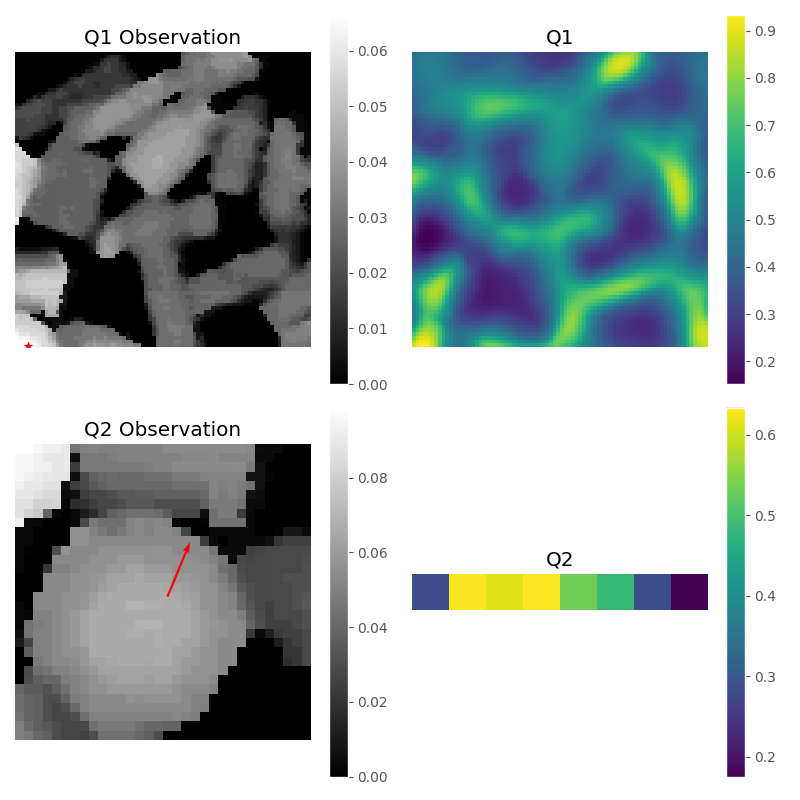}}
    \subfigure[]{\includegraphics[width=0.18\textwidth]{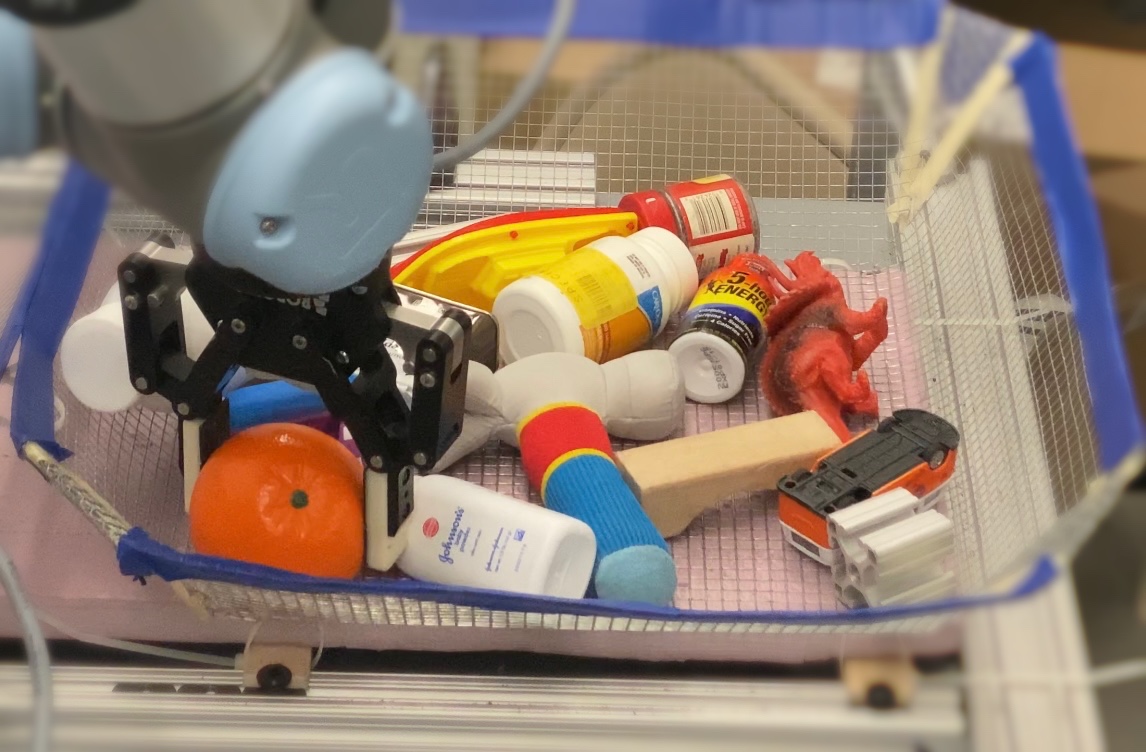}}
    \caption{Action space constraint for action selection. (a) The test set easy cluttered scene. (b) The state $s$. (c) The action space $x_\text{positive}$, overlays the binary mask $x_\text{positive}$ with the state $s$ for visualization. (d) The $Q$-values within the action space. (e) Selecting an action. (f) Executing a grasp.}
    \label{fig:action_space}
\end{figure*}

\section{Additional experiments}
\label{sec:add_ablations}

\begin{figure}
    \centering
    \subfigure[Training]{\includegraphics[width=0.23\textwidth]{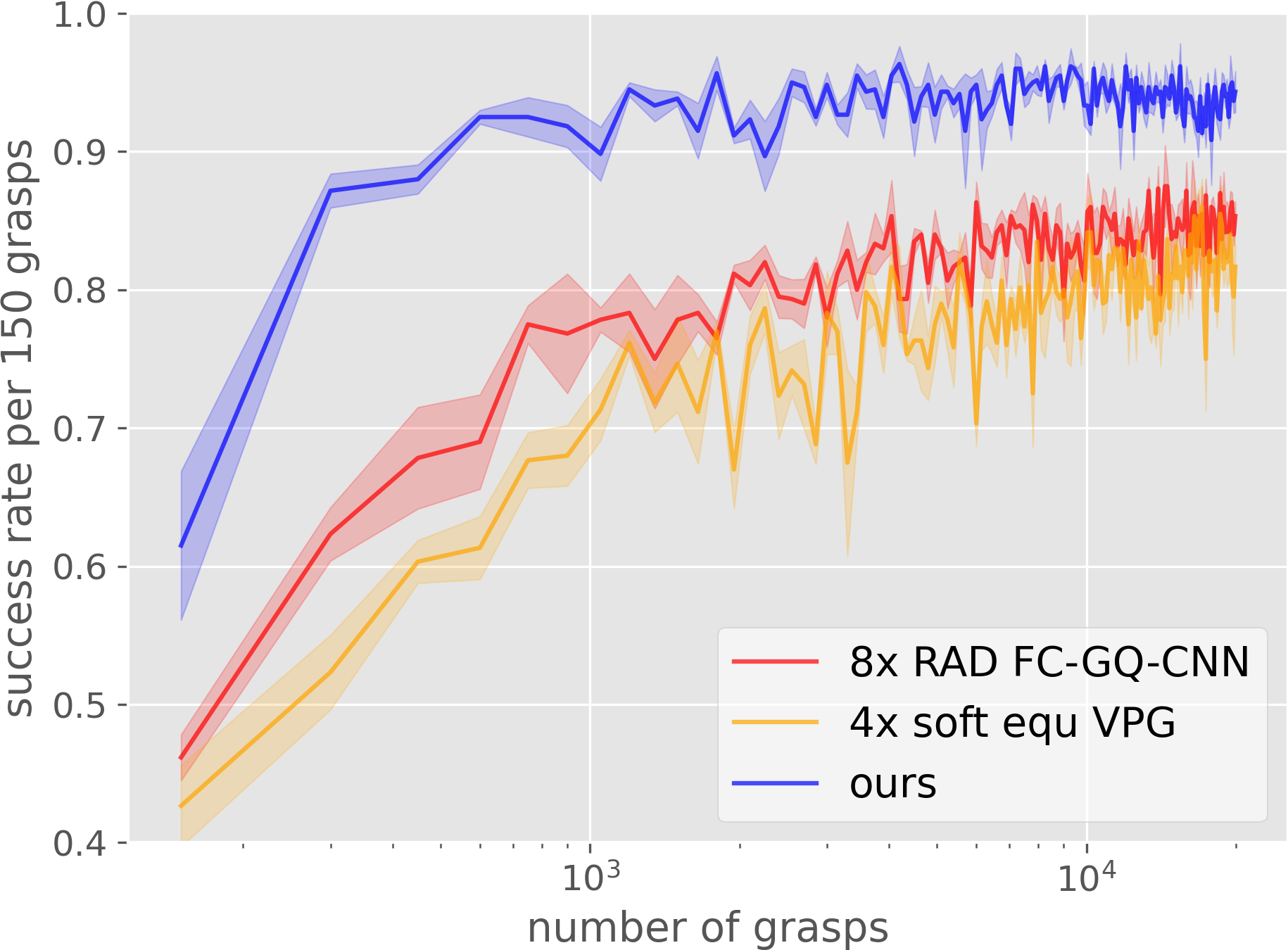}}
    \subfigure[Testing]{\includegraphics[width=0.23\textwidth]{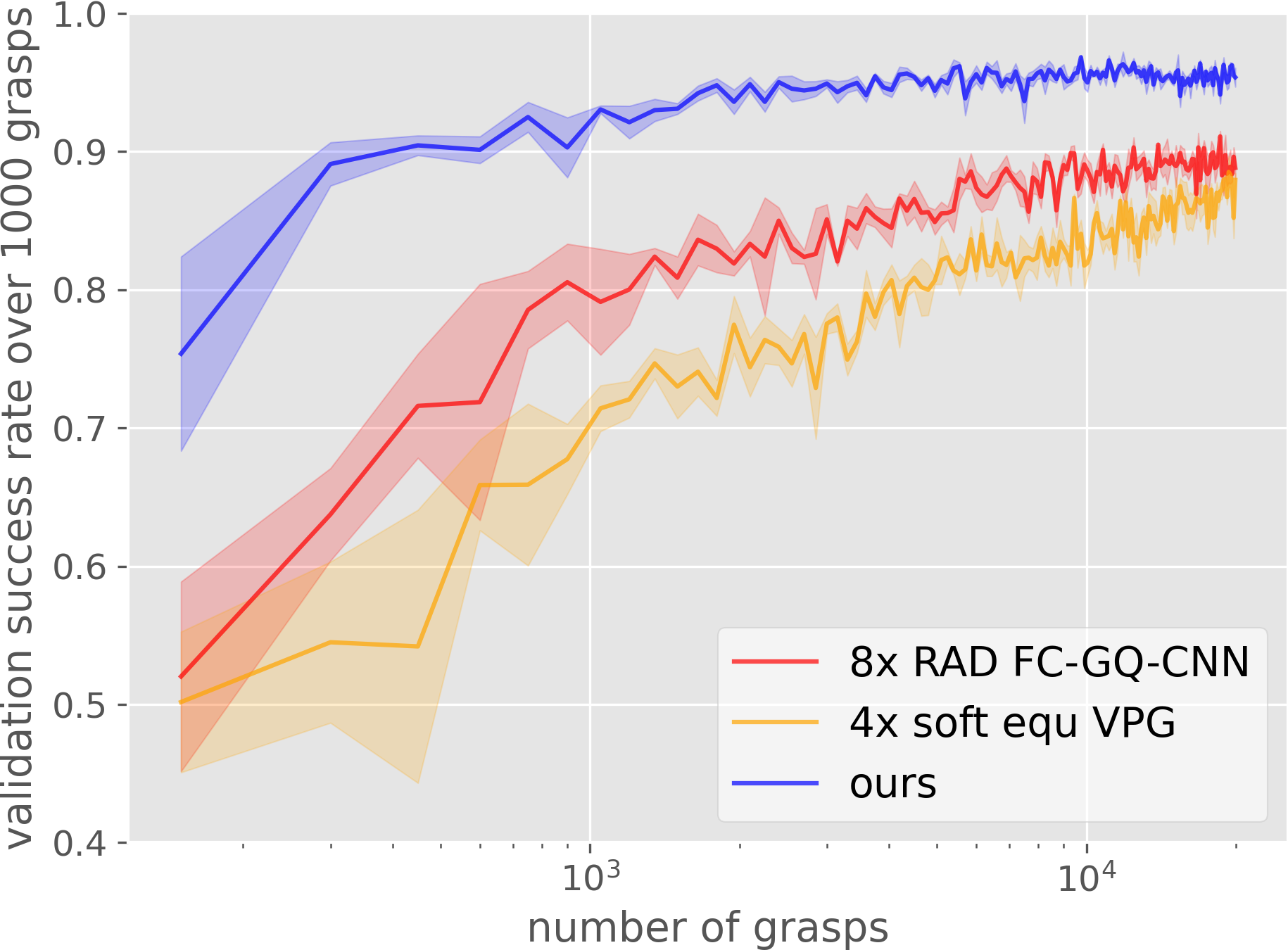}}
    \caption{Additional comparison with baselines for Section.\ref{sect:baselines} with depth modality. Lines are an average over $4$ runs. Shading denotes standard error. (a) learning curves as a running average over the last 150 training grasps. (b) average near-greedy performance of 1000 validation grasps performed every 150 training steps.}
    \label{fig:add_simulaiton_baseline}
\end{figure}

To investigate the sample efficiency of SymGrasp, we compare our method with the other two best methods in Figure.\ref{fig:simulation_baseline}. We train all the methods with $20$k grasps to illustrate the sample efficiency. As shown in Figure.\ref{fig:add_simulaiton_baseline}, our method not only learns faster but also converges to a better success rate, compared with the other two baselines.

We then ablate each components in Section.\ref{sec:other_optimizations}. Baselines are: In \underline{ASR loss}, both $q_1$ and $q_2$ minimize $l2$ loss between prediction and the reward $r$. In \underline{No prioritizing}, the mini-batch is uniformly sampled from the replay buffer. In \underline{e greedy}, we replace Boltzmann exploration with e greedy exploration that linearly anneals from 0.5 to 0.1 in 500 grasps. In \underline{No data aug}, no data augmentation is performed. In \underline{No softmax} the pixel-wise softmax in the last layer of $q_1$ and $q_2$ is removed.

\begin{figure}
    \centering
    \subfigure[Training]{\includegraphics[width=0.23\textwidth]{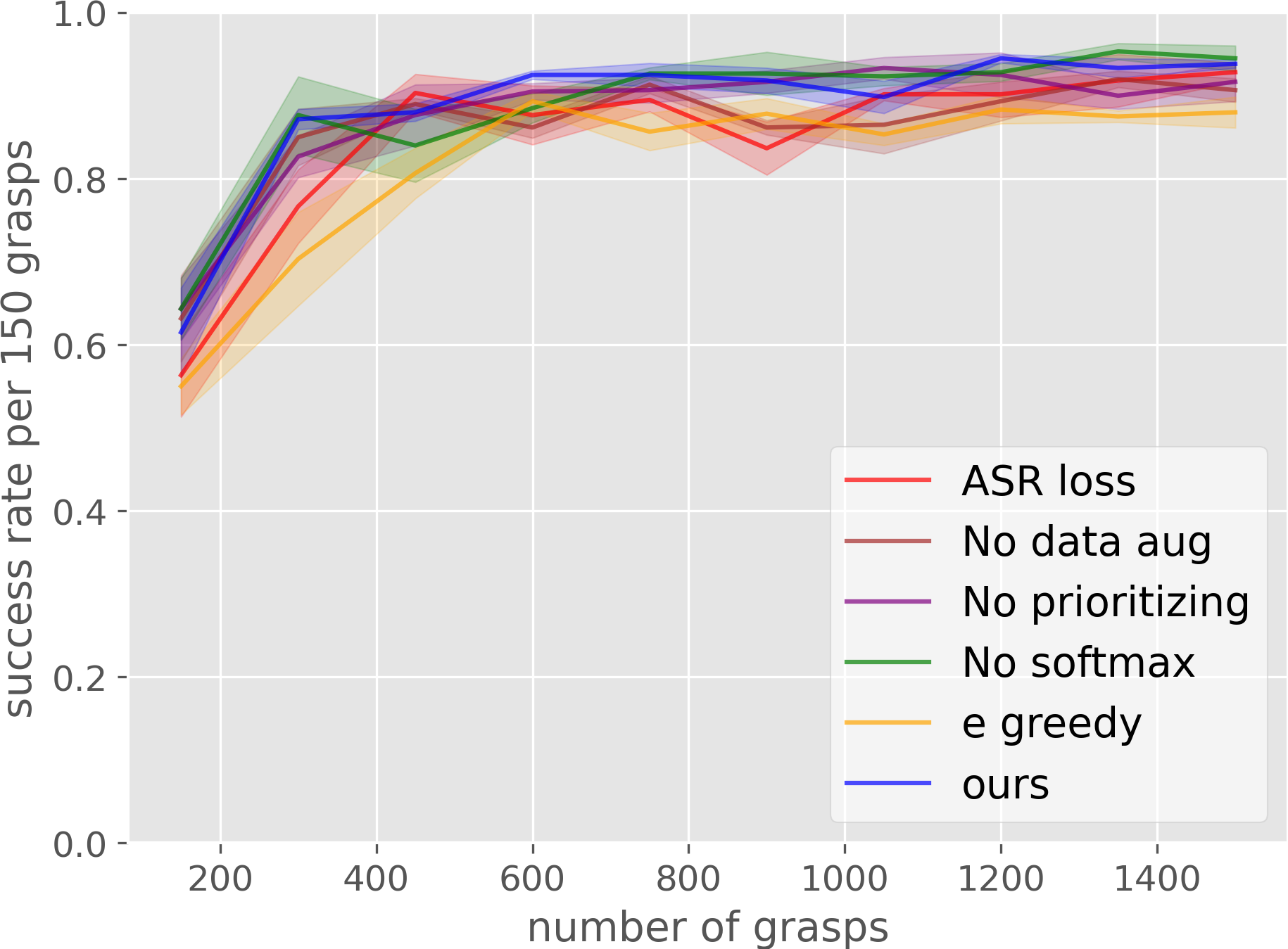}}
    \subfigure[Testing]{\includegraphics[width=0.23\textwidth]{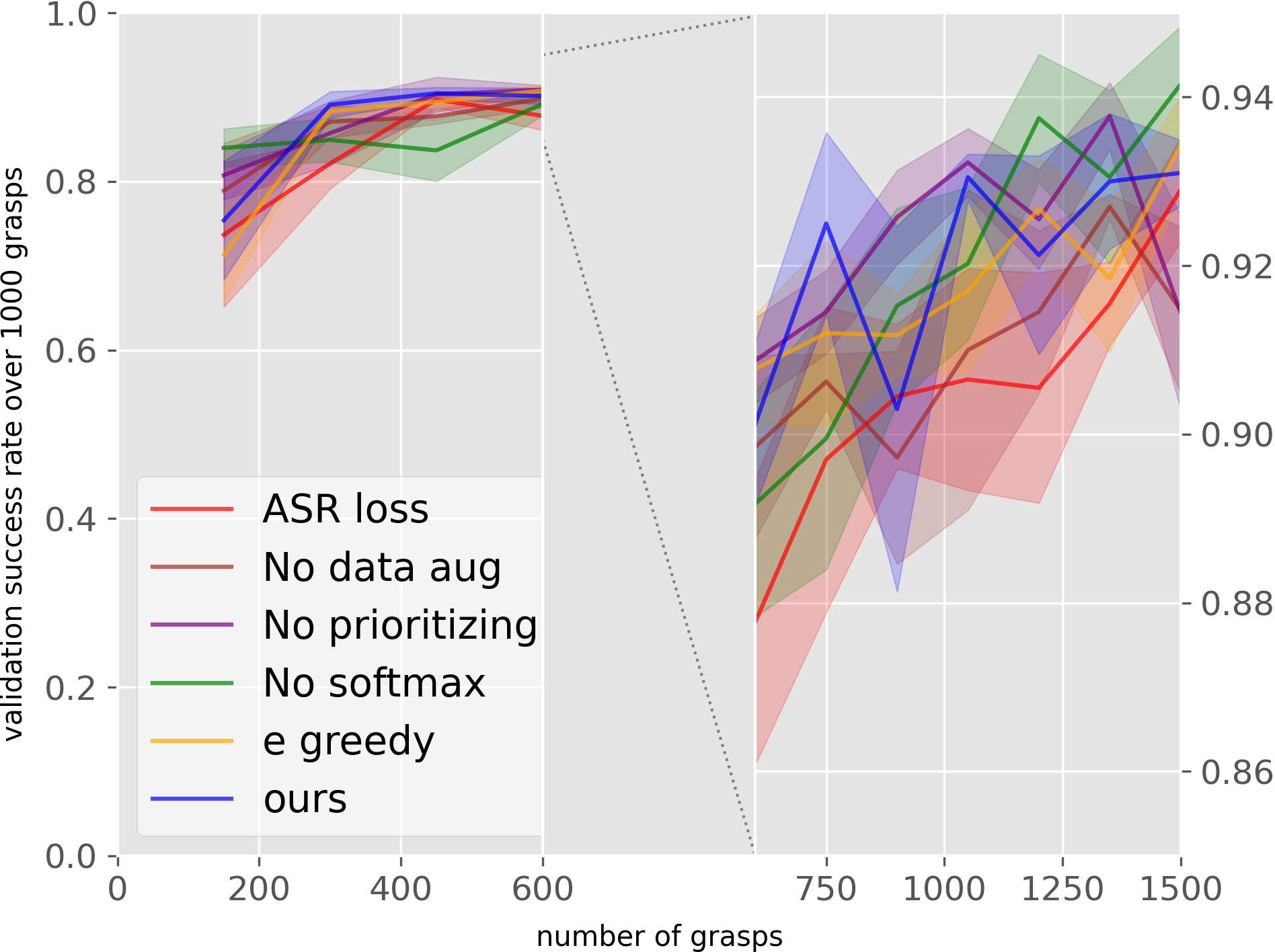}}
    \caption{Additional ablation study for Section.\ref{sec:other_optimizations} with depth modality. Lines are an average over $4$ runs. Shading denotes standard error. (a) learning curves as a running average over the last 150 training grasps. (b) average near-greedy performance of 1000 validation grasps performed every 150 training steps.}
    \label{fig:add_opt_ablation}
\end{figure}

Figure~\ref{fig:add_opt_ablation} a, b, shows the learning results. Figure.\ref{fig:add_opt_ablation}(b) shows that \underline{ASR loss} affects the performance the most, indicating the importance of the loss function in Section.\ref{sec:loss_function}. Other components \underline{No data aug}, \underline{e greedy} slightly reduces the performance indicating the marginal benefits of data augmentation and Boltzmann exploration. Lastly, \underline{No softmax} performs badly at the $600$ grasps and \underline{No prioritizing} performs badly at the $1500$ grasps, these suggest softmax and prioritizing failure grasp stabilizes learning.

We then ablate each component in Section.\ref{sec:opt_rgb}. Baselines are: In \underline{Cyclic group}, we replace $D16$ group with $C16$ group in $q2$. In \underline{No collision penalty}, we replace collision penalty by using binary grasp success reward.

\begin{figure}
    \centering
    \subfigure[Training]{\includegraphics[width=0.23\textwidth]{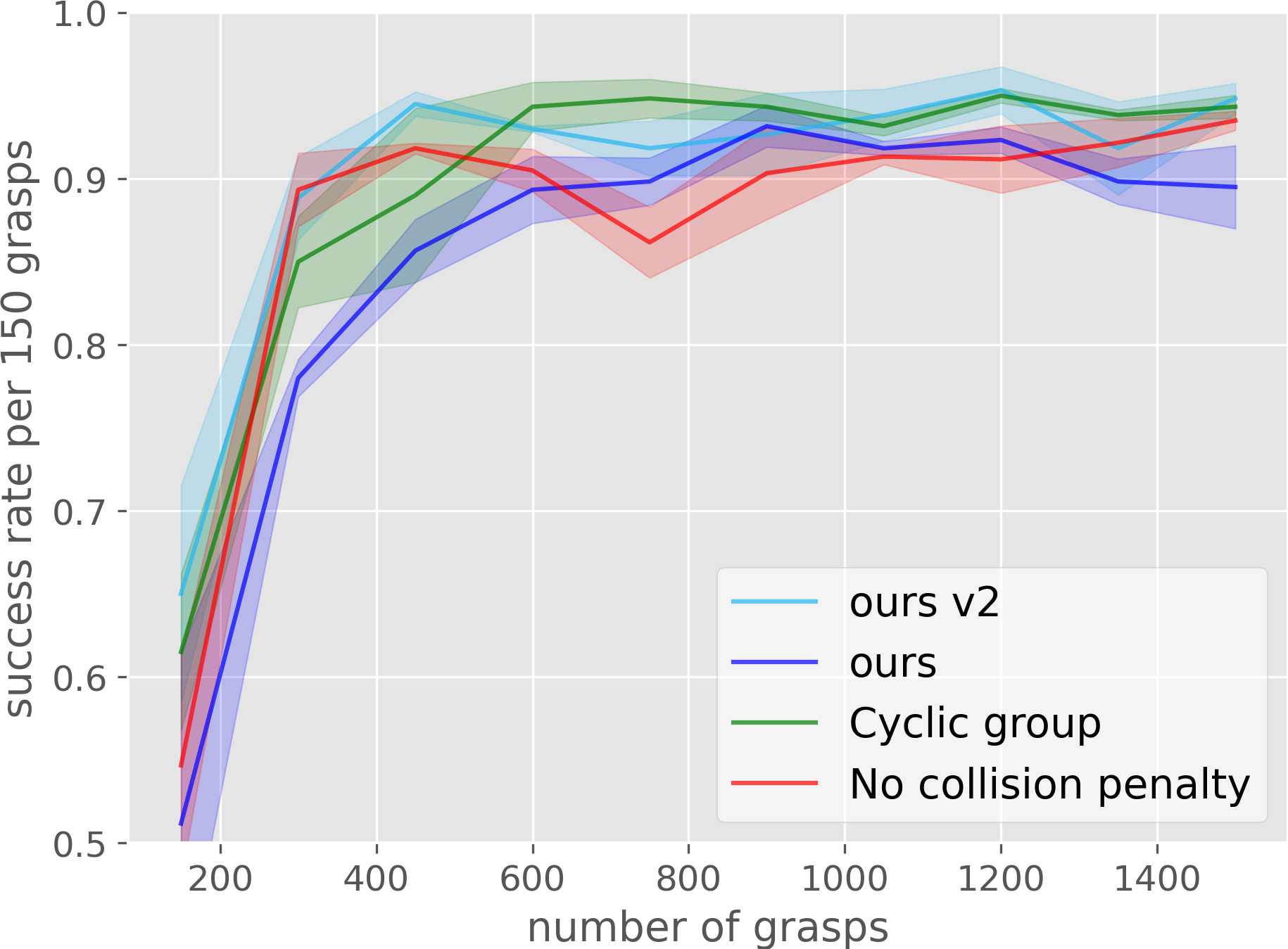}}
    \subfigure[Testing]{\includegraphics[width=0.23\textwidth]{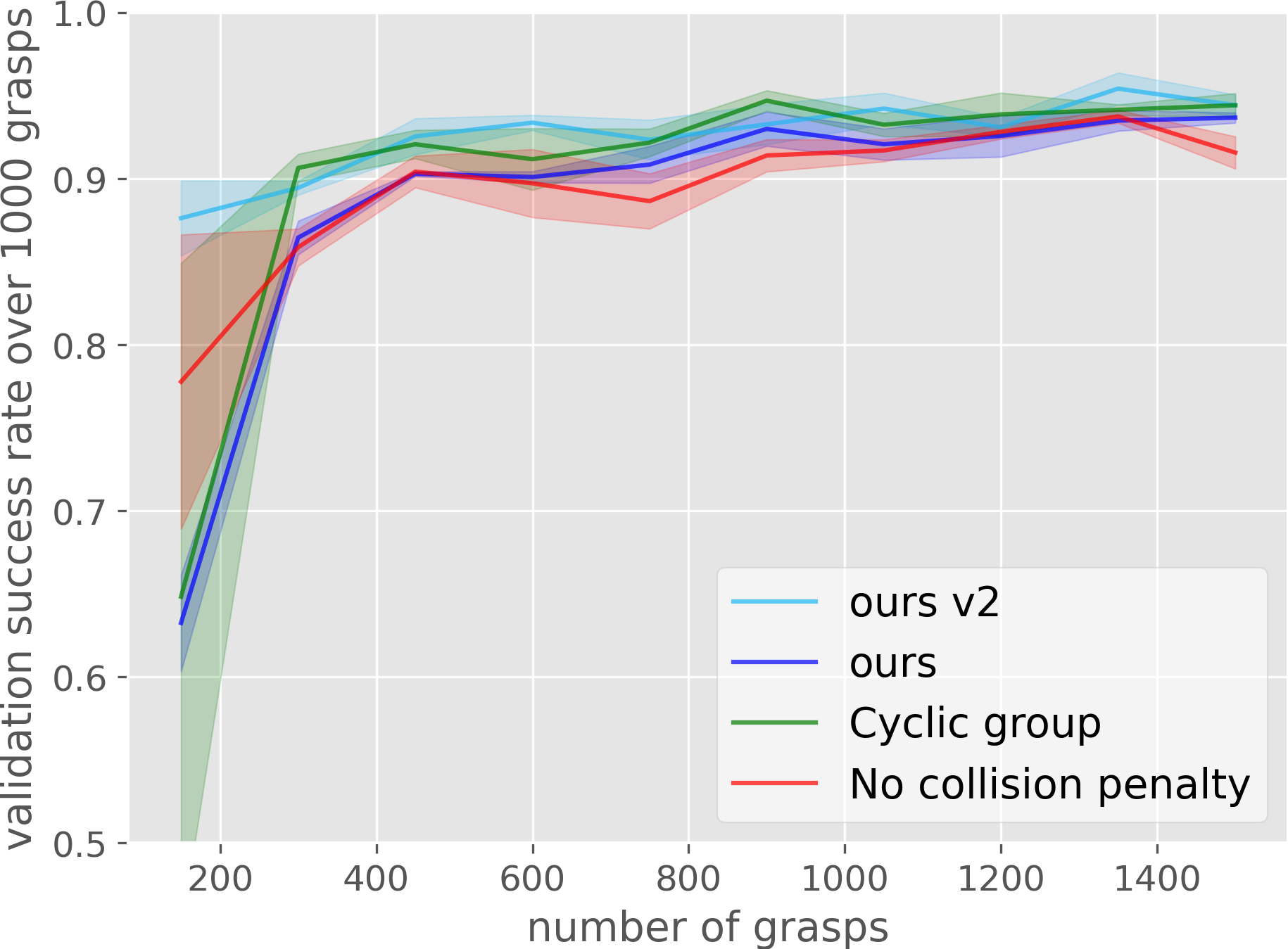}}
    \caption{Additional ablation study for Section.\ref{sec:opt_rgb} with depth modality. Lines are an average over $4$ runs. Shading denotes standard error. (a) learning curves as a running average over the last 150 training grasps. (b) average near-greedy performance of 1000 validation grasps performed every 150 training steps.}
    \label{fig:add_simulaiton_ablation}
\end{figure}

Figure~\ref{fig:add_simulaiton_ablation} a, b, shows the learning results. \underline{Cyclic group} shows Cyclic group in $q_2$ leads to slower learning before 300 grasp, compares to Dihedral group in $q_2$; \underline{No collision penalty} shows collision penalty encourages quicker and better convergence.





\begin{figure}
    \centering
    \includegraphics[width=0.3\textwidth]{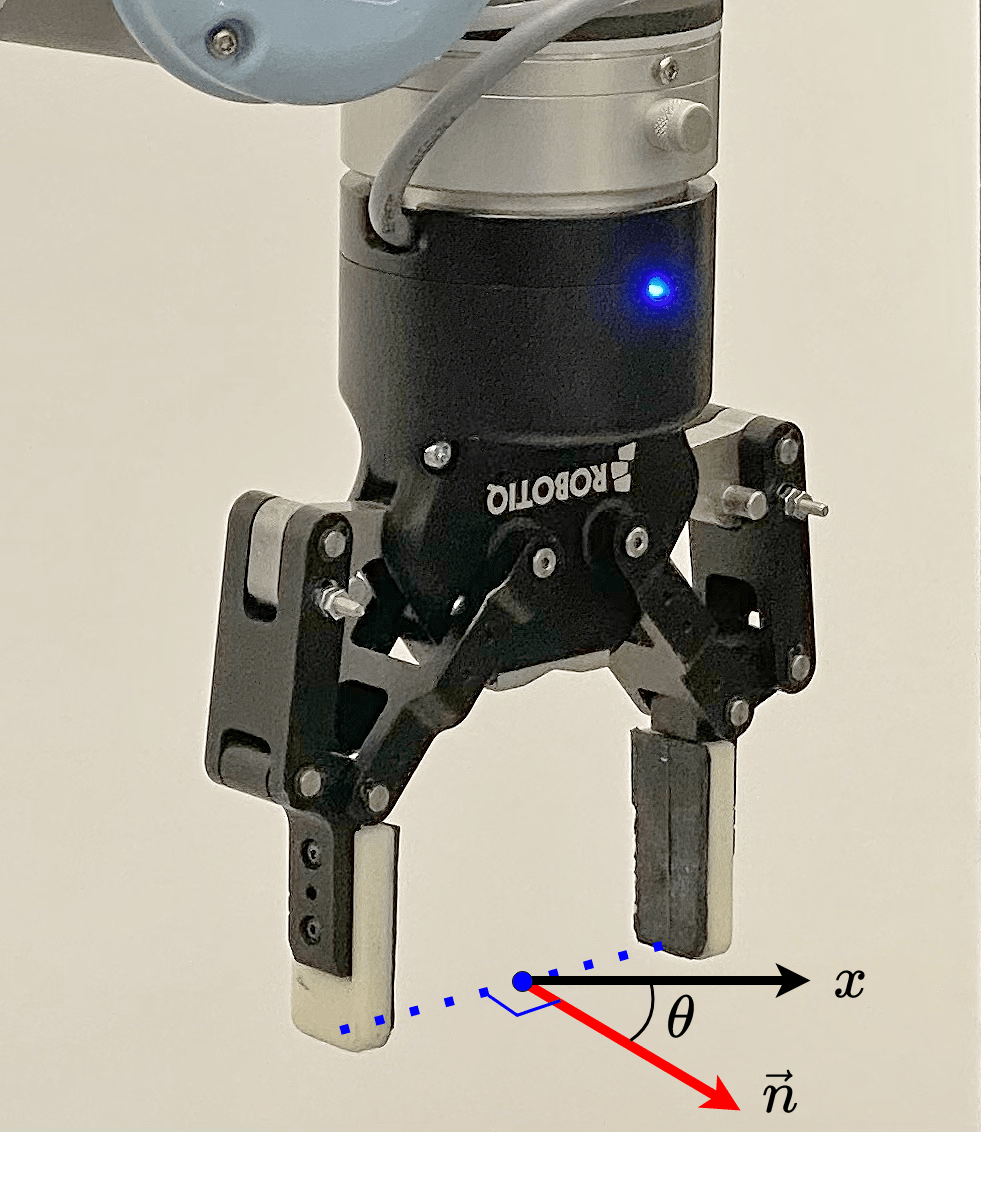}
    \caption{The definition of $\theta$. $x$ is the $x$-axle of the workspace while $\vec{n}$ is the normal of the gripper.}
    \label{fig:gripper}
\end{figure}


\begin{figure}
    \centering
    \subfigure[]{\includegraphics[width=0.17\textwidth]{figure/appendix/experiment_scene_map/black_grasp_scene_small_bottle.jpg}}
    \subfigure[]{\includegraphics[width=0.3\textwidth]{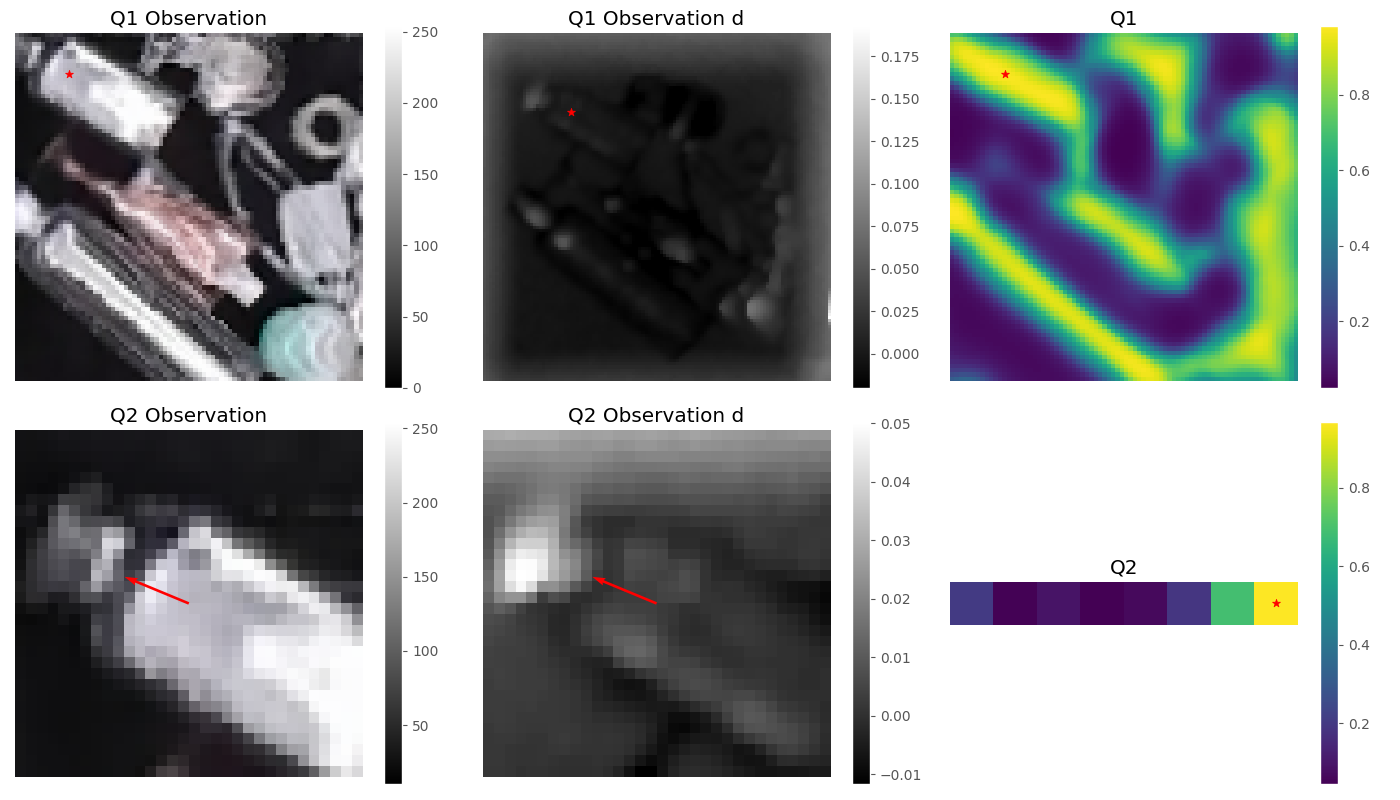}}\\
    \subfigure[]{\includegraphics[width=0.17\textwidth]{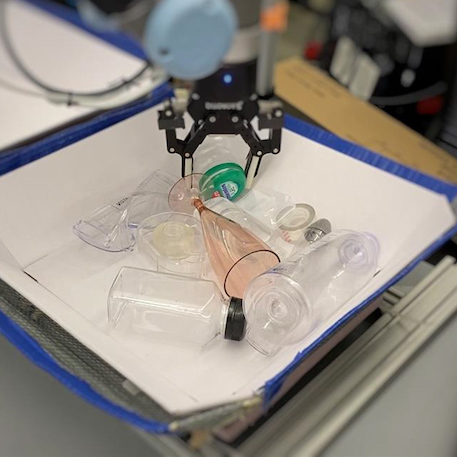}}
    \subfigure[]{\includegraphics[width=0.3\textwidth]{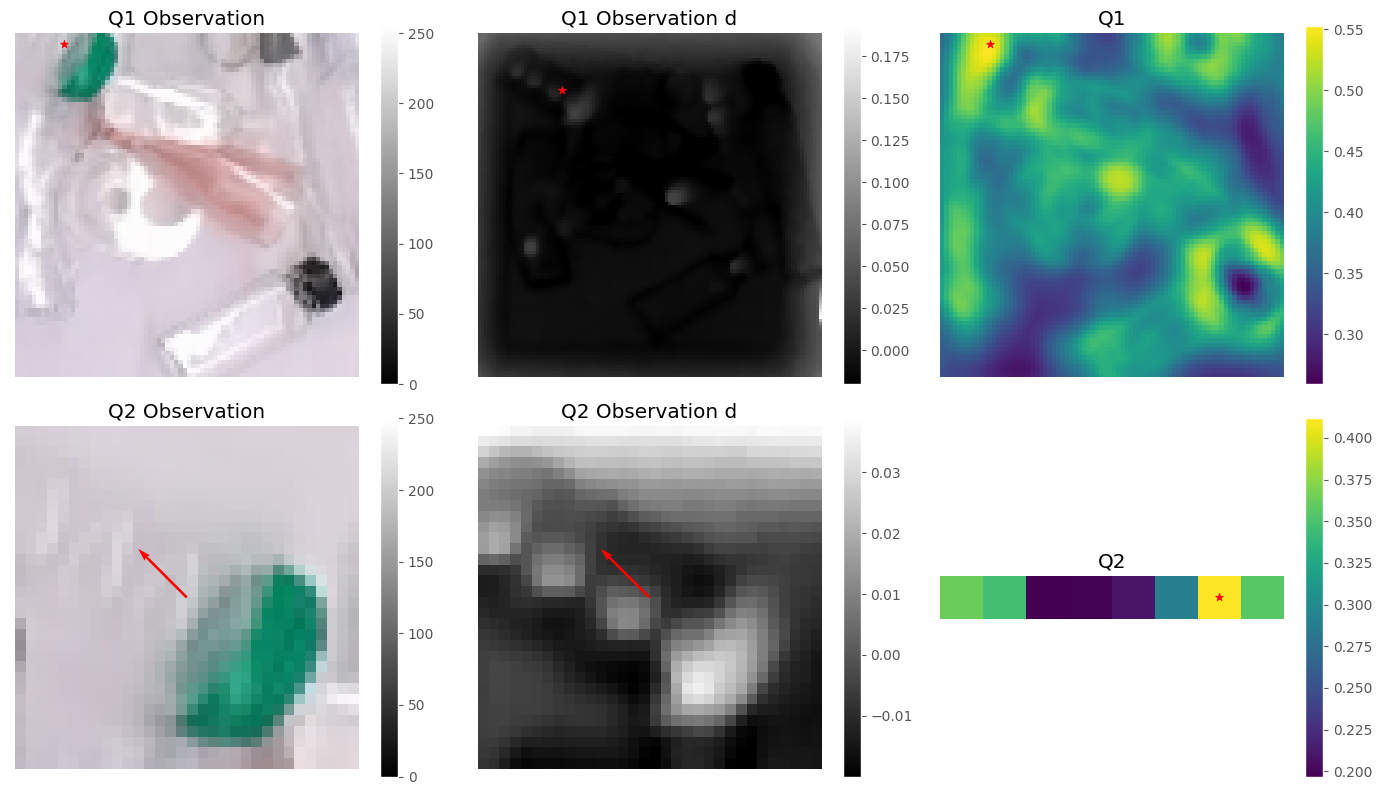}}
    \caption{Transparent object grasping scenes. First row: using black trays. Second row: using white trays.}
    \label{fig:tray_color}
\end{figure}


\section{Action space details}

The action $\theta$ is defined as the angle between the normal vector $\vec{n}$ of the gripper and the $x$-axle, see Figure~\ref{fig:gripper}.

While the grasping height is selected by heuristic, the grasping aperture is fixed with the maximum aperture of the gripper.

To prevent the grasps in the empty space where there is no object in $s$, we constrain the action space to $x_\text{positive} \in X$ to exclude the empty space, see Figure~\ref{fig:action_space}c. The constrain $x_\text{positive}$ is achieved by first thresholding the depth image: $s_\text{positive} = s > s_\text{threshold}$ ($s_\text{threshold}$ is 0.5cm in simulation and 1.5cm in hardware), then dilating this binary map $s_\text{positive}$ by radius $d_\text{dilation} = 4$ pixels. The parameter $s_\text{threshold}$ is selected according to sensor noise where $d_\text{dilation}$ is related to the half of the gripper aperture. Moreover, we constrain the action space within the tray to prevent collisions.

\section{Success and failure modes}

We list typical success and failure modes to evaluate the performance of SymGrasp.

\begin{figure*}
    \centering
    \subfigure[]{\includegraphics[width=0.2\textwidth]{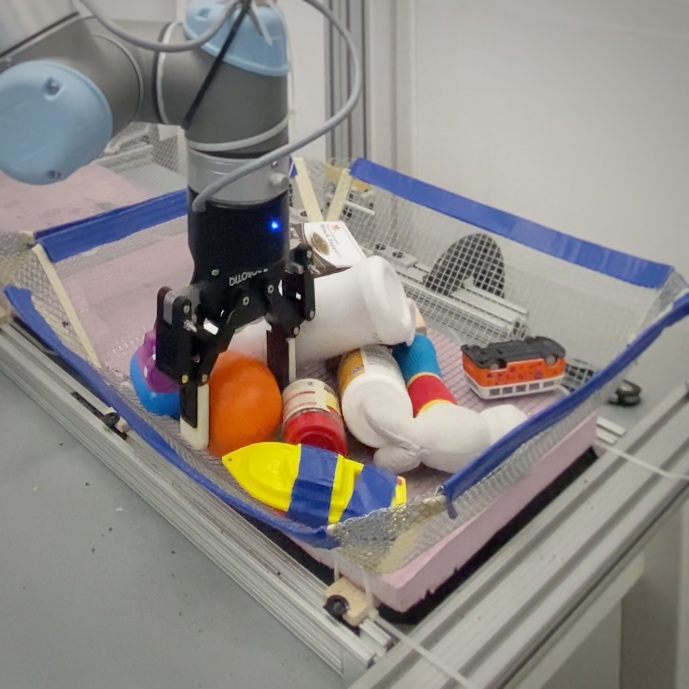}}
    \subfigure[]{\includegraphics[width=0.2\textwidth]{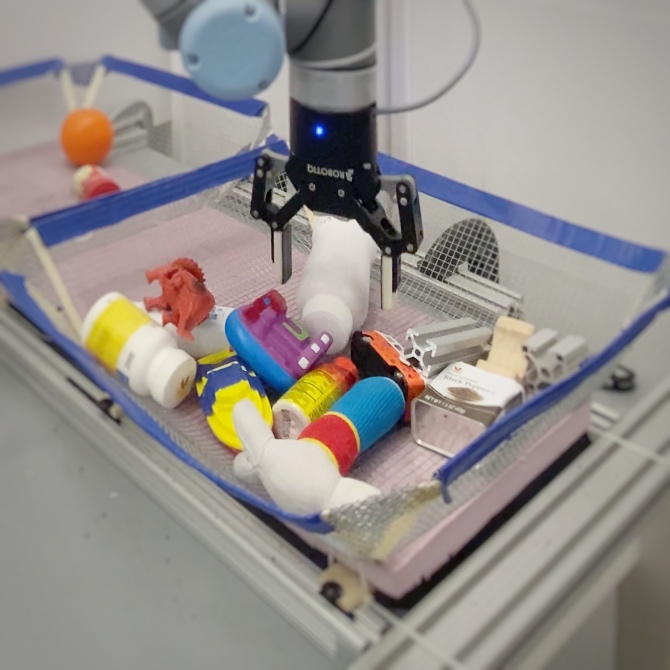}}
    \subfigure[]{\includegraphics[width=0.2\textwidth]{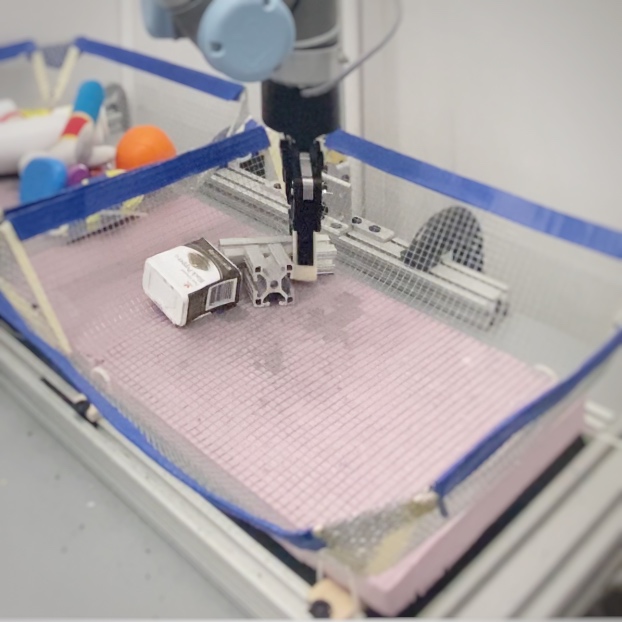}}
    \subfigure[]{\includegraphics[width=0.2\textwidth]{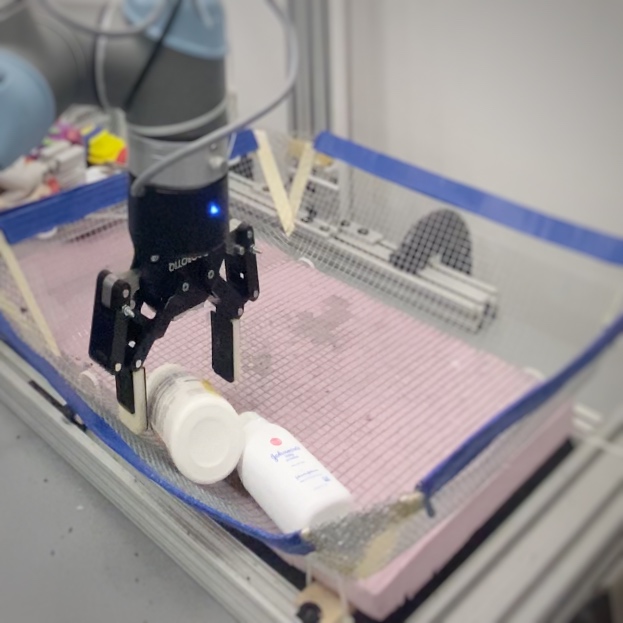}}
    \caption{Success modes in the test set easy, which has 15 hold out objects.}
    \label{fig:success_modes}
\end{figure*}

For success modes, the trained policy of SymGrasp showcases its capability. At the densely cluttered scene, SymGrasp prefers to grasp the relatively isolated part of the objects, see Figure~\ref{fig:success_modes}a, b. At the scene where the objects are close to each other, SymGrasp can find the grasp pose that is collision free with other objects, see Figure~\ref{fig:success_modes}c, d.

\begin{figure*}
    \centering
    \subfigure[Wrong $q_1$ (10/20)]{\includegraphics[width=0.22\textwidth]{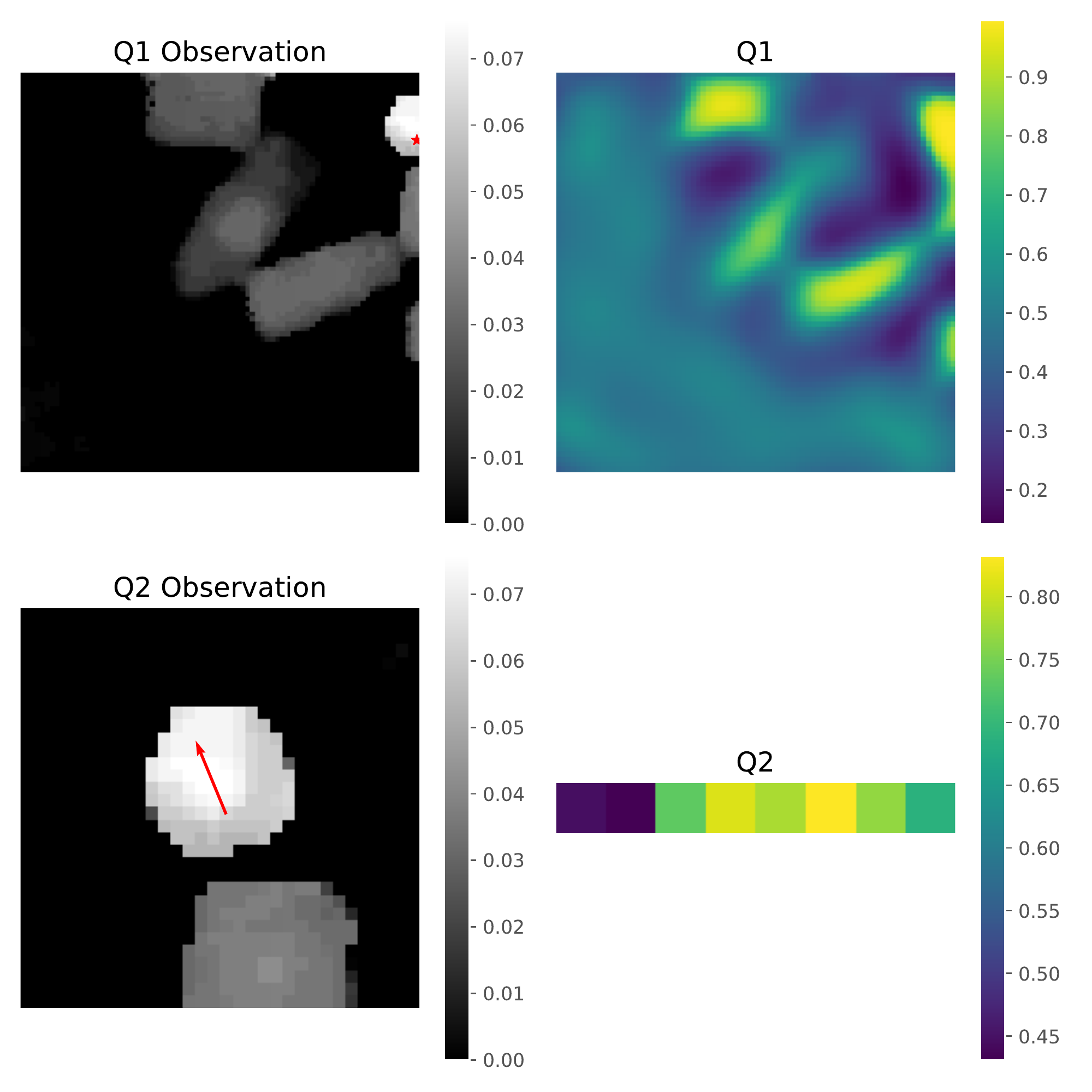}}
    \subfigure[Wrong $q_2$ (3/20)]{\includegraphics[width=0.22\textwidth]{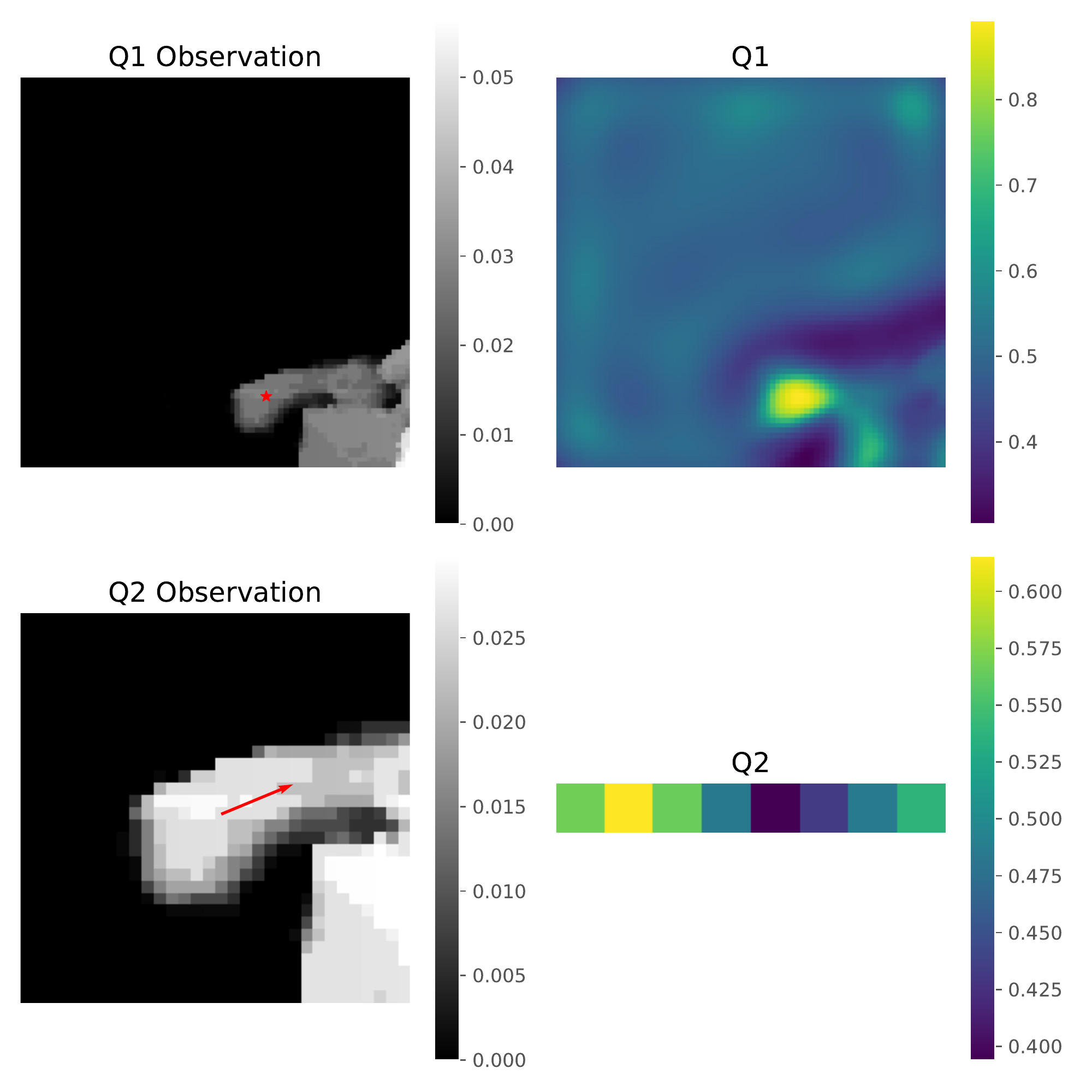}}
    \subfigure[Challenging scenes (3/20)]{\includegraphics[width=0.22\textwidth]{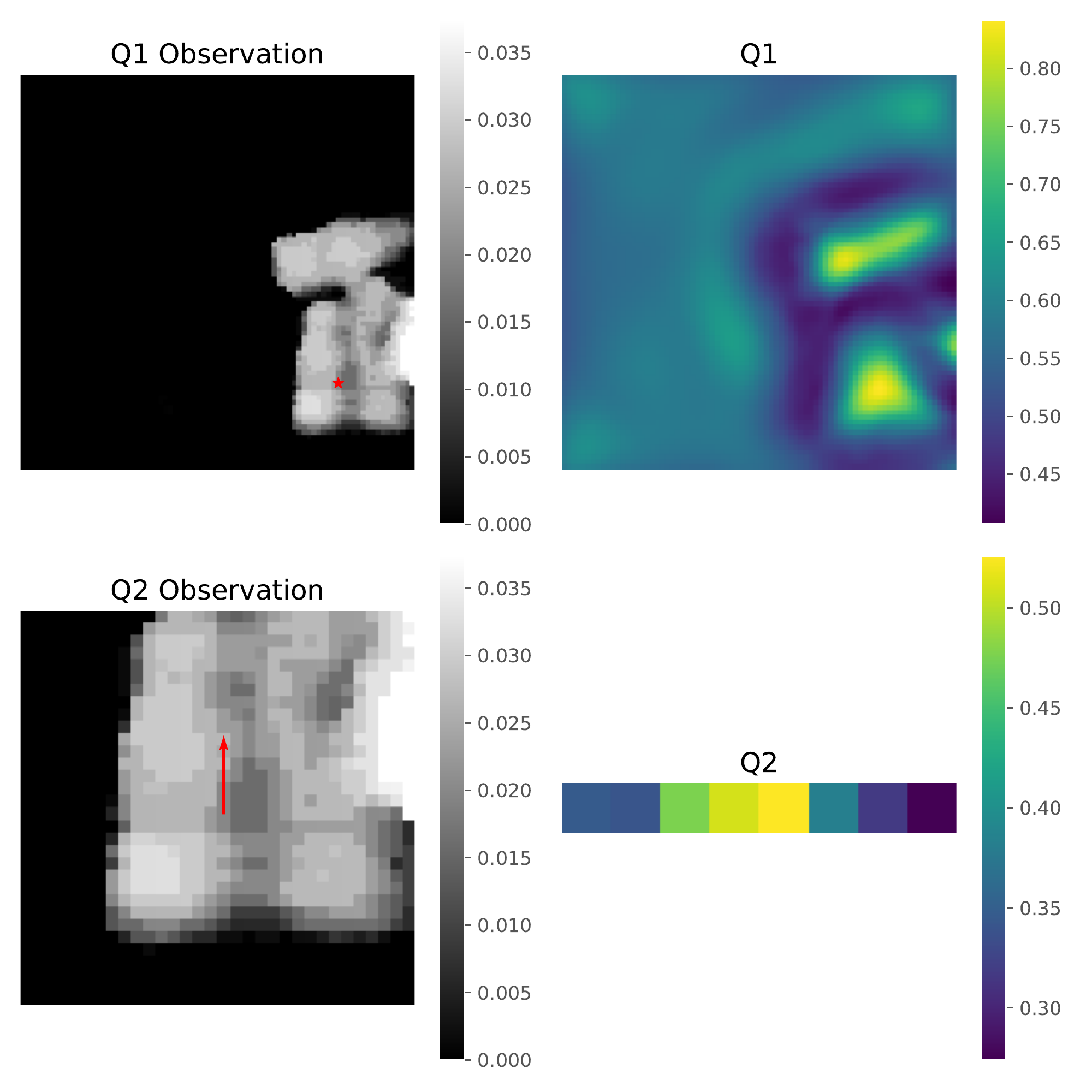}}
    \subfigure[Reasonable grasps (3/20)]{\includegraphics[width=0.22\textwidth]{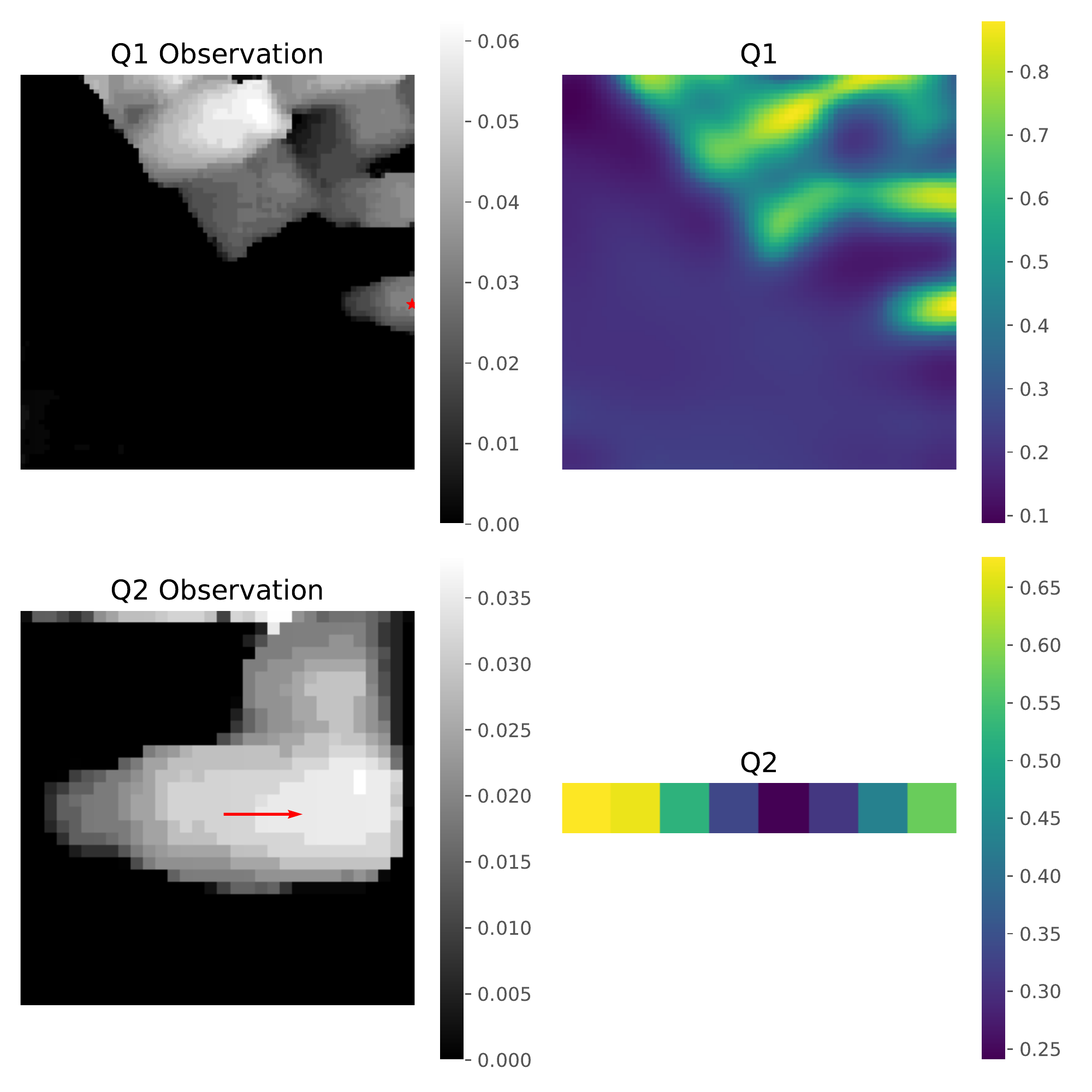}}
    \subfigure[Wrong $q_1$ (17/52)]{\includegraphics[width=0.22\textwidth]{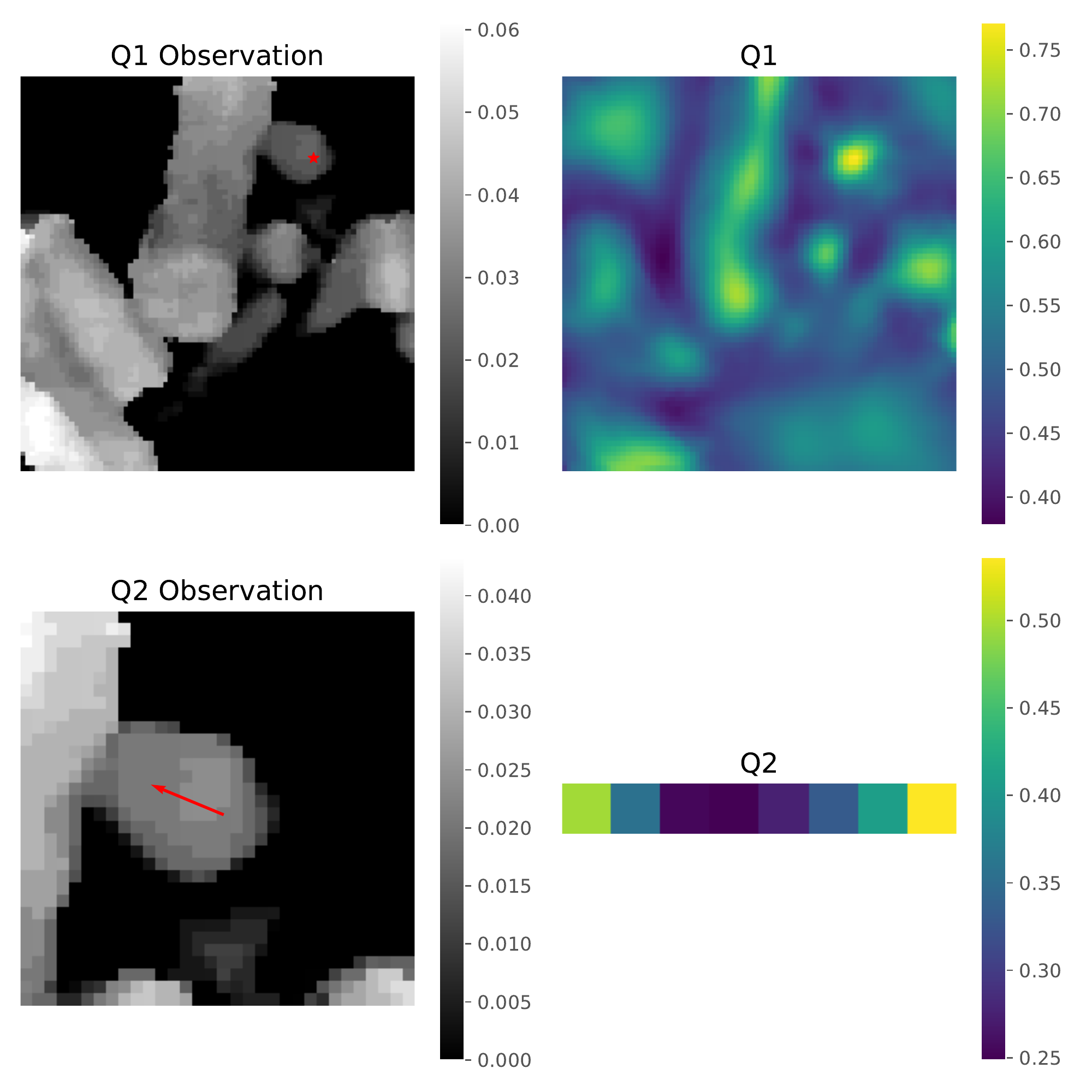}}
    \subfigure[Reasonable grasps (13/52)]{\includegraphics[width=0.22\textwidth]{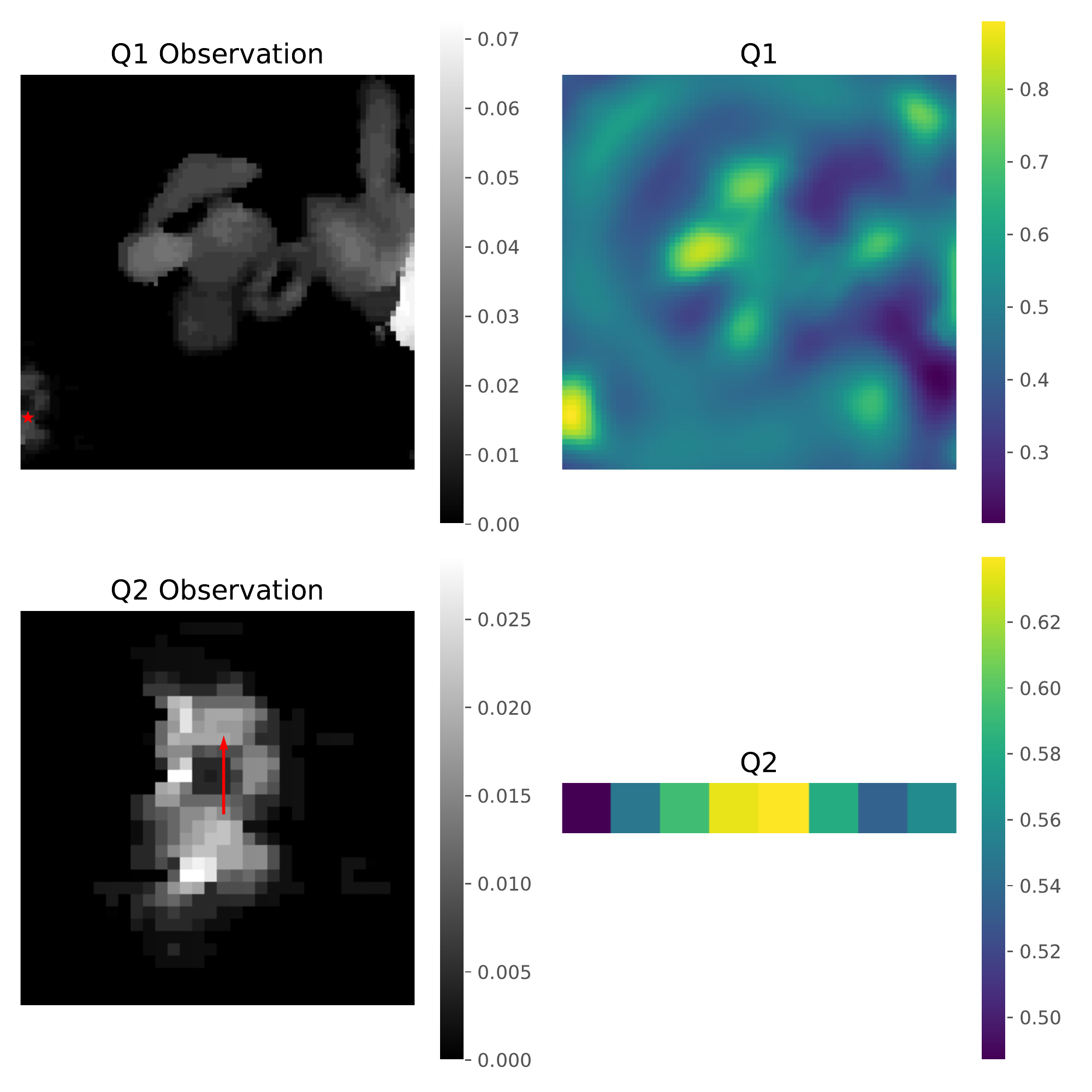}}
    \subfigure[Challenging scenes (11/52)]{\includegraphics[width=0.22\textwidth]{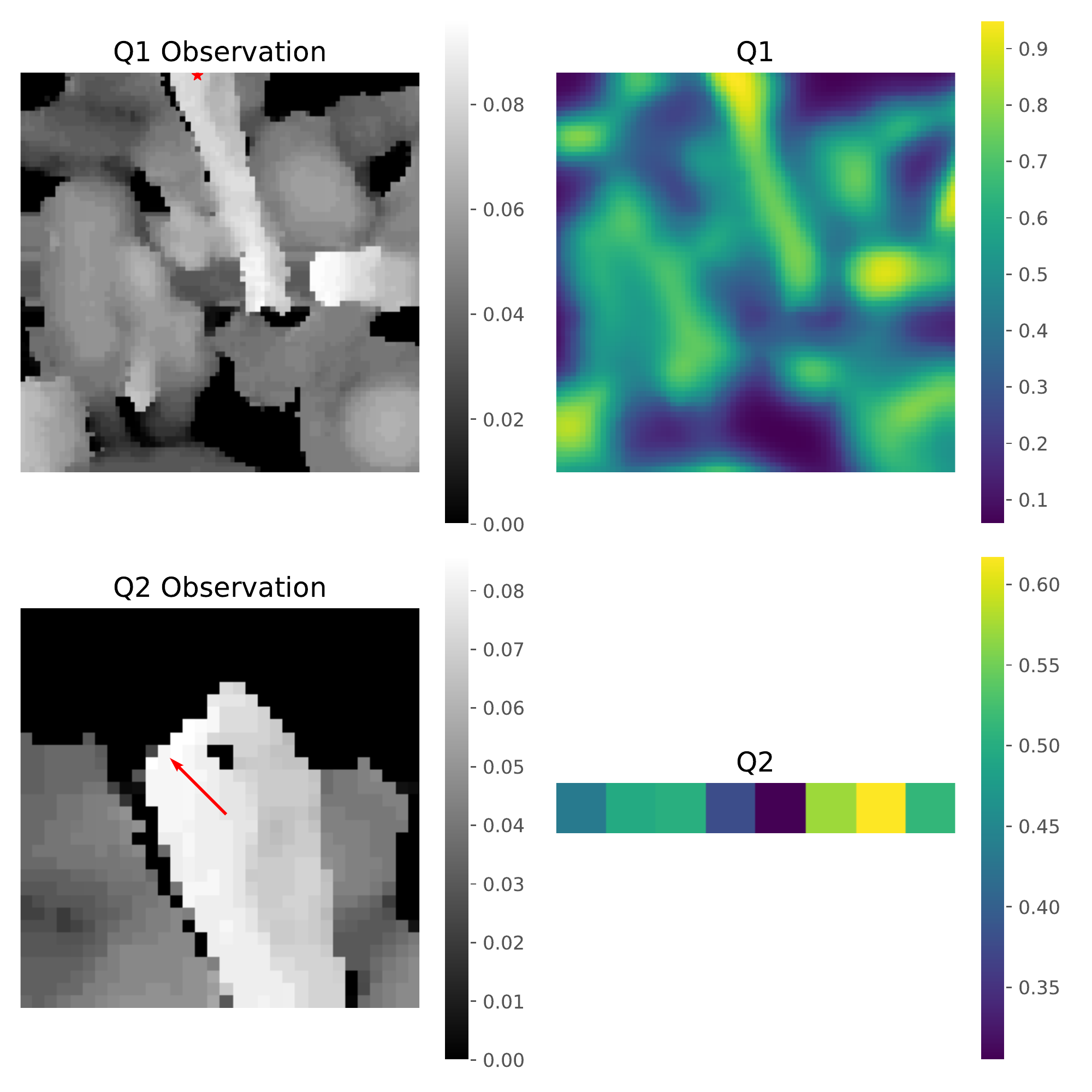}}
    \subfigure[Sensor distortion (8/52)]{\includegraphics[width=0.22\textwidth]{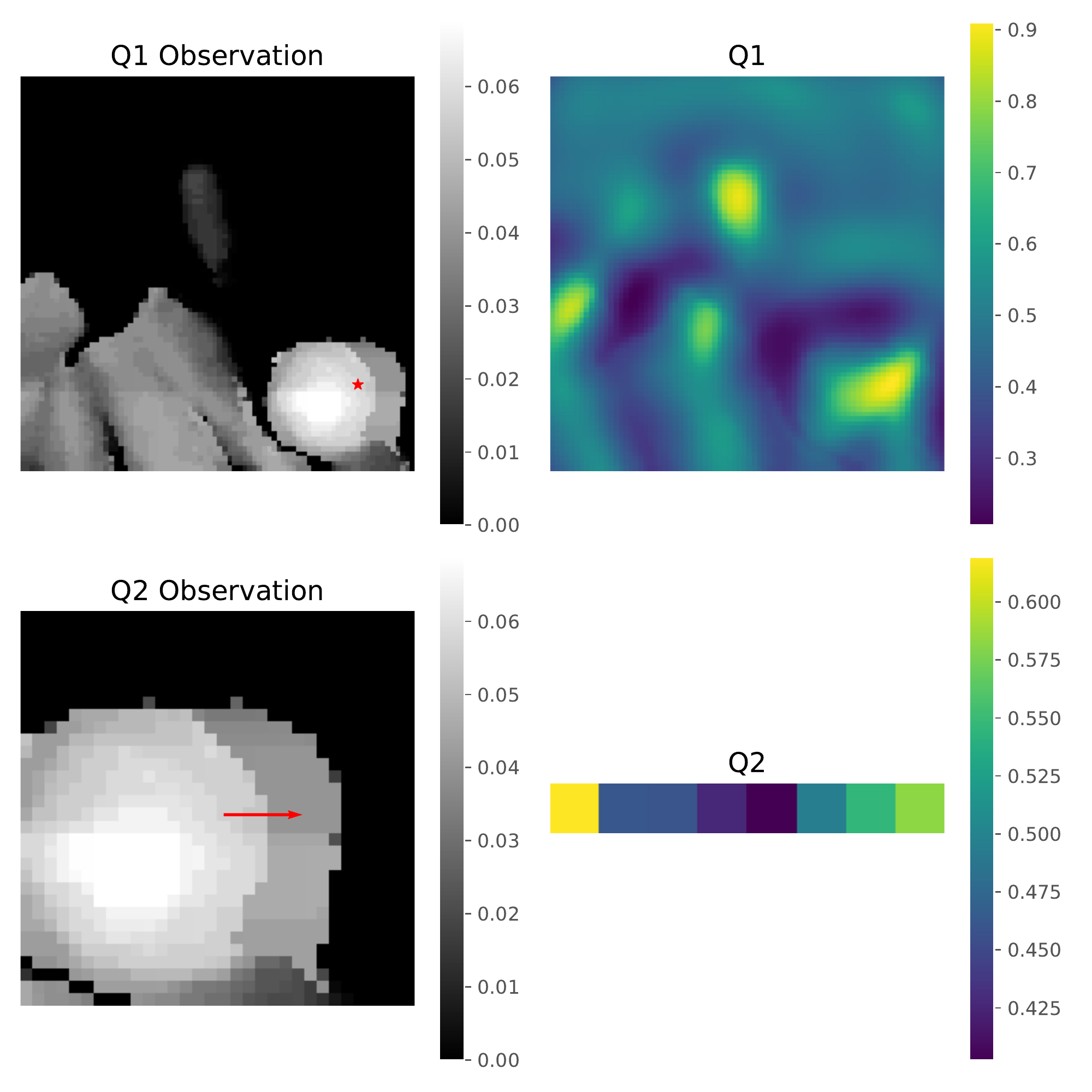}}
    \caption{Failure modes. The brackets show the failure times divided by the total number of failures in all four runs. The first row is test in the test set easy while the second row is test in the test set hard.}
    \label{fig:failure_modes}
\end{figure*}

For failure modes, we identify several typical scenarios: Bad grasp pose, either wrong translation or orientation grasp pose (Figure~\ref{fig:failure_modes}a, b, and e) indicates that there is a clear gap between our method and optimal policy. This might be caused by the biased dataset collected by the algorithm. Reasonable grasps failure (Figure~\ref{fig:failure_modes}d, f) means that the agent selects a reasonable grasp, but it fails due to the stochasticity of the real world, i.e., limited sensor resolution, contact dynamics, hardware calibration error, etc. Challenging scenes (Figure~\ref{fig:failure_modes}c, g) is the nature of densely cluttered objects, it can be alleviated by learning an optimal policy or synthesize higher DoFs grasps, e.x., predicting grasping high instead of using heuristic. The sensor distortion (Figure~\ref{fig:failure_modes}h) is caused by an low quality sensor or occlusion. Among all failure modes, wrong action selection takes the most part (65\% failure in the test set easy and 33\% failure in the test set hard). It is followed by reasonable grasps, challenging scenes, and then sensor distortion.

\section{Evaluation details in hardware}
\label{sec:evaluation_details}

The evaluation policy and environment are different from that of training in the following aspects. First, for all methods, the robot arm moves slower than that during training in the environment. This helps form stable grasps. Second, for our method, the evaluation policy uses a lower temperature ($\tau_\text{test}=0.002$) than training. After a failure grasp, ours performs $2$ SGD steps on this failure experience. The network weight will be reloaded after recovery from the failure~\cite{synergy}. For the baselines, the evaluation policy uses a greedy policy. After a failure grasp, baselines perform $8$ RAD SGD steps on this failure experience. The network weight will be reloaded after recovery from the failure~\cite{synergy}.

\end{appendices}

\end{document}